\documentclass[journal]{IEEEtran}

\usepackage{xcolor,soul,framed} 

\colorlet{shadecolor}{yellow}
\usepackage[pdftex]{graphicx}
\graphicspath{{../pdf/}{../jpeg/}}
\DeclareGraphicsExtensions{.pdf,.jpeg,.png}

\usepackage[cmex10]{amsmath}
\usepackage{array}
\usepackage{mdwmath}
\usepackage{mdwtab}
\usepackage{eqparbox}
\usepackage[utf8]{inputenc} 
\usepackage[T1]{fontenc}    
\usepackage{hyperref}       
\usepackage{url}            
\usepackage{booktabs}       
\usepackage{amsfonts}       
\usepackage{nicefrac}       
\usepackage{microtype}      
\usepackage{lipsum}
\usepackage{fancyhdr}       
\usepackage{graphicx}  
\usepackage{subcaption}  
\usepackage{xcolor}
\usepackage{amsmath}
\usepackage{subcaption}
\usepackage{pgffor}
\usepackage{subcaption}

\graphicspath{{media/}}     
\usepackage{float}
\usepackage{pifont}

\begin{document}
\bstctlcite{IEEEexample:BSTcontrol}

\title{FishDet-M: A Unified Large-Scale Benchmark for Robust Fish Detection and CLIP-Guided Model Selection in Diverse Aquatic Visual Domains
}

\author{
    Muayad~Abujabal,~Lyes~Saad~Saoud,~and~Irfan~Hussain
    \thanks{This work was conducted at the Khalifa University Center for Autonomous Robotic Systems (KUCARS), Abu Dhabi, United Arab Emirates.}
    \thanks{The authors are with Khalifa University, Abu Dhabi, UAE (e-mails: 100057733@ku.ac.ae; lyes.saoud@ku.ac.ae; irfan.hussain@ku.ac.ae).}
}

\markboth{Manuscript submitted to IEEE Transactions on Image Processing, 2025}
{Abujabal \MakeLowercase{\textit{et al.}}: FishDet-M: A Unified Large-Scale Benchmark for Robust Fish Detection}

\maketitle

\begin{abstract}
Accurate fish detection in underwater imagery is essential for ecological monitoring, aquaculture automation, and robotic perception. However, practical deployment remains limited by fragmented datasets, heterogeneous imaging conditions, and inconsistent evaluation protocols. To address these gaps, we present \textit{FishDet-M}, the largest unified benchmark for fish detection, comprising 13 publicly available datasets spanning diverse aquatic environments including marine, brackish, occluded, and aquarium scenes. All data are harmonized using COCO-style annotations with both bounding boxes and segmentation masks, enabling consistent and scalable cross-domain evaluation.
We systematically benchmark 28 contemporary object detection models, covering the YOLOv8 to YOLOv12 series, R-CNN based detectors, and DETR based models. Evaluations are conducted using standard metrics including mAP, mAP@50, and mAP@75, along with scale-specific analyses (AP$_S$, AP$_M$, AP$_L$) and inference profiling in terms of latency and parameter count. The results highlight the varying detection performance across models trained on FishDet-M, as well as the trade-off between accuracy and efficiency across models of different architectures.
To support adaptive deployment, we introduce a CLIP-based model selection framework that leverages vision-language alignment to dynamically identify the most semantically appropriate detector for each input image. This zero-shot selection strategy achieves high performance without requiring ensemble computation, offering a scalable solution for real-time applications.
FishDet-M establishes a standardized and reproducible platform for evaluating object detection in complex aquatic scenes. All datasets, pretrained models, and evaluation tools are publicly available to facilitate future research in underwater computer vision and intelligent marine systems.
\end{abstract}



%
\IEEEpeerreviewmaketitle

\begin{figure*}[t]
\centering
\includegraphics[width=0.95\linewidth]{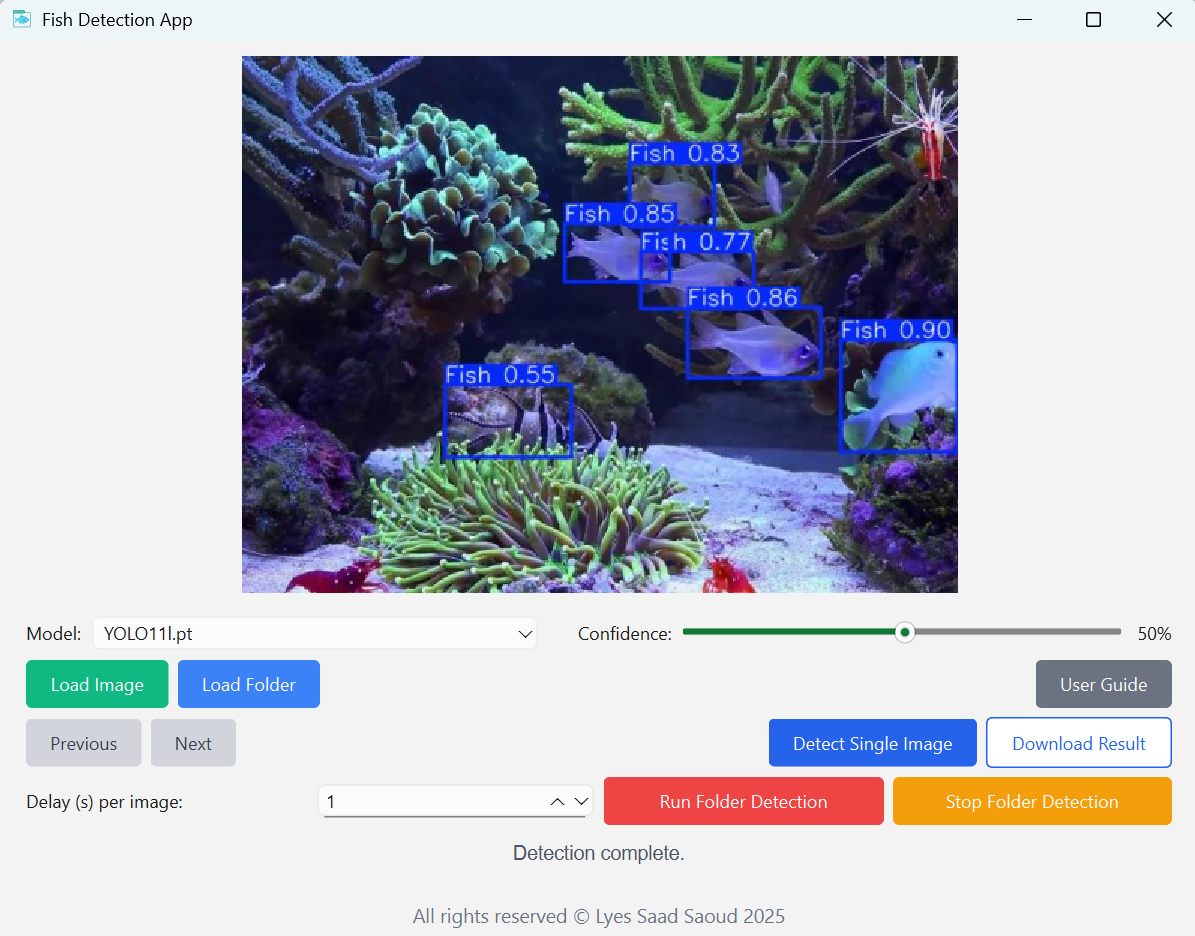}
\caption{Interactive application interface for comparing detection outputs on FishDet-M. Supports multiple models, bounding box overlays, and threshold control.}
\label{fig:interactive_app}
\end{figure*}
\section{Introduction}

Accurate fish detection in underwater environments remains a major challenge due to the visual degradation caused by turbidity, occlusion, light scattering, and dynamic backgrounds~\cite{liu2025fishfinder,salman2019realtime,fayaz2022review}. These conditions reduce contrast, distort color, and introduce clutter, impairing the reliability of vision-based detection models. Occlusions from vegetation, rocks, or dense schools of fish further obscure targets~\cite{li2023occlusion,anantha2024species}. Although deep learning has advanced underwater perception, its effectiveness is constrained by fragmented datasets that often lack diversity, consistent annotation formats, and representation of complex real-world conditions~\cite{pagire2022underwater,marrable2022species,sasithradevi2024depondfi,mohankumar2025benchmark}.

To address these limitations, several domain-specific datasets have emerged. The SmallFish dataset focuses on detecting small targets in poor visibility~\cite{liu2025fishfinder}, DUFish captures densely packed schooling behavior~\cite{jiang2024dufish}, and DePondFi introduces pond-based fish imagery from naturally challenging conditions~\cite{sasithradevi2024depondfi}. In parallel, detection models have evolved to address underwater visual distortions. FishDet-YOLO incorporates enhancement modules for low-contrast targets~\cite{yang2024fishdetyolo}, YOLOv8 TF introduces transformer-enhanced refinement and class-sensitive learning~\cite{shah2025yolov8tf}, and the IDLAFD UWSN framework leverages hybrid architectures to improve detection under blur and occlusion~\cite{duhayyim2022idlafd}.

Despite these advances, the field still lacks a unified benchmark capable of supporting large scale and cross-domain evaluation in realistic aquatic environments. To bridge this gap, we introduce \textit{FishDet-M}, a consolidated benchmark constructed by harmonizing 13 publicly available datasets into a single annotation format aligned with the COCO protocol. FishDet-M spans diverse underwater environments including coral reefs, aquaculture tanks, and brackish waters, and provides 296{,}885 annotated fish instances across 105{,}556 images. This benchmark supports the evaluation of 28 leading detection models, including YOLO variants, DETR-based architectures, and region proposal networks.

The contributions of this work are summarized as follows:

\begin{itemize}
    \item \textbf{FishDet-M Benchmark:} We introduce a unified benchmark dataset for fish detection that integrates 13 publicly available underwater datasets into a harmonized annotation format compatible with the COCO standard. The dataset includes diverse environments, species, and visual conditions.
    
    \item \textbf{Comprehensive Evaluation of Detection Models:} We evaluate 28 advanced object detection architectures including YOLOv8 through YOLOv12, DETR-based models, and R-CNN models. The evaluation uses standardized metrics such as mean average precision (mAP), scale-aware scores (AP$_S$, AP$_M$, AP$_L$), inference time, and parameter count.
    
    \item \textbf{CLIP-Based Adaptive Model Selection:} We propose a context-aware mechanism that utilizes vision-language alignment with CLIP to automatically select the most appropriate detection model based on input image content, enabling dynamic and robust inference.
    
    \item \textbf{Performance Insights and Deployment Guidelines:} We analyze model robustness under conditions such as occlusion and poor visibility, providing actionable recommendations for selecting models suitable for real-time ecological monitoring, aquaculture management, and robotic deployment.
    
    \item \textbf{Public Release to Support Reproducibility:} The full dataset, source code, checkpoints, and evaluation tools are released publicly through our GitHub repository to promote reproducible research and facilitate further development.
\end{itemize}

\section{Related Works}

\subsection{Underwater Fish Detection Challenges and Solutions}

Underwater fish detection remains a challenging task due to the complex optical properties of marine environments. Turbidity, light scattering, and occlusion degrade visibility and introduce background clutter~\cite{fayaz2022review,pagire2022underwater}, making accurate detection difficult, particularly for deformable and low contrast targets.

To mitigate these effects, recent methods combine underwater specific image enhancement, such as illumination correction and color restoration, with adapted deep learning detectors~\cite{reddy2024deepfish,yang2024fishdetyolo}. However, performance still drops in scenes with severe occlusion or high visual complexity~\cite{mao2015occlusion}.

Dataset limitations further compound these issues. Many existing datasets focus on controlled settings with limited species diversity or environmental variation~\cite{liu2025fishfinder,Liu2023MultitaskModel}. Newer datasets like DUFish~\cite{jiang2024dufish} and FishTrack23~\cite{dawkins2024fishtrack23} improve diversity and realism, but the broader dataset landscape remains fragmented. Table~\ref{tab:fish_datasets} summarizes representative datasets.

\begin{table}[t]
\caption{Representative fish detection datasets used in underwater computer vision research}
\centering
\begin{tabular}{lp{1.5cm}p{1.1cm}p{1cm}p{1.2cm}}
\toprule
Dataset & Environment & Size & Annotations & Species Diversity \\
\midrule
SmallFish~\cite{liu2025fishfinder} & Murky tanks & 5,000 & BBoxes & Low \\
DePondFi~\cite{sasithradevi2024depondfi} & Pond & 8,000 & BBoxes & Medium \\
DUFish~\cite{jiang2024dufish} & Open water & 6,300 & BBoxes and Masks & High \\
FishTrack23~\cite{dawkins2024fishtrack23} & Varied habitats & 20K & Tracks and BBoxes & High \\
OcclusionSet~\cite{li2023occlusion} & Reef zones & 3,500 & BBoxes & Medium \\
\bottomrule
\end{tabular}
\label{tab:fish_datasets}
\end{table}

Deep learning has improved detection performance with models such as YOLO, Cascade R-CNN, and DETR~\cite{yolov8_ultralytics,Cascade,detr}, enhanced by attention modules, deformable layers, and domain adaptation~\cite{yang2024fishdetyolo,chieza2025yolonas,shah2025yolov8tf}. Hybrid convolutional and transformer based architectures further improve robustness~\cite{shah2025yolov8tf}, while fine tuning on augmented datasets aids generalization to unseen conditions.

Challenges persist due to occlusion, camouflage, and interspecies variability~\cite{li2023occlusion,fabic2013population}. Addressing these requires context aware training and robust augmentation strategies. Evaluation protocols are expanding beyond mean average precision and intersection over union to include scale aware metrics, latency, and model size~\cite{liu2025fishfinder,jiang2024dufish}, supporting real time deployment.

Despite advances, cross domain variation and inconsistent benchmarks limit generalizability~\cite{saoud2024mars}. Overcoming these issues remains essential for enabling reliable fish detection in diverse marine environments.

\subsection{Fish Detection Datasets and Benchmarks}

The effectiveness of detection models relies heavily on diverse and well-annotated datasets. Many earlier datasets are restricted to clear or shallow environments, limiting their generalizability. Real-world aquatic scenes vary significantly in turbidity, lighting, and species composition, necessitating diverse benchmarks for robust evaluation~\cite{mohankumar2025benchmark,liu2025fishfinder}.

Recent efforts have addressed these limitations by expanding ecological and visual variability. DePondFi~\cite{mohankumar2025benchmark} offers annotated pond imagery where YOLO models achieve high mAP@50 scores. Fish4Knowledge supports both detection and classification in marine environments, with FishNet detector achieving over 92 percent detection precision~\cite{iqtait2024fishnet}. SmallFish is tailored to small object detection in murky conditions~\cite{liu2025fishfinder}.

FishTrack23~\cite{dawkins2024fishtrack23} emphasizes dense tracking across habitats, while LifeCLEF 2014~\cite{spampinato2014lifeclef} serves as a benchmark for fish recognition. Other datasets like UOMT~\cite{wei2024uomt} and FS48~\cite{fan2024fs48} address salient object detection and re-identification. OzFish~\cite{Saleh2020,aims2019ozfish} include species-level annotations for fish in Australian shores.

Table~\ref{tab:datasets} provides a comparative overview of these datasets, highlighting environments, tasks, and public availability. Despite their contributions, the field still lacks unified annotation standards, benchmark protocols, and cross-study comparability. Upcoming datasets such as FishDet-M aim to address these challenges through structured metadata and realistic test conditions.

\begin{table*}[t]
\centering
\caption{Representative fish detection datasets and benchmarks.}
\label{tab:datasets}
\begin{tabular}{@{}llllcl@{}}

\toprule
\textbf{Dataset} & \textbf{Environment} & \textbf{Main Task} & \textbf{Performance Highlights} & \textbf{Availability} & \textbf{Link} \\ 
\midrule
DePondFi~\cite{mohankumar2025benchmark} & Pond, South India & Real-time object detection & YOLOv8: mAP@50 = 0.92; Ensemble = 0.94 & \ding{52} & \mbox{\href{https://onlinelibrary.wiley.com/doi/full/10.4218/etrij.2024-0383}{Link}}
\\
Fish4Knowledge-2010~\cite{iqtait2024fishnet} & Marine scenes & Detection and classification & mAP = 92.3\%, Accuracy = 89.7\% & \ding{52} &  \mbox{\href{https://homepages.inf.ed.ac.uk/rbf/Fish4Knowledge/}{Link}} \\
SmallFish~\cite{liu2025fishfinder} & Murky, cluttered & Detection of small targets & Enhanced mAP via Fish-Finder algorithm & \ding{55} & -- \\
FishTrack23~\cite{dawkins2024fishtrack23} & Multi-habitat & Multi-object tracking & 20,000 expert tracks & \ding{52} & \mbox{\href{https://academictorrents.com/details/70695b973afa53be67dbfb72a2478775885598b9}{Link}} \\
LifeCLEF 2014~\cite{spampinato2014lifeclef} & Video footage & Detection and recognition & Evaluated in open competition & \ding{52} & \mbox{\href{https://www.imageclef.org/2014/lifeclef/fish}{Link}} \\
UOMT~\cite{wei2024uomt} & Mixed marine & Salient object detection & Supports multitask learning & \ding{52} & \mbox{\href{https://drive.google.com/file/d/1iZUVU1moc1nzvTsXOe1k5vdL-vI6dm81/view?usp=drive_link}{Link}} \\
FS48~\cite{fan2024fs48} & Controlled lighting & Re-identification & Multi-view detection using FSNet & \ding{55} & \mbox{\href{}{--}} \\
DeepFish~\cite{Saleh2020} & Tropical Australia & Classification and sizing & Evaluated with SOTA models & \ding{52} & \mbox{\href{https://alzayats.github.io/DeepFish/}{Link}} \\
OzFish~\cite{chieza2025yolonas} & Coastal marine & YOLO-based detection & Robust YOLOv8/NAS results & \ding{52} & \mbox{\href{https://github.com/open-AIMS/ozfish}{Link}} \\
\bottomrule
\end{tabular}
\end{table*}

\subsection{Deep Learning Models for Underwater Object Detection}

Deep learning has substantially advanced underwater detection performance, particularly in environments where image quality is degraded by turbidity, occlusion, and variable lighting~\cite{yang2024fishdetyolo,chieza2025yolonas, Saleh2020}. Foundational models like YOLO \cite{shah2025yolov8tf}, Faster R-CNN~\cite{Fasterrcnn}, and DETR have been widely adapted for marine detection pipelines.

To address visual limitations, researchers have introduced attention mechanisms, deformable layers, and fusion modules that enhance feature representation for partially visible targets. Transformers have also been incorporated into convolutional backbones, enabling long-range feature modeling in low-contrast conditions~\cite{shah2025yolov8tf,duhayyim2022idlafd}.

Specialized variants such as FishDet-YOLO and YOLOv8-TF focus on underwater data distributions and address class imbalance~\cite{yang2024fishdetyolo,shah2025yolov8tf}.Domain adaptation techniques, including adversarial learning and style transfer, further help bridge performance gaps between synthetic and real-world datasets~\cite{saoud2024review}.

Hybrid models that fuse convolution and transformer components leverage the spatial localization strengths of CNNs and the semantic context modeling of transformers. This combination enhances detection accuracy in cluttered, occluded, and dynamic marine scenes \cite{shah2025yolov8tf}.

These architectures are increasingly designed with deployment in mind. Their applications range from ecological monitoring and fishery management to robotic exploration and real-time fish tracking \cite{brackishMOT, hong2020trashcan, Lian_2023_ICCV}.

\subsection{Occlusion Handling and Species Variability}

Occlusion is a persistent challenge in underwater scenes, particularly in crowded habitats or reef zones. Solutions include 3D geometric modeling, plan mirror-based occlusion mitigation, and rotated bounding box regression~\cite{zhang2023eornet,mao2016tracking}. Transformer-enhanced YOLO models and approaches using repulsion loss have also shown improved performance on overlapping targets~\cite{li2023occlusion}.

Generative augmentation using models such as DCGAN and UIEGAN contributes occlusion-rich samples, enhancing training diversity~\cite{sudhakara2022classification,qiu2018transfer}. Temporal tracking methods further improve robustness by associating occluded objects across frames~\cite{vanessen2021tracking}.

Species variability adds further complexity due to morphological and chromatic diversity. Fine-grained models use attention mechanisms, patch localization, and multi-branch architectures to detect subtle inter-species differences~\cite{veiga2024transformers,geng2024fusion}. Camouflaged species, which exploit background matching or polarized light, remain especially challenging~\cite{brady2015polarocrypsis}.

Advanced strategies such as active detection and few-shot learning are being used to adapt models efficiently with minimal data~\cite{shah2024active}. Lightweight models like MobileNetV2, when fine-tuned with semantic modules, offer promising trade-offs between performance and computational load~\cite{berlia2024transfer}.

Table~\ref{tab:occlusion_species_strategies} outlines the key approaches for addressing occlusion and species-level variation in detection systems.

\begin{table}[t]

\centering
\caption{Summary of methods addressing occlusion, species variability, and camouflage.}
\label{tab:occlusion_species_strategies}
\begin{tabular}{@{}p{2.5cm}p{5.5cm}@{}}
\toprule
\textbf{Technique} & \textbf{Description} \\
\midrule
3D tracking models~\cite{mao2015occlusion,mao2016tracking} & Geometric modeling for occlusion tracking using mirror imaging and spatial cues  \\
Rotating box detection~\cite{zhang2023eornet} & Uses oriented boxes to resolve overlaps in dense scenes  \\
Enhanced YOLO models~\cite{li2023occlusion} & Incorporates modules and repulsion loss for occlusion robustness  \\
Synthetic data generation~\cite{sudhakara2022classification,qiu2018transfer} & Adversarial networks used to balance datasets with rare and occluded samples  \\
Underwater image enhancement~\cite{sudhakara2022classification} & Improves contrast and clarity prior to detection  \\
Multi-object tracking~\cite{vanessen2021tracking} & Maintains temporal consistency in sequential data  \\
Vision transformers with FGVC~\cite{veiga2024transformers} & Discriminative attention modules for fine-grained classification  \\
Dual-branch fusion networks~\cite{geng2024fusion} & Integrates inter-species similarity and semantic fusion  \\
Attention mechanisms~\cite{zhai2023sacovamnet} & Enhances localized feature extraction in camouflage detection  \\
Polarocrypsis~\cite{brady2015polarocrypsis} & Biological camouflage using polarization cues  \\
Active detection models~\cite{shah2024active} & Epistemic uncertainty used for sample selection  \\
Transfer learning~\cite{berlia2024transfer} & Fine-tunes pretrained models using edge and texture-aware modules  \\
\bottomrule
\end{tabular}
\end{table}

\subsection{Evaluation Metrics and Benchmarking Protocols}

Reliable benchmarking in underwater detection depends on consistent and comprehensive evaluation criteria. Standard metrics such as mAP, IoU, precision, and recall are commonly reported~\cite{tejas2024comparison,assem2024rov}, but variations in implementation and dataset usage often hinder reproducibility~\cite{shen2025review}.

To address detection at different object scales, researchers employ APS, APM, and APL metrics that reflect performance on small, medium, and large objects. These are critical for understanding behavior in turbid conditions, where smaller fish are more likely to be missed~\cite{ma2024entropy,bhalla2024dataset}.

Practical deployments also require runtime metrics such as inference time, memory usage, and model size. These determine whether models can be implemented on embedded devices or underwater robots with limited computational resources~\cite{liu2023sonar}.

Image quality assessment metrics, including UWEQM~\cite{guo2023uweqm} and QDE~\cite{shen2025review}, are also important. These help quantify the impact of enhancement and restoration steps on overall detection reliability. Recent multi-exposure and fusion-based schemes offer richer evaluations that integrate contrast, chroma, and structural consistency~\cite{jiang2025dimension}.

Standardizing benchmarking protocols and dataset splits will improve comparability across studies. New efforts should aim to combine performance metrics with deployment considerations and perceptual quality assessments to guide model selection for real-world marine applications.
\subsection{Applications in Aquaculture, Robotics, and Marine Monitoring}

Fish detection models play a vital role across aquaculture, robotics, and environmental monitoring. In aquaculture, deep learning enables automated fish counting, size estimation, and health assessment under turbid and low-light conditions~\cite{elmezain2025underwater,akram2022visual,akram2024aquaculture}. Lightweight models such as AquaYOLO and CUIB YOLO~\cite{vijayalakshmi2025aquayolo,zhang2024YOLOv8n} are optimized for embedded deployment and integrate with smart feeding and water monitoring systems~\cite{teixeira2022feedfirst}.

In underwater robotics, fish detection supports navigation, obstacle avoidance, and targeted sampling, which are essential for autonomous missions such as coral reef surveys, pollution tracking, and marine mapping~\cite{ryuh2015robotic,wang2023swarm,ahmed2023vision,din2023marine}. Recent works have also emphasized aquaculture infrastructure inspection and defect segmentation using semantic segmentation and detection pipelines~\cite{akram2024enhancing}.

In conservation, detection models assist biodiversity tracking, mariculture zone mapping, and ecosystem health monitoring. Integration with drones and satellite imaging enables detection of trends driven by environmental changes and human activity~\cite{zhang2025mariculture,falconer2025climate,saoud2024beyond}.

Despite progress, variability in species, morphology, and water conditions continues to challenge robustness. Efforts in model compression, domain adaptation, and multi-sensor fusion are ongoing. Efficient models such as AASNet and quantized YOLO variants are promising for edge deployment in marine settings~\cite{kong2025aasnet,wang2021rird}. These applications demonstrate the growing importance of fish detection in sustainable aquaculture, autonomous exploration, and ecological assessment~\cite{saoud2024mars,elmezain2025underwater,saoud2024beyond,din2025benchmarking}.

\section{Data Records}
\label{sec:data-records}

The FishDet-M benchmark comprises 105{,}556 images and 296{,}885 annotated fish instances, partitioned into training, validation, and test sets using stratified sampling to ensure balanced representation across habitat types, visibility conditions, and fish densities. Key statistics for each split are summarized in Table~\ref{tab:dataset_stats}.

\begin{table}[t]
\centering
\caption{Overview of the FishDet-M dataset splits and annotations.}
\label{tab:dataset_stats}
\begin{tabular}{lccc}
\toprule
\textbf{Feature} & \textbf{Training Set} & \textbf{Validation Set} & \textbf{Test Set} \\
\midrule
Number of Images & 83{,}093 & 10{,}654 & 11{,}809 \\
Number of Fish Instances & 228{,}558 & 32{,}961 & 35{,}366 \\
Annotated Species & $>$50 & $>$50 & $>$50 \\
Bounding Boxes & Yes & Yes & Yes \\
Environmental Diversity & Broad & Broad & Broad \\
\bottomrule
\end{tabular}
\end{table}

FishDet-M exhibits extensive visual heterogeneity, with image resolutions ranging from 78$\times$53 to 4608$\times$3456 pixels (mean: 715$\times$468) and instance densities spanning from single fish to dense aggregations with up to 256 individuals per frame (mean: 2.81 instances per image). 

\begin{figure}[t]
\centering
\includegraphics[width=\linewidth]{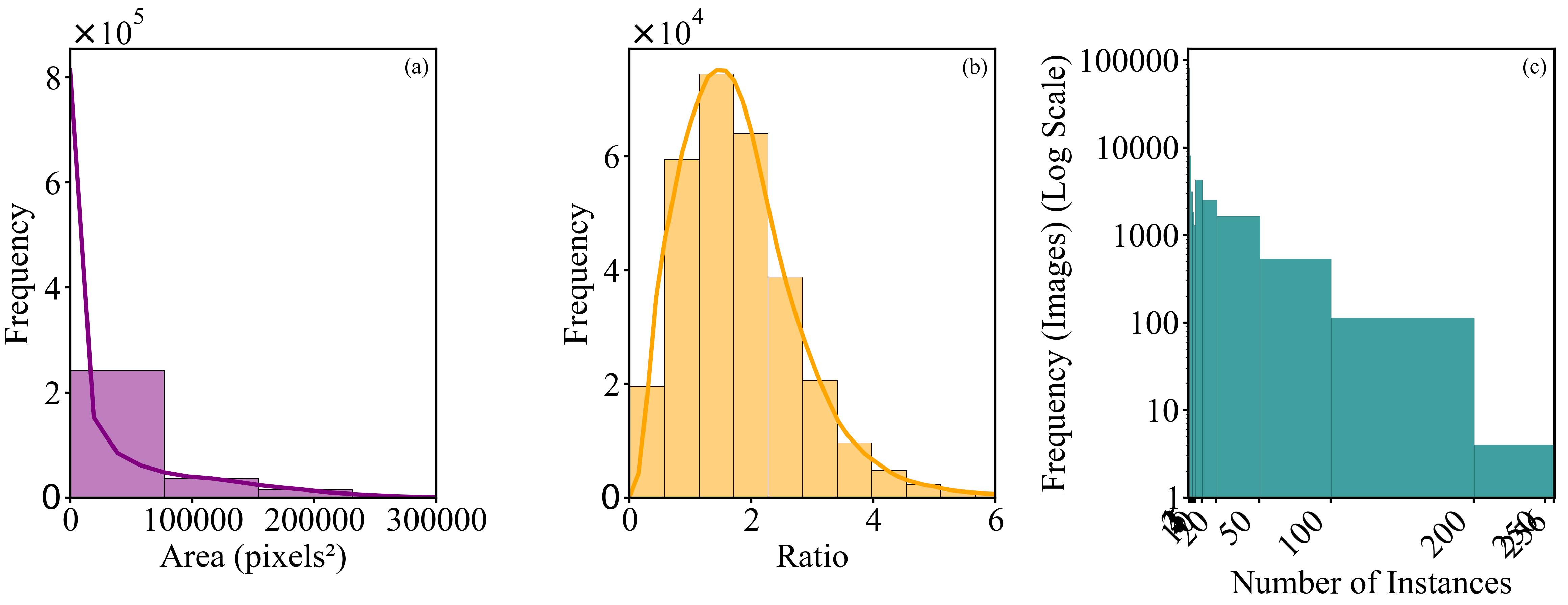}
\begin{subfigure}{0.3\linewidth}
    \phantomsubcaption
    \label{fig:bbox_characteristics:a}
\end{subfigure}
\begin{subfigure}{0.3\linewidth}
    \phantomsubcaption
    \label{fig:bbox_characteristics:b}
\end{subfigure}
\begin{subfigure}{0.3\linewidth}
    \phantomsubcaption
    \label{fig:bbox_characteristics:c}
\end{subfigure}

\caption{Distribution of object-level statistics across FishDet-M. (a) Bounding box area. (b) Aspect ratio. (c) Number of instances per image (log scale).}
\label{fig:bbox_characteristics}
\end{figure}

Object-level statistics are visualized in Fig.~\ref{fig:bbox_characteristics}. The distribution of bounding box areas in Fig. ~\ref{fig:bbox_characteristics:a} shows a right-skewed curve dominated by small and medium-sized objects. Aspect ratios in Fig.~\ref{fig:bbox_characteristics:b} peak between 1 and 2. The histogram of instances per image in Fig.~\ref{fig:bbox_characteristics:c}, plotted on a log scale, reveals a long-tail pattern where most images contain few fish while a smaller subset includes densely packed schools.

\begin{figure}[t]
\centering
\includegraphics[width=\linewidth]{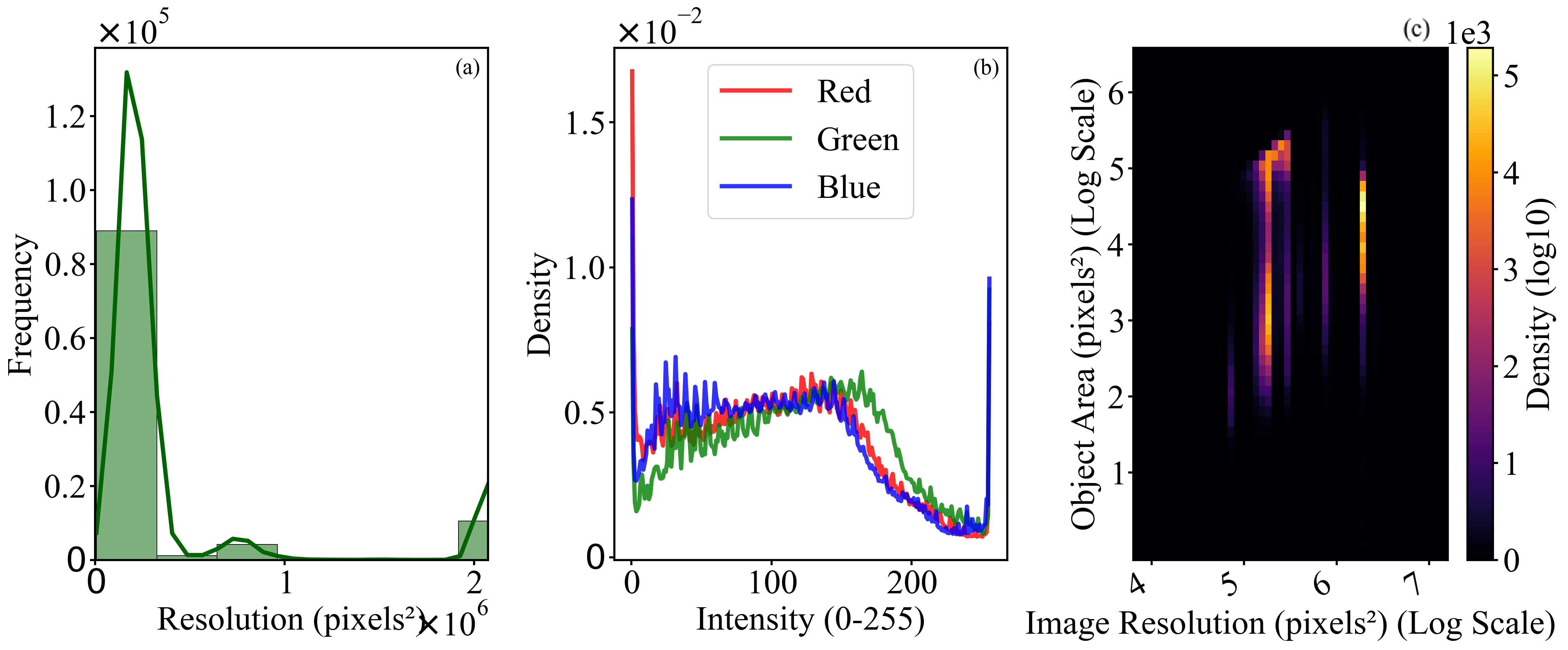}

\begin{subfigure}{0.3\linewidth}
    \phantomsubcaption
    \label{fig:resolution_rgb_obj_res:a}
\end{subfigure}
\begin{subfigure}{0.3\linewidth}
    \phantomsubcaption
    \label{fig:resolution_rgb_obj_res:b}
\end{subfigure}
\begin{subfigure}{0.3\linewidth}
    \phantomsubcaption
    \label{fig:resolution_rgb_obj_res:c}
\end{subfigure}

\caption{Visual property distributions. (a) Image resolution. (b) RGB intensity histograms. (c) Object area vs. image resolution (log-log scale).}
\label{fig:resolution_rgb_obj_res}
\end{figure}

Visual properties at the image level are summarized in Fig.~\ref{fig:resolution_rgb_obj_res}. The image resolution histogram in Fig.~\ref{fig:resolution_rgb_obj_res:a} illustrates the diversity in spatial coverage. RGB intensity histograms in Fig.~\ref{fig:resolution_rgb_obj_res:b} reveal that most scenes are low-light and dominated by green-blue hues, typical of underwater capture \cite{usod10k}. The object-to-image resolution heatmap in Fig.~\ref{fig:resolution_rgb_obj_res:c}, shown on a log-log scale, emphasizes the prevalence of small objects across both low- and high-resolution imagery.

\begin{figure}[t]
\centering
\includegraphics[width=\linewidth]{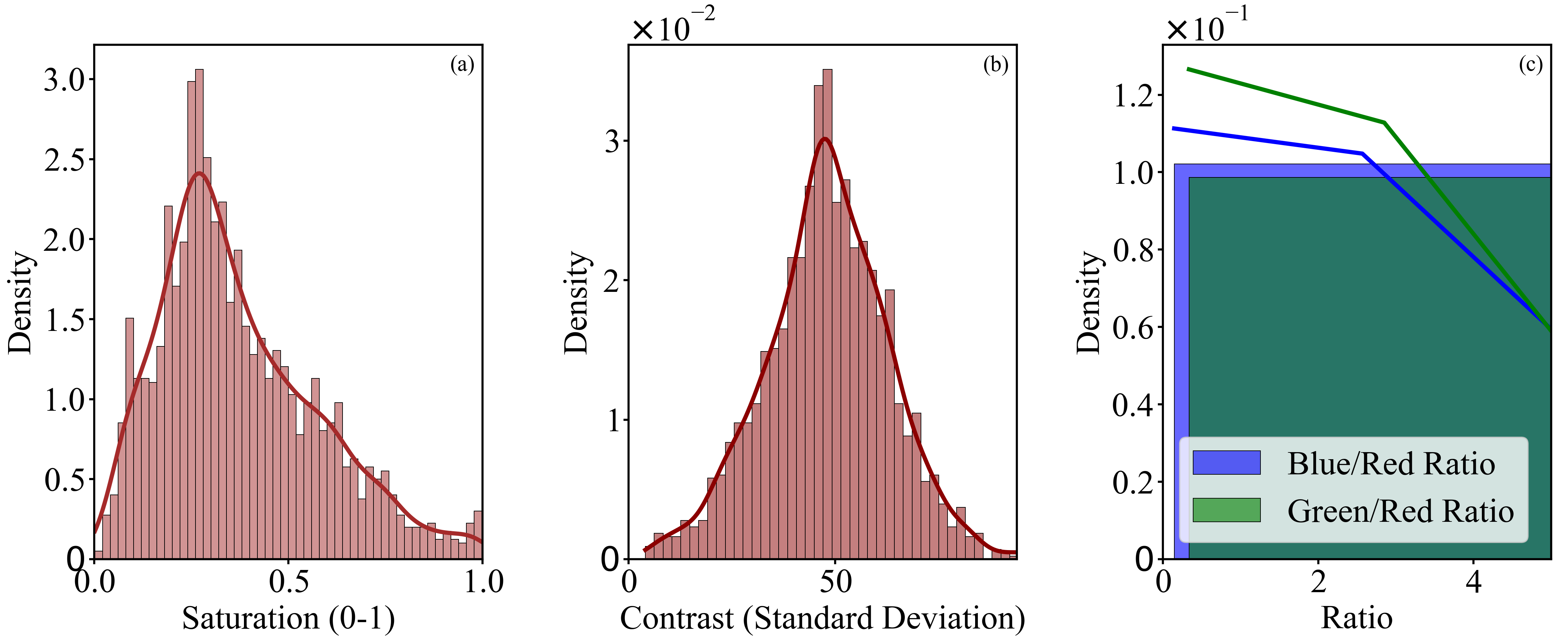}

\begin{subfigure}{0.3\linewidth}
    \phantomsubcaption
    \label{fig:saturation_contrast_tint:a}
\end{subfigure}
\begin{subfigure}{0.3\linewidth}
    \phantomsubcaption
    \label{fig:saturation_contrast_tint:b}
\end{subfigure}
\begin{subfigure}{0.3\linewidth}
    \phantomsubcaption
    \label{fig:saturation_contrast_tint:c}
\end{subfigure}

\caption{Underwater visual characteristics. (a) Saturation. (b) Contrast (std. dev.). (c) Color channel ratios (Blue/Red, Green/Red).}
\label{fig:saturation_contrast_tint}
\end{figure}

Additional underwater-specific imaging characteristics are highlighted in Fig.~\ref{fig:saturation_contrast_tint}. The saturation histogram in Fig.~\ref{fig:saturation_contrast_tint:a} shows a dominance of muted tones, consistent with light absorption and scattering . Contrast distribution in Fig.~\ref{fig:saturation_contrast_tint:b}, computed via grayscale standard deviation, confirms low overall contrast. The channel ratio plots in Fig.~\ref{fig:saturation_contrast_tint:c} illustrate strong blue and green biases relative to red, confirming the spectral distortion typical of aquatic environments \cite{usod10k}.

\begin{figure}[t]
\centering
\includegraphics[width=\linewidth]{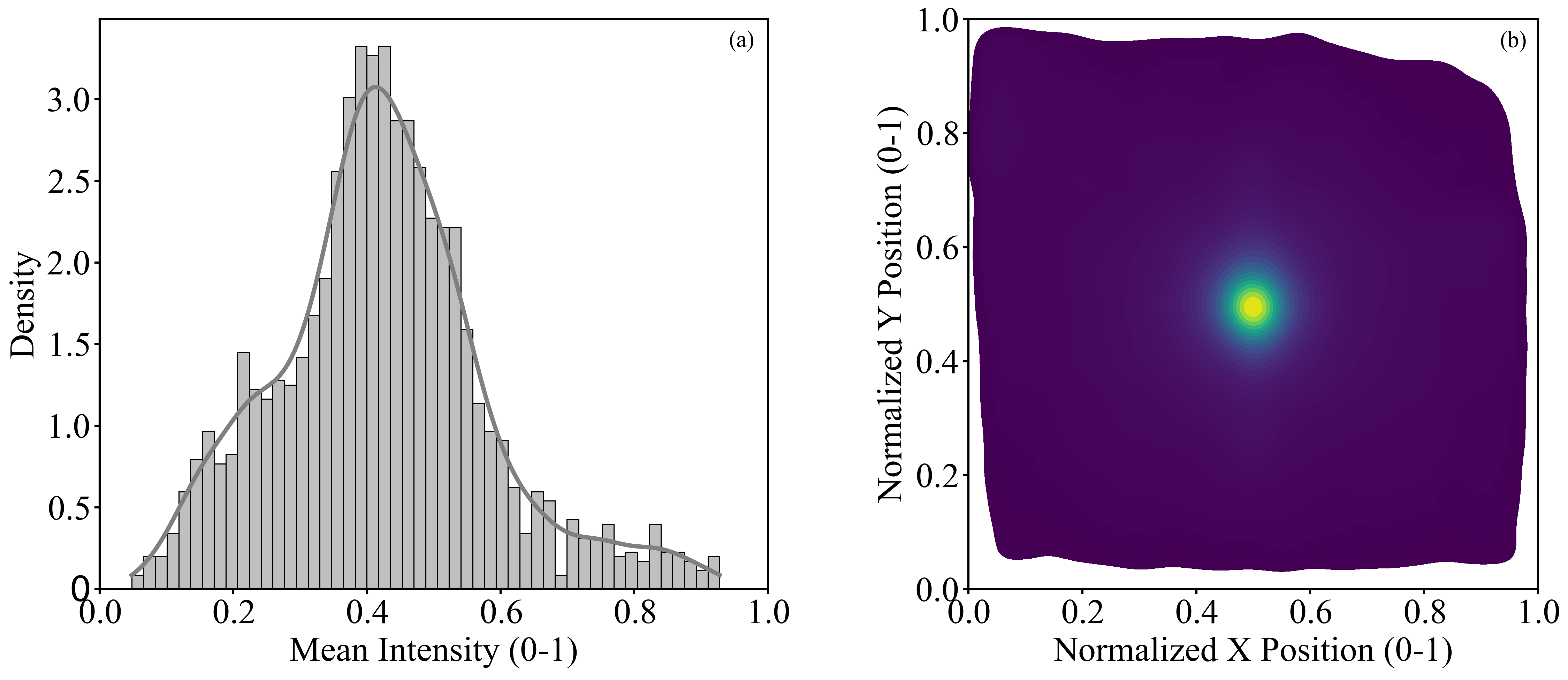}

\begin{subfigure}{0.45\linewidth}
    \phantomsubcaption
    \label{fig:haze_heatmap:a}
\end{subfigure}
\begin{subfigure}{0.45\linewidth}
    \phantomsubcaption
    \label{fig:haze_heatmap:b}
\end{subfigure}

\caption{Scene-level statistics. (a) Image mean intensity (haze proxy). (b) Spatial distribution of bounding box centers.}
\label{fig:haze_heatmap}
\end{figure}

Additional spatial insights are provided in Fig.~\ref{fig:haze_heatmap}. The histogram of mean image intensities in Fig.~\ref{fig:haze_heatmap:a} serves as a proxy for visibility degradation due to haze and backscatter. The bounding box center heatmap in Fig.~\ref{fig:haze_heatmap:b} reveals a central bias in object placement, reflecting typical framing tendencies in human-operated data acquisition.

To facilitate exploratory analysis and model transparency, we developed an interactive desktop GUI as shown in Fig.~\ref{fig:interactive_app}. The tool enables comparative evaluation of detection outputs from YOLO-based and transformer-based models, with support for adjusting confidence thresholds and overlaying predicted bounding boxes. Additional metadata such as mAP scores and inference times are presented in real time, assisting researchers in identifying model suitability for varied aquatic conditions.

\section{Methods}
\label{sec:methods}

\textbf{Dataset Aggregation and Annotation.} FishDet-M consolidates 13 underwater fish detection datasets, combining publicly available repositories and datasets. Data originate from coral reefs, aquariums, coastal zones, and estuaries, comprising both still images and video frames. All annotations were standardized to the COCO format\cite{cocosite}, with bounding boxes harmonized into a unified, species-agnostic \textit{fish} category. This ensured annotation integrity across varying formats and naming conventions. Invalid or ambiguous annotations were removed during quality control.

\textbf{Environmental and Visual Diversity.} FishDet-M covers a broad range of real-world conditions including clear and turbid waters, shallow and deep scenes, artificial and natural lighting, motion blur, and object occlusion. These variations simulate practical deployment scenarios and improve model robustness. The dataset also includes terrestrial and laboratory views for extended diversity and evaluation of domain transfer.

\textbf{Harmonization and Quality Assurance.} Dataset integration involved coordinate format unification, validation of box dimensions, and manual correction of missing or erroneous annotations. All bounding boxes follow the COCO format $x_{\text{min}}, y_{\text{min}}, \text{width}, \text{height}$~\cite{cocosite}. Rigorous validation steps ensured dataset consistency and excluded corrupted samples.

\textbf{Partitioning Strategy.} A source-aware stratified split maintained proportional representation of each dataset within training (80\%), validation (10\%), and test (10\%) sets. This preserved the ecological and visual diversity of smaller datasets and prevented bias from large contributors like FishNet\cite{10377207}.

\begin{figure}[t]
\centering
\includegraphics[width=\linewidth]{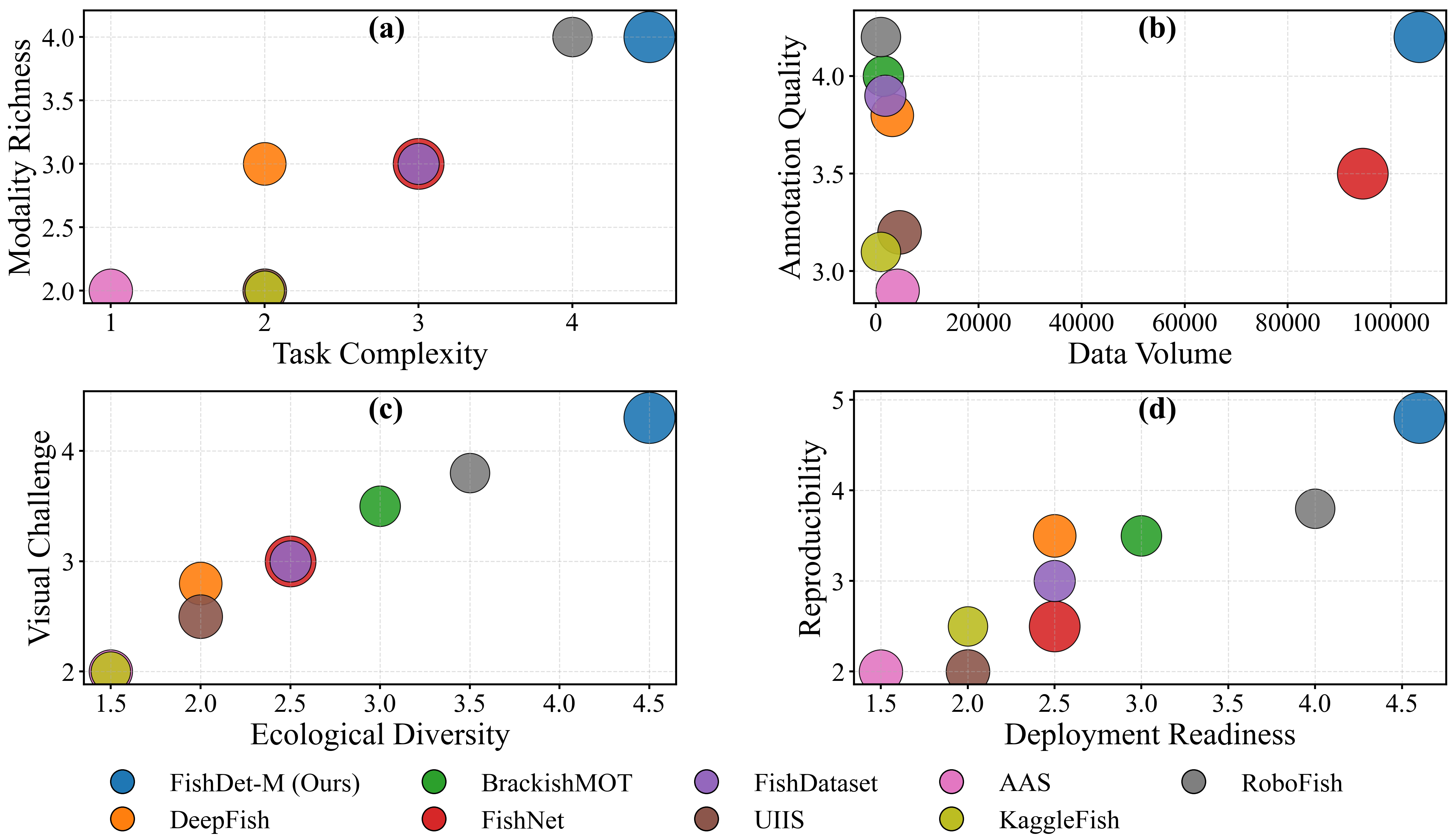}
\caption{Comparative positioning of major fish detection datasets across four key axes: (a) task complexity vs. modality richness, (b) data volume vs. annotation quality, (c) ecological diversity vs. visual challenge, and (d) deployment readiness vs. reproducibility. Bubble size indicates dataset scale (number of annotated fish instances).}
\label{fig:dataset_benchmark_bubble}
\end{figure}

\textbf{Dataset Selection Criteria.} Selection was guided by eight criteria: task complexity, modality richness, volume, annotation quality, ecological diversity, visual challenge and deployment readiness. Representative datasets included DeepFish~\cite{Saleh2020}, FishNet~\cite{10377207}, Brackish-MOT~\cite{brackishMOT}, and TrashCan 1.0~\cite{hong2020trashcan}. The complete list of source datasets is listed in table~\ref{tab:dataset-summary}. Non-representative or duplicate frames were removed to focus on annotated fish instances and prevent data leaks. To visualize representativeness, Fig.~\ref{fig:dataset_benchmark_bubble} presents a four-panel bubble chart contrasting key dimensions (e.g., task complexity vs. modality richness), with FishDet-M consistently positioned in the upper-right quadrant, highlighting its diversity, annotation volume, and readiness for real-world deployment.

\begin{figure}[t]
    \centering
    \includegraphics[width=1\linewidth]{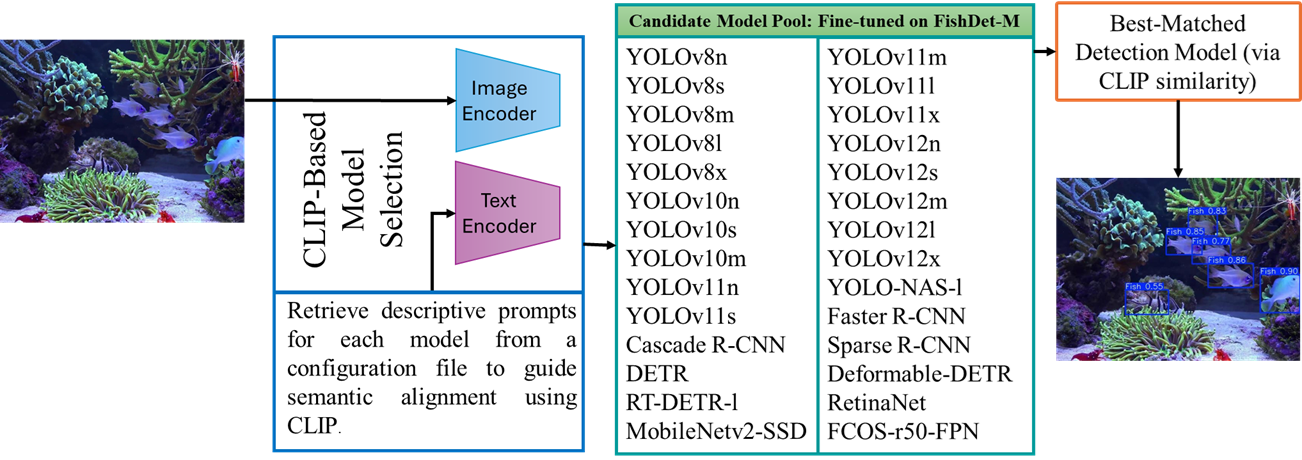}
    \caption{CLIP-based model selection pipeline. The input image is embedded via the CLIP image encoder, and its similarity is computed with a set of model-specific textual prompts embedded using the CLIP text encoder. The model with the highest alignment score is automatically selected for downstream detection.}
    \label{fig:clip_model_selection}
\end{figure}

\textbf{CLIP-Based Model Selection.} We integrated a CLIP-based module to automatically select the most appropriate detection model from 28 candidates using semantic similarity between image embeddings and prompt texts as shown in Fig.~\ref{fig:clip_model_selection}. This mechanism enables context-aware inference without manual selection or heuristics.

\textbf{Technical Validation.} We benchmarked 28 models across YOLOv8--v12 \cite{yolov8_ultralytics,THU-MIGyolov10,yolo11_ultralytics,yolo12}, YOLO-NAS\cite{supergradients}, Cascade R-CNN\cite{Cascade}, Sparse R-CNN\cite{Sparsercnn}, DETR variants\cite{detr, RTDETR, Deformable}, RetinaNet\cite{Retinanet}, and MobileNetv2-SSD\cite{SSD}. Models were trained on FishDet-M using defualt hyperparameters. Evaluation used mAP, mAP$_{50}$, mAP$_{75}$, AP$_S$, AP$_M$, and AP$_L$, precision-recall curves, and F1 scores. Efficiency was assessed via FPS, model size, and latency.

\textbf{Hardware and Software.} All experiments were run on an RTX 4090 GPU with an Intel i9-14900K CPU and 64GB RAM under Windows 11. Codebases included PyTorch \cite{paszke2019pytorchimperativestylehighperformance}, Ultralytics \cite{Jocher_Ultralytics_YOLO_2023} and MMDetection\cite{chen2019mmdetectionopenmmlabdetection}. All detection metrics were computed using pycocotools\cite{pycocotools} with mixed-precision evaluation.

\begin{table*}[t]
\caption{Summary of existing datasets used as image sources for constructing our comprehensive FishDet-M dataset, detailing their types, sizes, and descriptions}
\centering
\begin{tabular}{@{}llll@{}}
\toprule
\textbf{Dataset} & \textbf{Type} & \textbf{Size} & \textbf{Description} \\
\midrule
Brackish-MOT~\cite{brackishMOT} & MOT & 89 videos & Hazy underwater videos with multi-object fish tracking bounding boxes \\
UIIS~\cite{Lian_2023_ICCV} & Instance Seg. & 4628 images & General-purpose underwater instance segmentation across categories \\
TrashCan 1.0~\cite{hong2020trashcan} & Object Detection & 7212 images & Debris detection dataset including fish annotations \\
Fish4Knowledge-2023~\cite{fish4knowledge-dataset_dataset} & Fish Detection & 1897 images & Bounding-box labeled images of fish from underwater videos \\
Fish Dataset~\cite{Liu2023MultitaskModel} & Detection, Seg. & 1850 images & Aquarium images of Crucian carp with detection and segmentation masks \\
FishNet~\cite{10377207} & Cls., Det & 94,532 images & Large-scale dataset for multi-species classification with bounding boxes \\
Fish Video~\cite{fish-video-ls42k_dataset} & Fish Detection & 322 images & Video-derived frames with annotated fish instances \\
FISH-video ~\cite{fish-video_dataset} & Fish Detection & 506 images &  Fish tank video frames with annotated fish instances \\
Med. Fish~\cite{10.3389/fmars.2023.1151758} & Fish Detection & 1247 images & Labeled fish images from Mediterranean Sea species \\
DeepFish~\cite{Saleh2020} & Fish Det., Seg. & 3200 / 620 & Images with labeled fish instances in tropical Australia \\
Aquarium~\cite{detect-aqurium_dataset} & Fish Detection & 638 images & Aquarium animal dataset with fish-focused bounding boxes \\
AAS~\cite{Le2023} & Fish Detection & 4239 images & Multi-class marine dataset with 72\% fish representation \\
FishExtend~\cite{fish-clean_dataset} & Fish Detection & 3122 images & Diverse fish images under various environmental conditions \\
\hline
\textbf{FishDet-M (Ours)} & \textbf{Fish Detection} & \textbf{105,556 images} & Unified benchmark combining 13 datasets for large-scale, diverse fish detection \\
\bottomrule
\end{tabular}
\label{tab:dataset-summary}
\end{table*}

\section{Results and Discussion}

This section provides a comprehensive evaluation of object detection models for underwater fish detection, presenting both quantitative performance metrics and qualitative analyses of model behavior. We assess model performance on the aggregate FishDet-M benchmark, examine their efficiency characteristics, and investigate their generalization capabilities across various sub-test splits, reflecting diverse underwater conditions.
\subsection{Quantitative Evaluation}

We evaluated 28 object detection models on the FishDet-M benchmark to analyze performance and computational efficiency across a range of architectures. Table~\ref{tab:detection-test} presents accuracy, scale-specific detection, and inference metrics.

The YOLO family consistently achieved the highest detection accuracy. YOLO12x reached 0.491 mAP, followed by YOLO12l (0.487) and YOLO11l (0.484). These models performed strongly across small, medium, and large object scales (AP$_S$, AP$_M$, AP$_L$), confirming their adaptability to varying fish sizes. YOLO-NAS-l also provided strong results (0.470 mAP) with efficient inference.

Lighter models such as YOLOv8n and YOLOv10n offered excellent speed-accuracy balance. YOLOv8n achieved 251 FPS using only 3.2 million parameters, making it ideal for real-time and resource-constrained scenarios. 


In contrast, transformer-based models including DETR, RT-DETR-l, and Deformable-DETR showed lower mAP (0.317–0.390), struggled with small object detection, and exhibited higher latency (15 to 45 milliseconds), making them less suitable for real-time deployment in aquatic environments.

\begin{table*}
\small
\caption{Detection performance summary of 28 deep learning object detection models, including YOLO, R-CNN, and Transformer-based architectures, evaluated on the FishDet-M test set. Metrics reported include mAP@50, mAP@50:95, and inference speed (ms), Training is done using Ultralytics \cite{Jocher_Ultralytics_YOLO_2023}, mmdetection \cite{chen2019mmdetectionopenmmlabdetection} and Supergradients \cite{supergradients}}
\centering
\begin{tabular}{lccccccccr}
\toprule
Model & mAP & mAP$_{50}$ & mAP$_{75}$ & AP$_S$ & AP$_M$ & AP$_L$ & Params (M) & Inf. Speed (ms) & FPS \\
\midrule
YOLOv8n \cite{yolov8_ultralytics}          & 0.433 & 0.776       & 0.445       & 0.251  & 0.409  & 0.602  & {3.2}          & \textbf{3.97}                  & \textbf{251.78}   \\
YOLOv8s \cite{yolov8_ultralytics}          & 0.455 & 0.808       & 0.469       & 0.274  & 0.434  & 0.618  & 11.2          & 4.02                   & 248.85   \\
YOLOv8m \cite{yolov8_ultralytics}          & 0.471 & 0.828       & 0.490       & 0.286  & 0.456  & 0.632  & 25.9          & 5.10                   & 196.01   \\
YOLOv8l \cite{yolov8_ultralytics}          & 0.478 & 0.834       & 0.505       & 0.294  & 0.466  & 0.638  & 43.7          & 6.12                   & 163.50   \\
YOLOv8x \cite{yolov8_ultralytics}          & 0.481 & 0.837       & 0.505       & 0.297  & 0.472  & 0.638  & 68.2          & 7.07                   & 141.52   \\
YOLOv10n \cite{THU-MIGyolov10}       & 0.442 & 0.784       & 0.460       & 0.261  & 0.413  & 0.610  & 2.3           & 4.98                   & 200.77   \\
YOLOv10s \cite{THU-MIGyolov10}         & 0.461 & 0.812       & 0.483       & 0.277  & 0.439  & 0.626  & 7.2           & 5.03                   & 198.91   \\
YOLOv10m \cite{THU-MIGyolov10}         & 0.476 & 0.833       & 0.501       & 0.295  & 0.461  & 0.633  & 15.4          & 6.65                   & 150.41   \\
YOLO11n \cite{yolo11_ultralytics}          & 0.450 & 0.796       & 0.465       & 0.262  & 0.426  & 0.620  & 2.6           & 5.24                   & 190.68   \\
YOLO11s \cite{yolo11_ultralytics}          & 0.471 & 0.824       & 0.491       & 0.283  & 0.460  & 0.629  & 9.4           & 5.27                   & 189.71   \\
YOLO11m \cite{yolo11_ultralytics}          & 0.480 & 0.838       & 0.502       & 0.297  & 0.468  & 0.635  & 20.1          & 6.52                   & 153.45   \\
YOLO11l \cite{yolo11_ultralytics}          & 0.484 & 0.842       & 0.506       & 0.302  & 0.470  & 0.639  & 25.3          & 9.72                   & 102.90   \\
YOLO11x \cite{yolo11_ultralytics}          & 0.483 & 0.840       & 0.509       & 0.300  & 0.472  & 0.639  & 56.9          & 9.76                   & 102.40   \\
YOLO12n \cite{yolo12}          & 0.443 & 0.789       & 0.456       & 0.255  & 0.415  & 0.616  & \textbf{2.6}           & 7.56                   & 132.20   \\
YOLO12s \cite{yolo12}          & 0.470 & 0.823       & 0.490       & 0.281  & 0.451  & 0.634  & 9.3           & 7.76                   & 128.89   \\
YOLO12m \cite{yolo12}          & 0.483 & 0.841       & 0.507       & 0.299  & 0.470  & \textbf{0.642}  & 20.2          & 8.12                   & 123.17   \\
YOLO12l \cite{yolo12}          & 0.487 & 0.847       & 0.516       & 0.306  & 0.483  & 0.639  & 26.4          & 13.66                  & 73.20    \\
YOLO12x \cite{yolo12}          & \textbf{0.491} & \textbf{0.848}       & \textbf{0.521}       & \textbf{0.315}  & \textbf{0.485}  & 0.641  & 59.1          & 13.41                  & 74.59    \\
YOLO-NAS-l \cite{supergradients}        & 0.470 & 0.805       & 0.488       & 0.285  & 0.453  & 0.620  & 66.9          & 16.01                  & 62.5     \\
Faster R-CNN \cite{Fasterrcnn}      & 0.379 & 0.690       & 0.374       & 0.192  & 0.348  & 0.538  & 41.0          & 34.8                   & 28.7     \\
Cascade R-CNN \cite{Cascade}      & 0.449 & 0.758       & 0.470       & 0.260  & 0.422  & 0.607  & 69.1             & 40.5                   & 24.7     \\
Sparse R-CNN \cite{Sparsercnn}      & 0.357 & 0.615       & 0.375       & 0.127  & 0.305  & 0.565  & 62             & 50.4                   & 19.9     \\
DETR \cite{detr}            & 0.390 & 0.708       & 0.390       & 0.154  & 0.335  & 0.600  & 41.0          & 36.0                   & 27.8     \\
Deformable-DETR \cite{Deformable}   & 0.317 & 0.648       & 0.269       & 0.12   & 0.24   & 0.507  & 34.0          & 45                     & 22.1     \\
RT-DETR-l \cite{RTDETR}        & 0.381 & 0.696       & 0.383       & 0.156  & 0.354  & 0.588  & 32            & 15.1                   & 66       \\
RetinaNet \cite{Retinanet} & 0.448 & 0.764       & 0.465       & 0.238  & 0.416  & 0.627  & 34.0          & 26.6                   & 37.6     \\
MobileNetv2-SSD \cite{SSD}& 0.288 & 0.519       & 0.294       & 0.032  & 0.185  & 0.554  & 3             & 40                     & 25       \\
FCOS-r50-FPN \cite{Fcos}      & 0.475 & 0.838       & 0.472       & 0.203  & 0.383  & 0.599  & 32.1           & 29.7                   & 33.7     \\

FishDet-M-CLIP & 0.444 & 0.742 & 0.472 & 0.235 & 0.415 & 0.620 & -- & 12.46 & 80.28 \\
\bottomrule
\end{tabular}
\label{tab:detection-test}
\end{table*}

Region proposal models such as Cascade R-CNN (0.449 mAP) offered solid accuracy but lower speed. Single-stage alternatives like RetinaNet (0.448 mAP) and FCOS (0.475 mAP) provided better efficiency at 30–38 FPS.

\subsubsection{Precision-Recall Analysis}
The Precision-Recall curves in Fig.~\ref{fig:PR_F1_combined:a} show most models begin with high precision and drop as recall increases. MobileNetV2-SSD declines steeply at recall 0.4, and other models such as Sparse R-CNN and Deformable-DETR also exhibit earlier degradation. At IoU=0.75 Fig.~\ref{fig:PR_F1_combined:b}, the precision starts lower and declines earlier, particularly for transformer models.
\subsubsection{F1 Score}

\begin{figure}
\centering
\includegraphics[width=\linewidth]{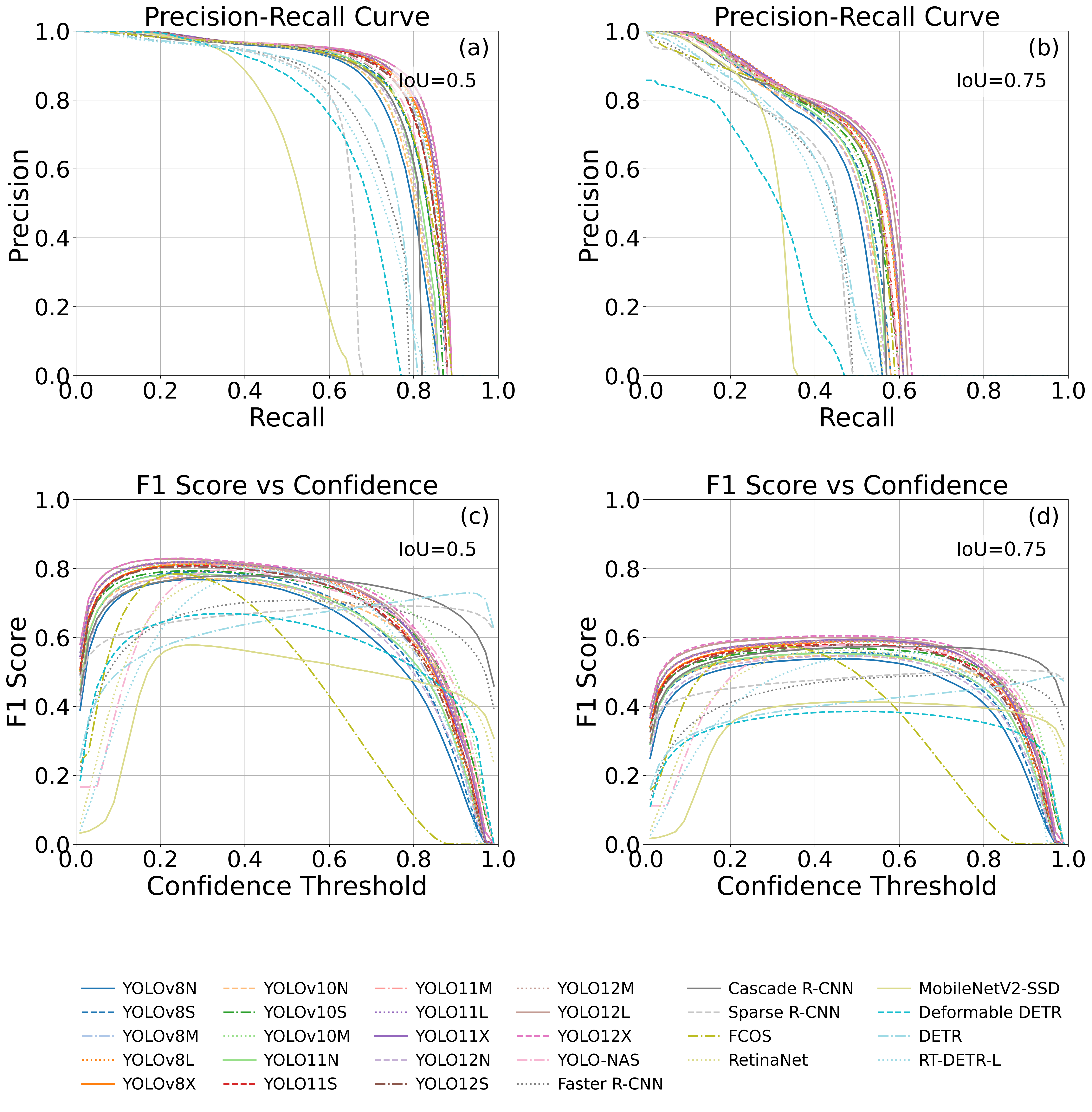}

\begin{subfigure}{0.23\linewidth}
    \phantomsubcaption
    \label{fig:PR_F1_combined:a}
\end{subfigure}
\begin{subfigure}{0.23\linewidth}
    \phantomsubcaption
    \label{fig:PR_F1_combined:b}
\end{subfigure}
\begin{subfigure}{0.23\linewidth}
    \phantomsubcaption
    \label{fig:PR_F1_combined:c}
\end{subfigure}
\begin{subfigure}{0.23\linewidth}
    \phantomsubcaption
    \label{fig:PR_F1_combined:d}
\end{subfigure}

\caption{F1 score and Precision-Recall curves for 28 object detection models at IoU thresholds of 0.50 and 0.75. Subfigures (a) and (b) show F1 score trends, and (c) and (d) depict precision-recall curves.}
\label{fig:PR_F1_combined}
\end{figure}

F1 score trends for IoU thresholds of 0.50 and 0.75 are shown in Fig.~\ref{fig:PR_F1_combined:c} and~\ref{fig:PR_F1_combined:d} . YOLO variants peaked early and declined smoothly, while Sparse R-CNN and DETR peaked later. Deformable-DETR displayed a flatter curve with early decline. At stricter IoU thresholds, models showed more uniform patterns, with FCOS and MobileNetV2-SSD dropping sharply beyond 0.6 confidence.
\subsubsection{Evaluation on Sub-Test Splits}

To better understand the contribution of each constituent dataset, we conducted a detailed evaluation of model performance on the individual test partitions of the thirteen source datasets. Results indicate notable variability depending on dataset complexity. For example, YOLO12x achieved an mAP of 0.846 on DeepFish \cite{Saleh2020}, which features underwater scenes with large prominent fish instances. In contrast, the same model scored 0.371 on FishDataset \cite{Liu2023MultitaskModel}, where overlapping and occluded, similar-looking fish in a tank present significant visual challenges.
 Across the full benchmark, FishDet-M produced consistent performance in the range of 0.481 to 0.491 for the highest-performing models (see Table~\ref{tab:detection-test}), demonstrating generalization across diverse visual domains.

The individual datasets can be broadly grouped into three performance tiers based on their environmental characteristics. Datasets in the high-performance tier, such as DeepFish \cite{Saleh2020} and FishVideo \cite{fish-video_dataset}, typically include large fish with minimal background interference and stable lighting conditions. The moderate tier includes datasets like FishNet \cite{10377207} and Fish4Knowledge \cite{fish4knowledge-dataset_dataset}, which involve variable fish sizes, mixed lighting conditions, and moderate occlusion. The most challenging tier comprises datasets such as TrashCan 1.0 \cite{hong2020trashcan} and Brackish-MOT \cite{brackishMOT}, where visual noise, low image quality, high turbidity, and frequent occlusion significantly hinder detection accuracy.
Detailed metrics for all evaluated models on each dataset are provided in the supplementary material (Table SI through SXIII ). These results confirm that while FishDet-M does not achieve the highest accuracy on any single dataset, it consistently promotes generalizable feature learning across domains. This is further evidenced by the narrow variance in model performance across subsets. An average precision value of --1 indicates that no qualifying objects were present in the corresponding sub-test split based on size thresholds.

\subsection{Qualitative and Generalization Analysis}

\begin{figure*}[t]
    \centering
    \begin{subfigure}[t]{\textwidth}
        \centering
        \setlength{\tabcolsep}{0pt}
        \begin{tabular}{@{}c@{}c@{}c@{}c@{}c@{}c@{}c@{}c@{}c@{}c@{}c@{}c@{}c@{}c@{}c@{}}
                    \textbf{\tiny GT} &
        \textbf{\tiny Y8n} &
        \textbf{\tiny Y8s} &
        \textbf{\tiny Y8m} &
        \textbf{\tiny Y8l} &
        \textbf{\tiny Y8x} &
        \textbf{\tiny Y10n} &
        \textbf{\tiny Y10s} &
        \textbf{\tiny Y10m} &
        \textbf{\tiny Y11n} &
        \textbf{\tiny Y11s} &
        \textbf{\tiny Y11m} &
        \textbf{\tiny Y11l} &
        \textbf{\tiny Y11x} &
        \textbf{\tiny Y12n} \\

        \includegraphics[width=1.1cm, height=1.1cm]{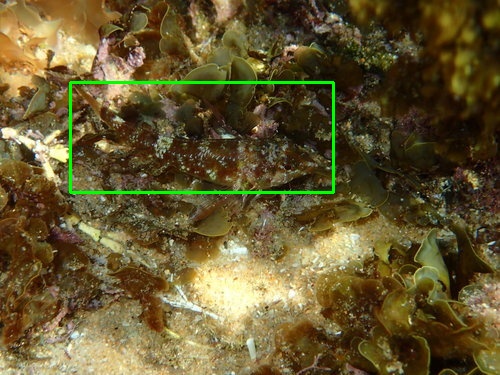} &
        \includegraphics[width=1.1cm, height=1.1cm]{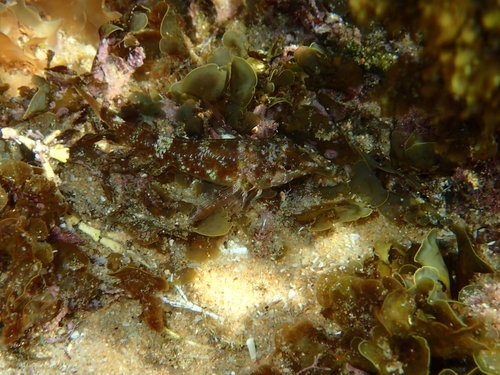} &
        \includegraphics[width=1.1cm, height=1.1cm]{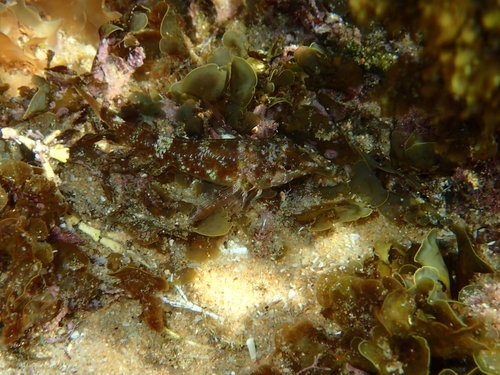} &
        \includegraphics[width=1.1cm, height=1.1cm]{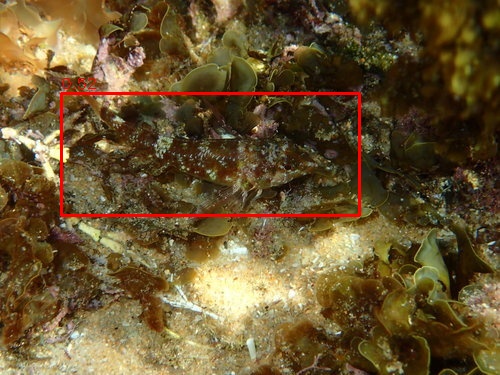} &
        \includegraphics[width=1.1cm, height=1.1cm]{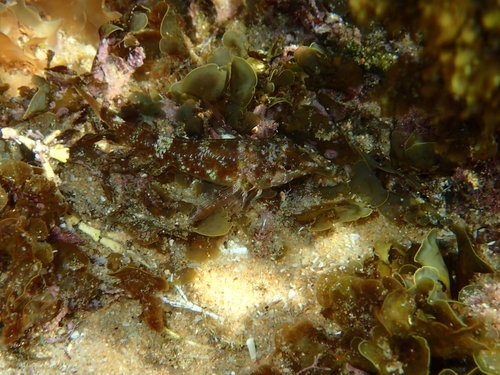} &
        \includegraphics[width=1.1cm, height=1.1cm]{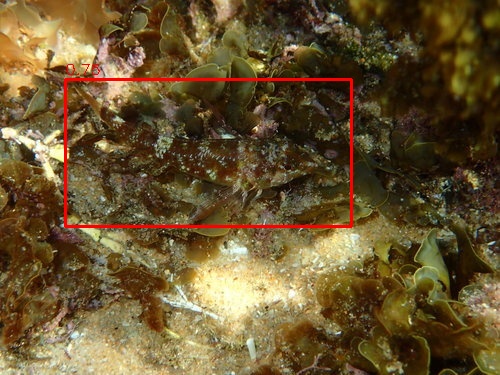} &
        \includegraphics[width=1.1cm, height=1.1cm]{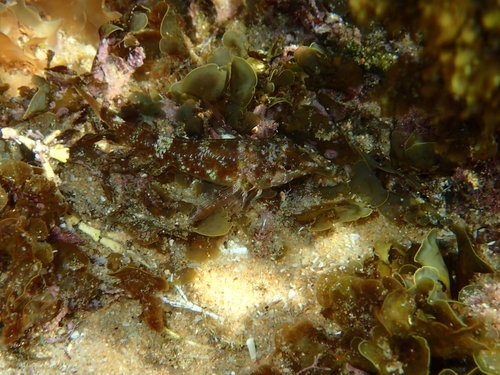} &
        \includegraphics[width=1.1cm, height=1.1cm]{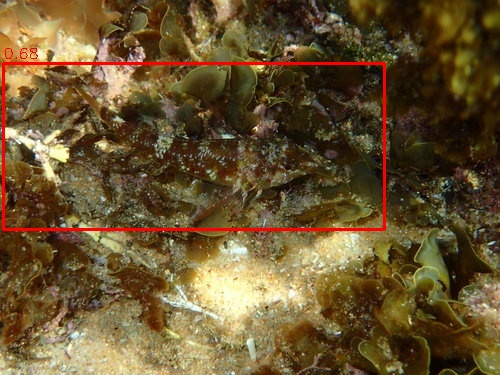} &
        \includegraphics[width=1.1cm, height=1.1cm]{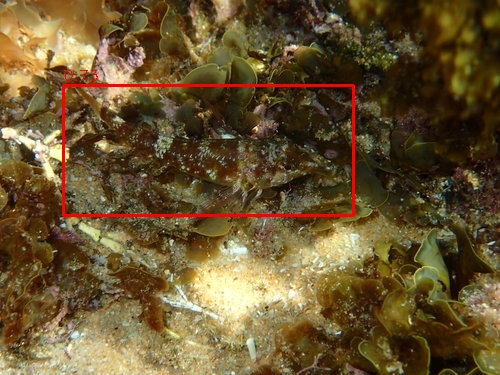} &
        \includegraphics[width=1.1cm, height=1.1cm]{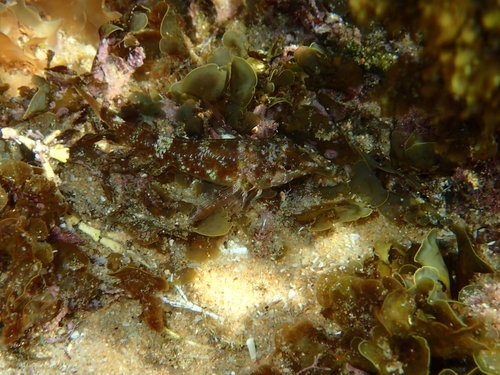} &
        \includegraphics[width=1.1cm, height=1.1cm]{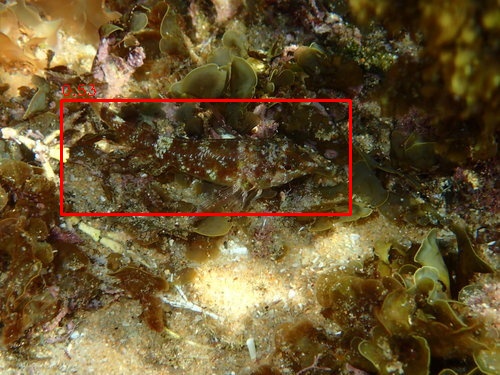} &
        \includegraphics[width=1.1cm, height=1.1cm]{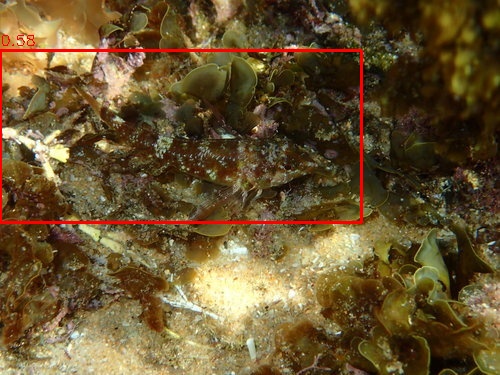} &
        \includegraphics[width=1.1cm, height=1.1cm]{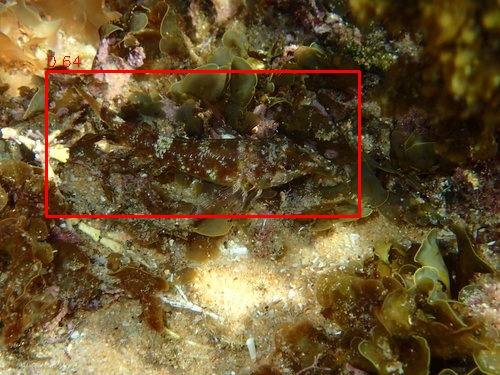} &
        \includegraphics[width=1.1cm, height=1.1cm]{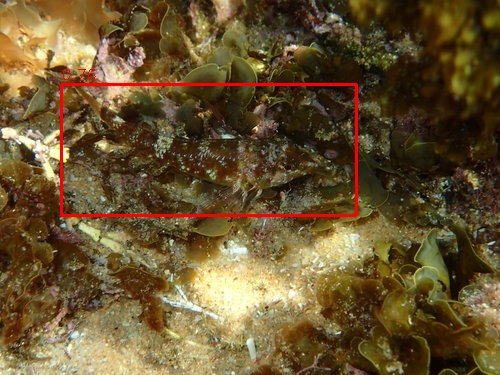} &
        \includegraphics[width=1.1cm, height=1.1cm]{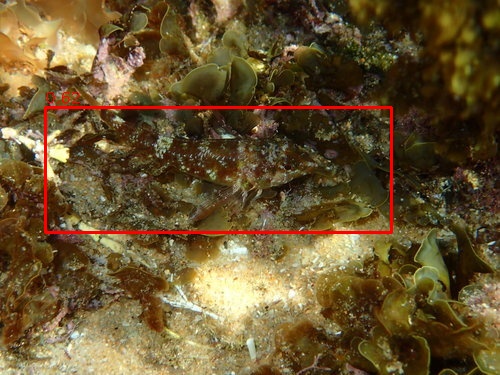} \\
        \includegraphics[width=1.1cm, height=1.1cm]{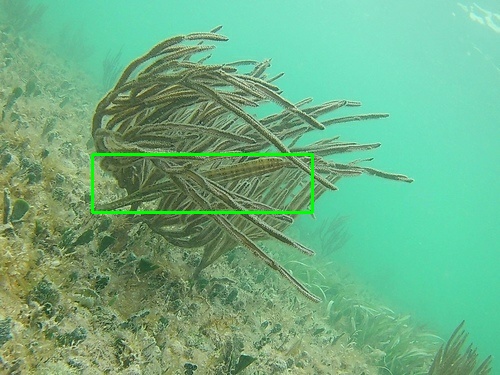} &
        \includegraphics[width=1.1cm, height=1.1cm]{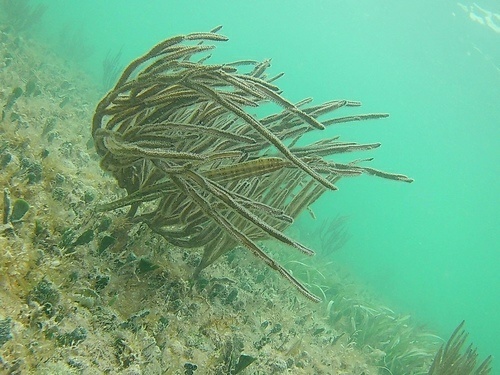} &
        \includegraphics[width=1.1cm, height=1.1cm]{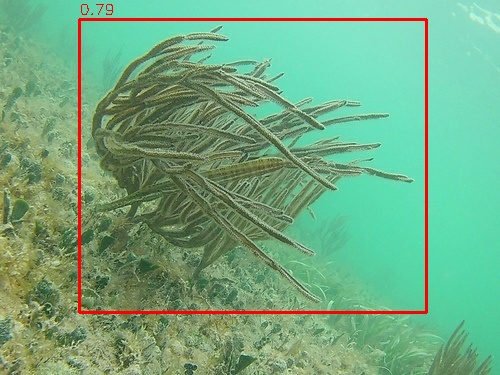} &
        \includegraphics[width=1.1cm, height=1.1cm]{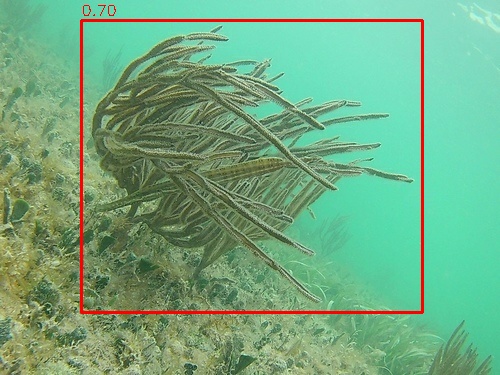} &
        \includegraphics[width=1.1cm, height=1.1cm]{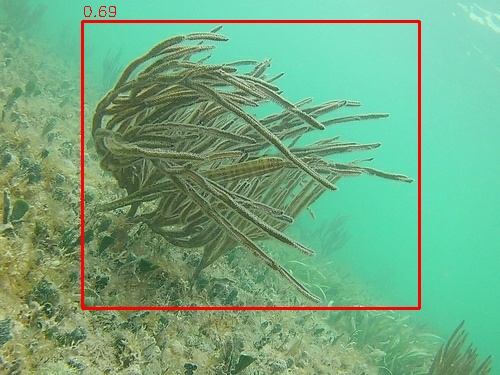} &
        \includegraphics[width=1.1cm, height=1.1cm]{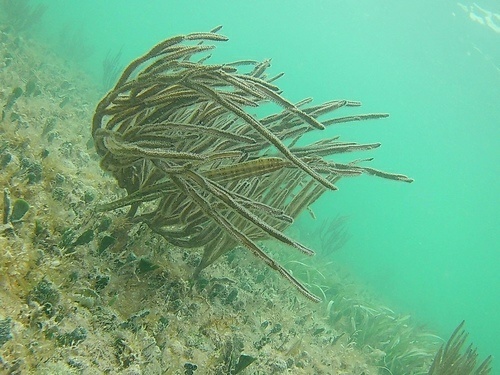} &
        \includegraphics[width=1.1cm, height=1.1cm]{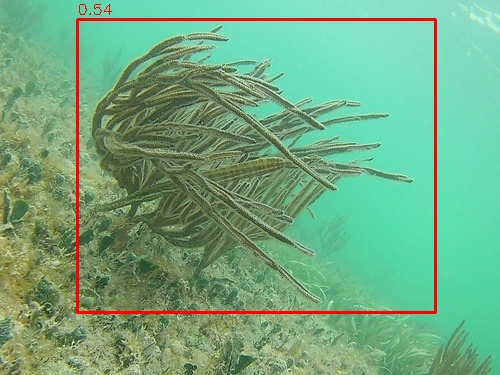} &
        \includegraphics[width=1.1cm, height=1.1cm]{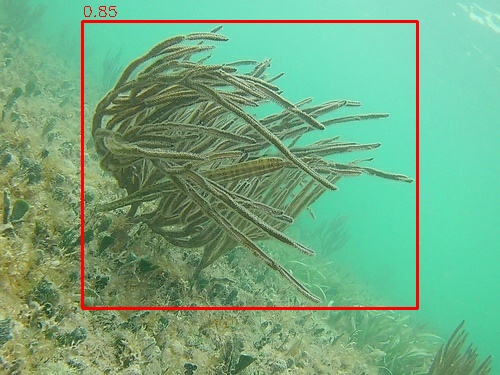} &
        \includegraphics[width=1.1cm, height=1.1cm]{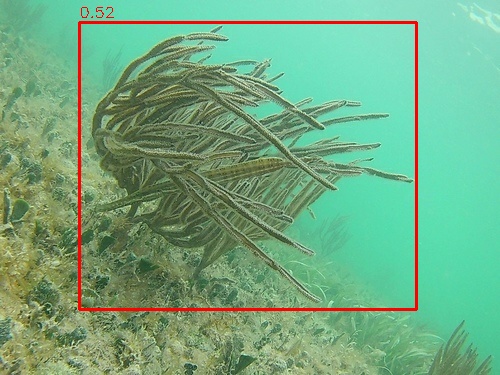} &
        \includegraphics[width=1.1cm, height=1.1cm]{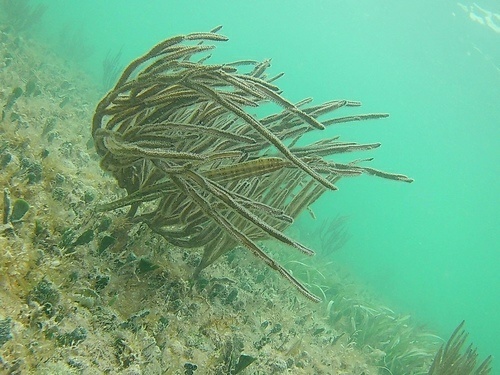} &
        \includegraphics[width=1.1cm, height=1.1cm]{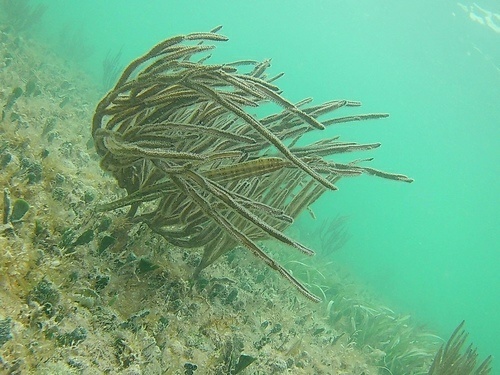} &
        \includegraphics[width=1.1cm, height=1.1cm]{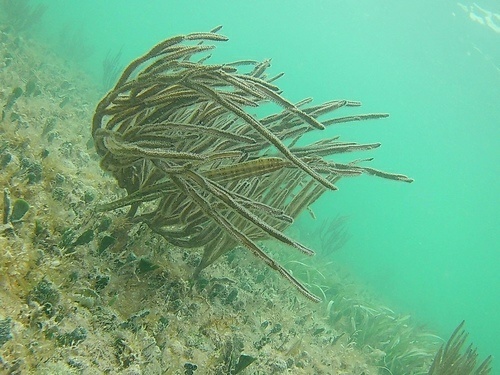} &
        \includegraphics[width=1.1cm, height=1.1cm]{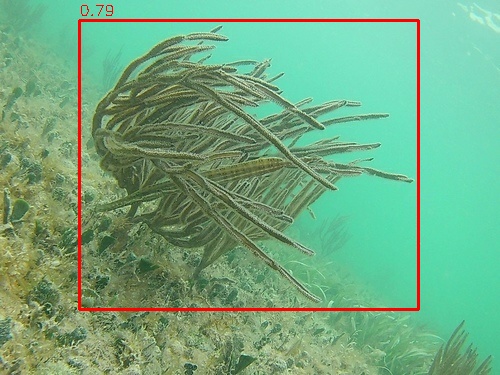} &
        \includegraphics[width=1.1cm, height=1.1cm]{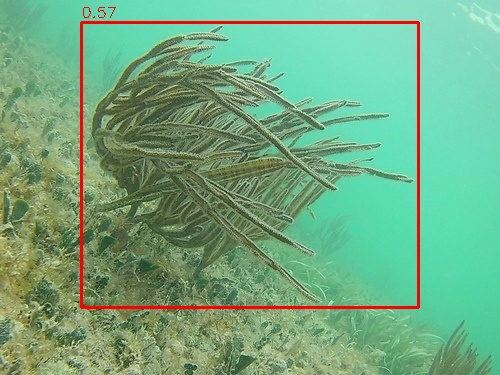} &
        \includegraphics[width=1.1cm, height=1.1cm]{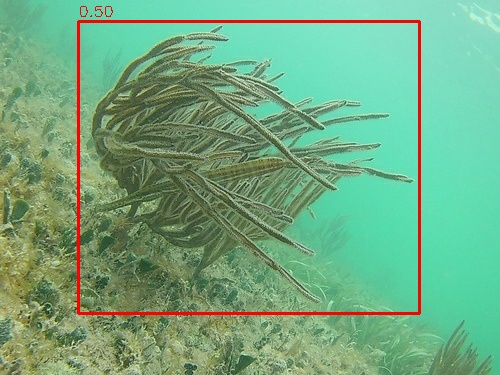} \\
        \includegraphics[width=1.1cm, height=1.1cm]{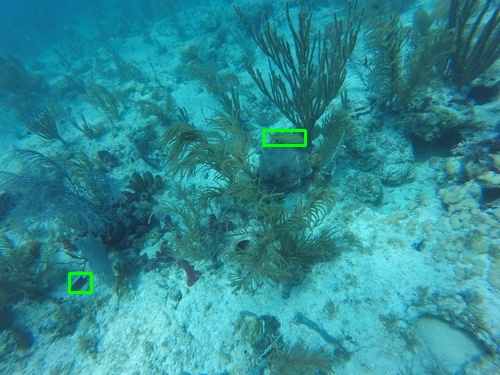} &
        \includegraphics[width=1.1cm, height=1.1cm]{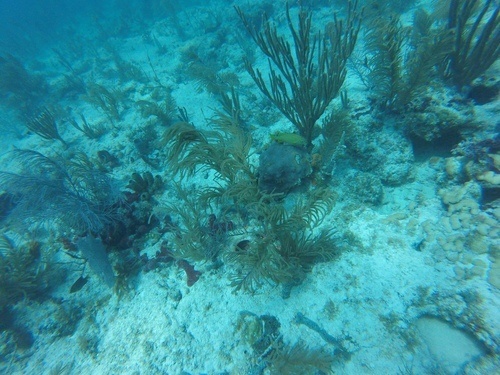} &
        \includegraphics[width=1.1cm, height=1.1cm]{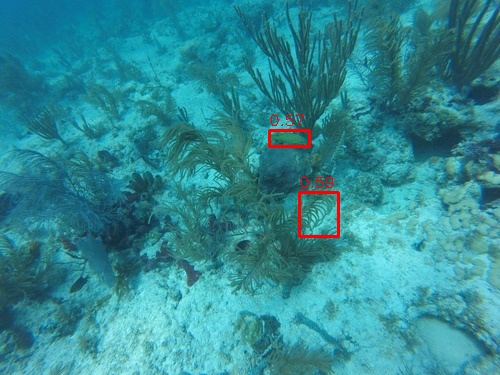} &
        \includegraphics[width=1.1cm, height=1.1cm]{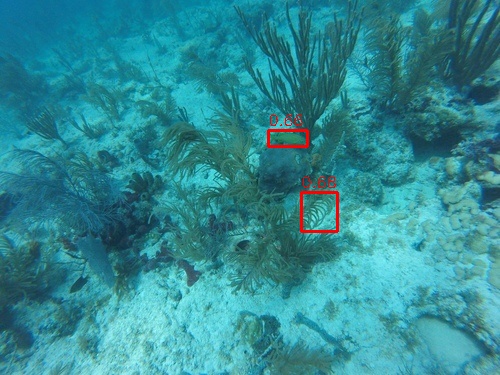} &
        \includegraphics[width=1.1cm, height=1.1cm]{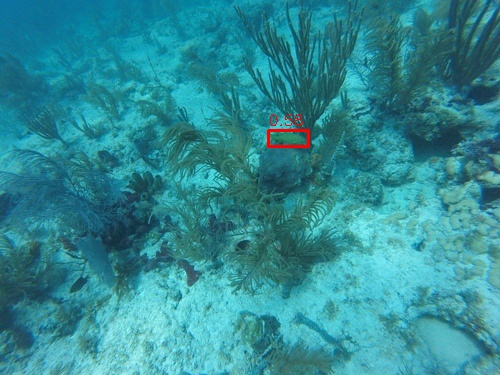} &
        \includegraphics[width=1.1cm, height=1.1cm]{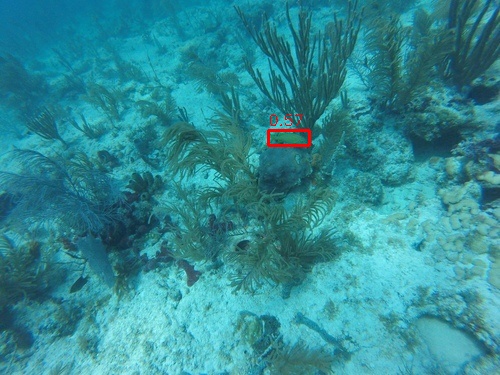} &
        \includegraphics[width=1.1cm, height=1.1cm]{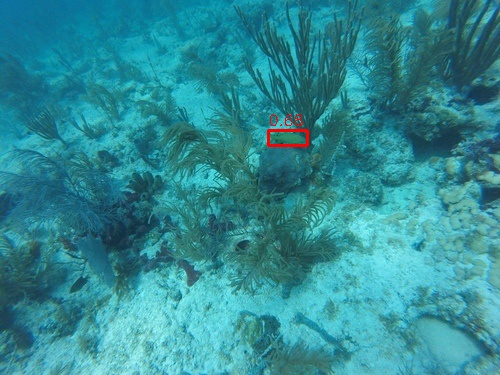} &
        \includegraphics[width=1.1cm, height=1.1cm]{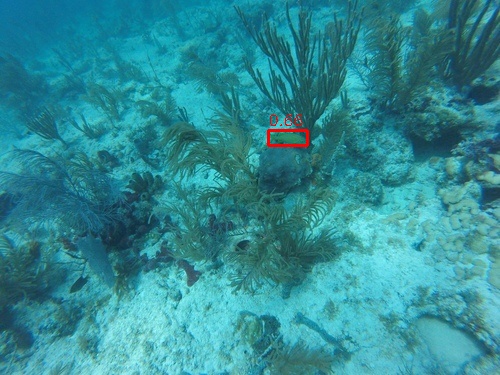} &
        \includegraphics[width=1.1cm, height=1.1cm]{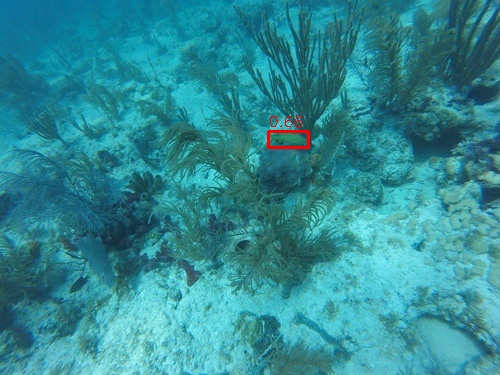} &
        \includegraphics[width=1.1cm, height=1.1cm]{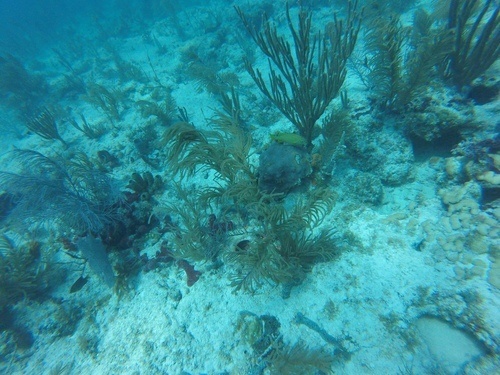} &
        \includegraphics[width=1.1cm, height=1.1cm]{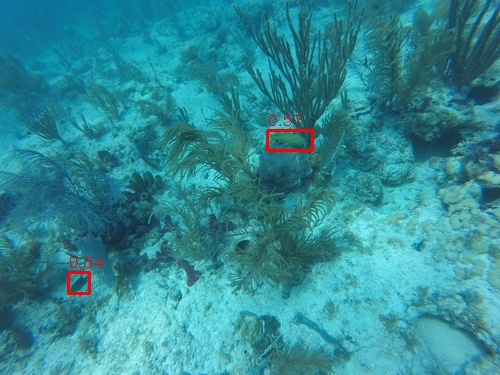} &
        \includegraphics[width=1.1cm, height=1.1cm]{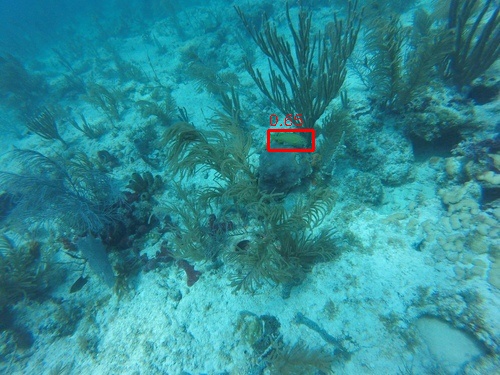} &
        \includegraphics[width=1.1cm, height=1.1cm]{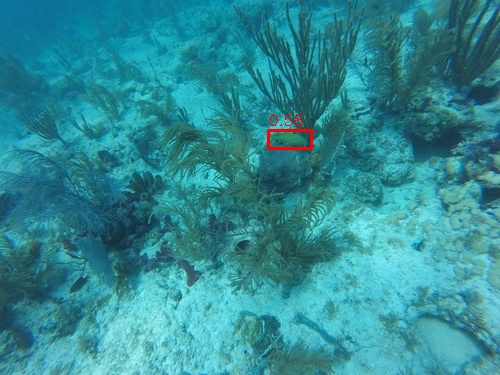} &
        \includegraphics[width=1.1cm, height=1.1cm]{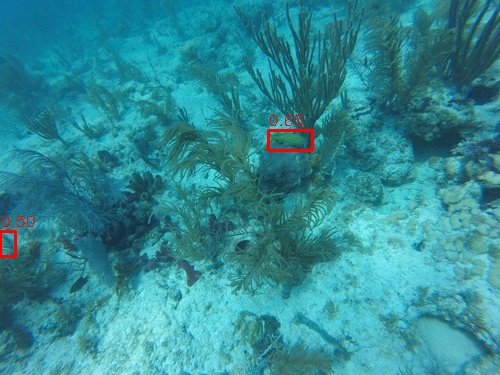} &
        \includegraphics[width=1.1cm, height=1.1cm]{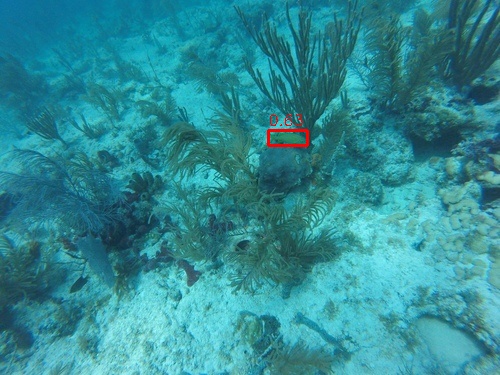} \\
        \includegraphics[width=1.1cm, height=1.1cm]{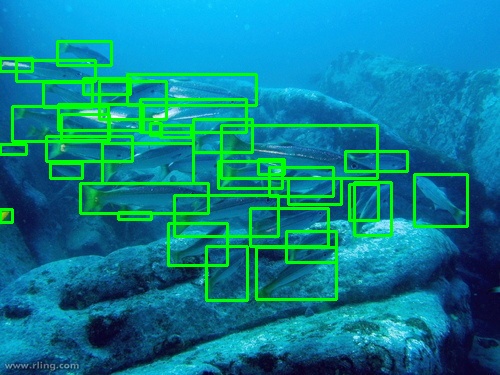} &
        \includegraphics[width=1.1cm, height=1.1cm]{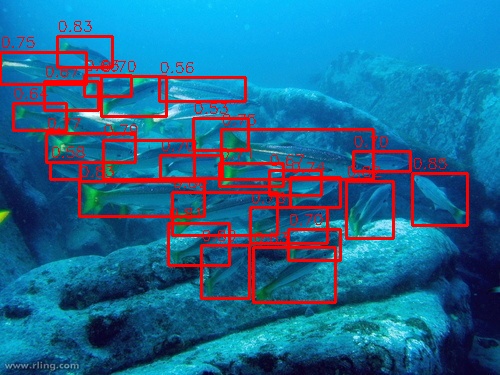} &
        \includegraphics[width=1.1cm, height=1.1cm]{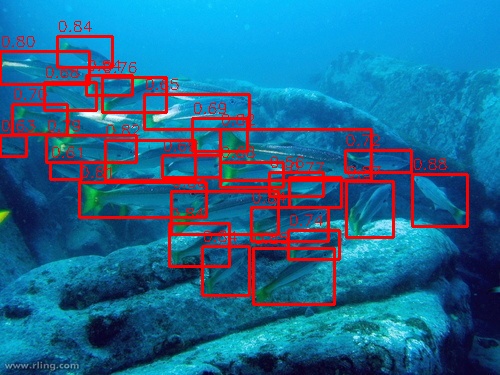} &
        \includegraphics[width=1.1cm, height=1.1cm]{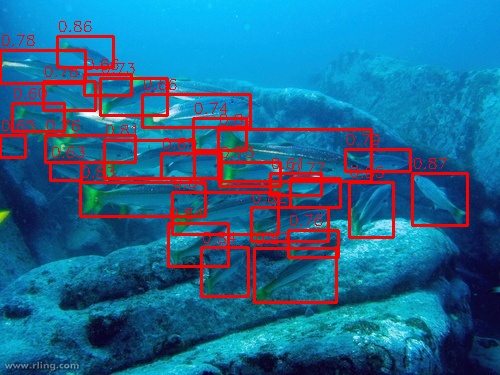} &
        \includegraphics[width=1.1cm, height=1.1cm]{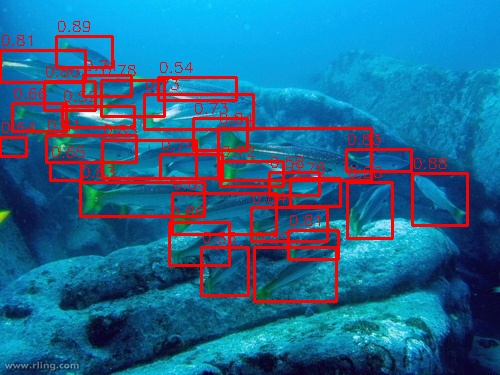} &
        \includegraphics[width=1.1cm, height=1.1cm]{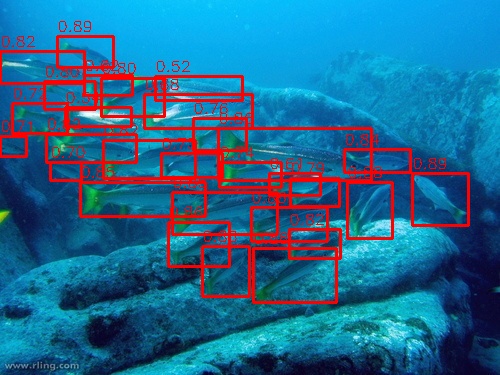} &
        \includegraphics[width=1.1cm, height=1.1cm]{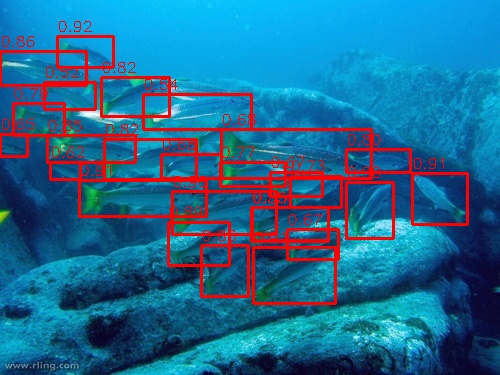} &
        \includegraphics[width=1.1cm, height=1.1cm]{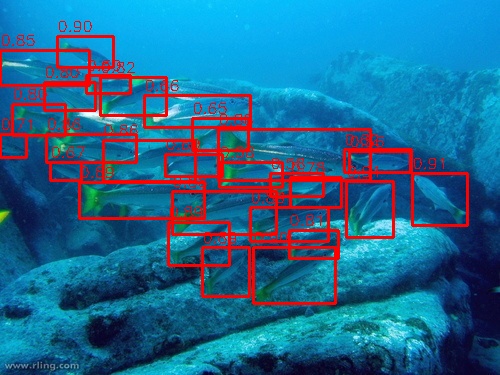} &
        \includegraphics[width=1.1cm, height=1.1cm]{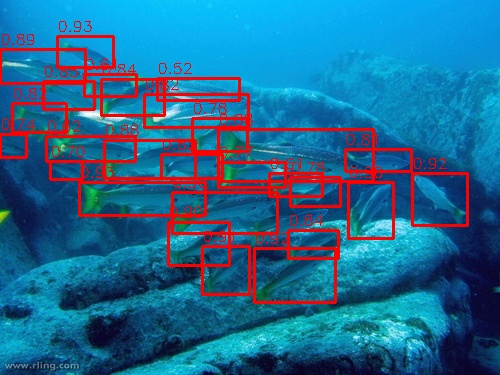} &
        \includegraphics[width=1.1cm, height=1.1cm]{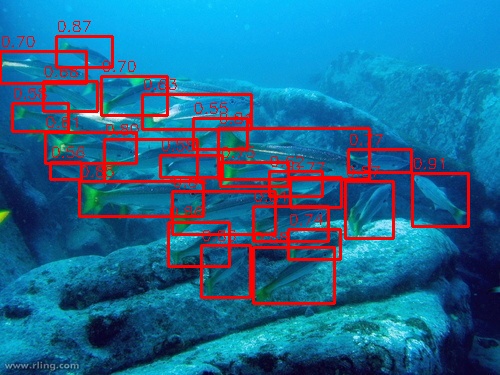} &
        \includegraphics[width=1.1cm, height=1.1cm]{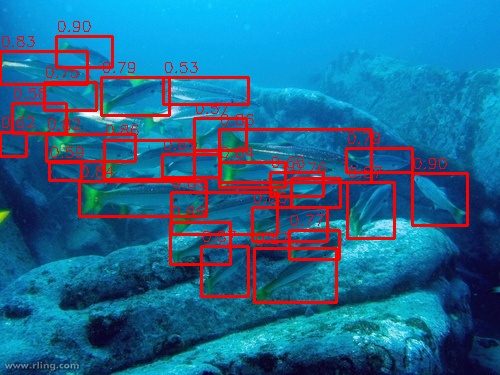} &
        \includegraphics[width=1.1cm, height=1.1cm]{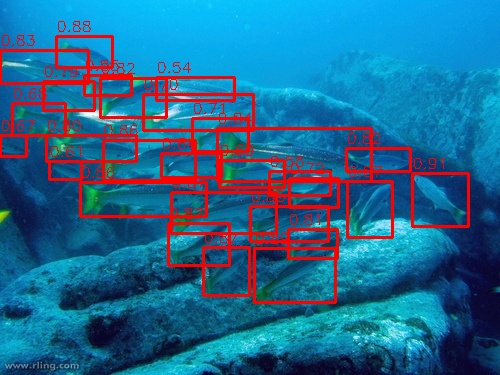} &
        \includegraphics[width=1.1cm, height=1.1cm]{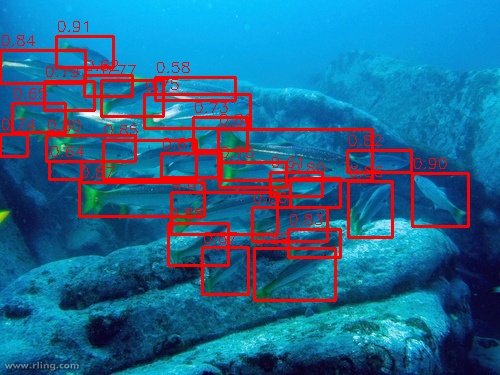} &
        \includegraphics[width=1.1cm, height=1.1cm]{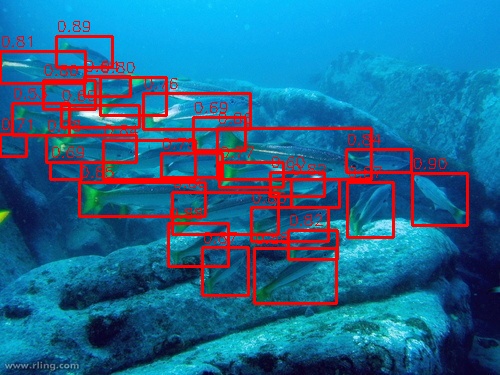} &
        \includegraphics[width=1.1cm, height=1.1cm]{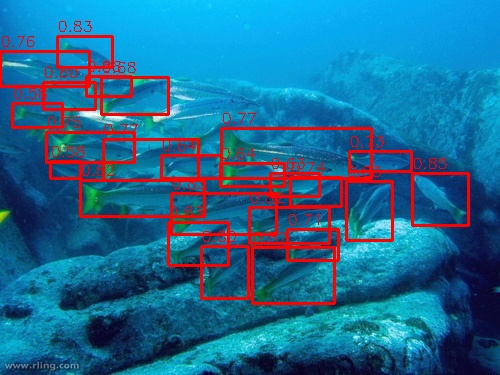} \\
        \end{tabular}
    \end{subfigure}
    \vspace{0.05em}

    \begin{subfigure}[t]{\textwidth}
        \centering
        \setlength{\tabcolsep}{0pt}
        \begin{tabular}{@{}c@{}c@{}c@{}c@{}c@{}c@{}c@{}c@{}c@{}c@{}c@{}c@{}c@{}c@{}c@{}}
\textbf{\tiny GT} &
\textbf{\tiny Y12s} &
\textbf{\tiny Y12m} &
\textbf{\tiny Y12l} &
\textbf{\tiny Y12x} &
\textbf{\tiny YNAS} &
\textbf{\tiny DETR} &
\textbf{\tiny DefD} &
\textbf{\tiny RTD} &
\textbf{\tiny FRCNN} &
\textbf{\tiny CRCNN} &
\textbf{\tiny SRCNN} &
\textbf{\tiny RNet} &
\textbf{\tiny FCOS} &
\textbf{\tiny MobSSD} \\

        \includegraphics[width=1.1cm, height=1.1cm]{figures/pred_figure/1/GT.jpg} &
        \includegraphics[width=1.1cm, height=1.1cm]{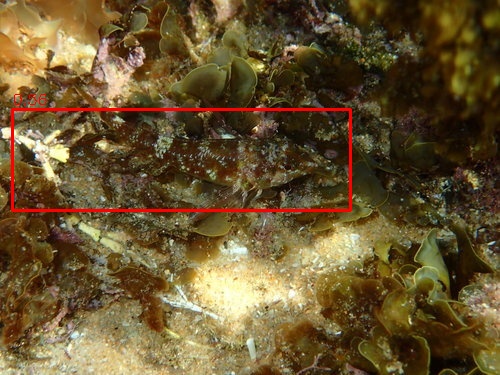} &
        \includegraphics[width=1.1cm, height=1.1cm]{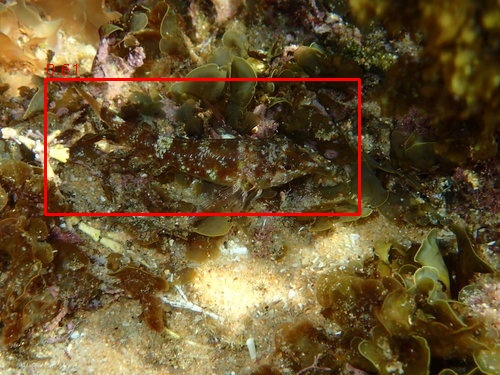} &
        \includegraphics[width=1.1cm, height=1.1cm]{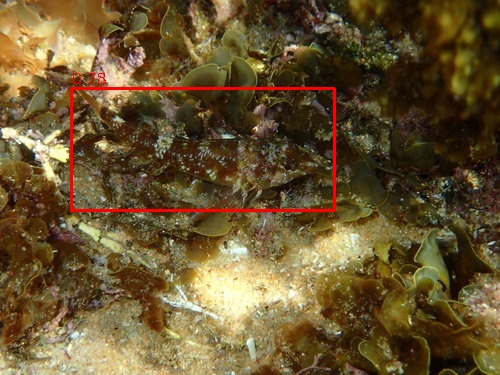} &
        \includegraphics[width=1.1cm, height=1.1cm]{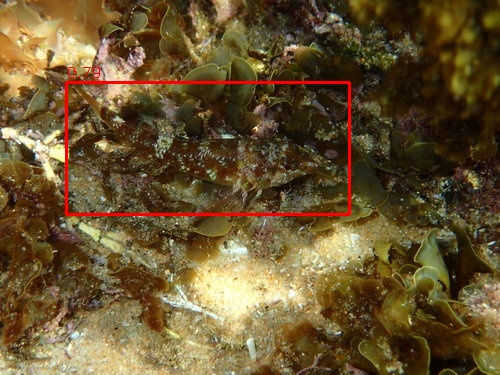} &
        \includegraphics[width=1.1cm, height=1.1cm]{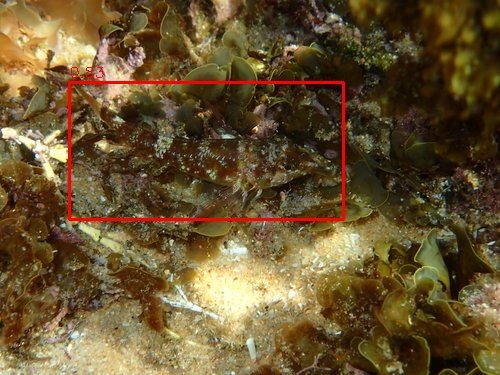} &
        \includegraphics[width=1.1cm, height=1.1cm]{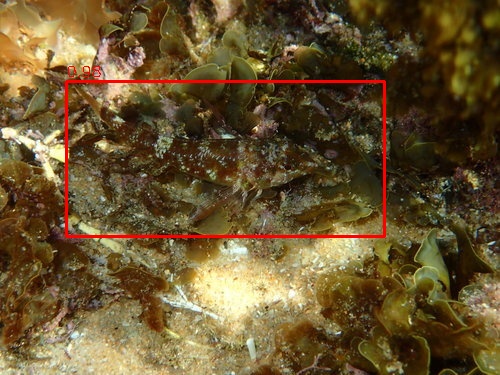} &
        \includegraphics[width=1.1cm, height=1.1cm]{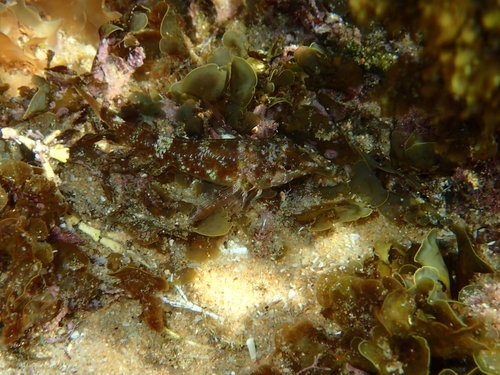} &
        \includegraphics[width=1.1cm, height=1.1cm]{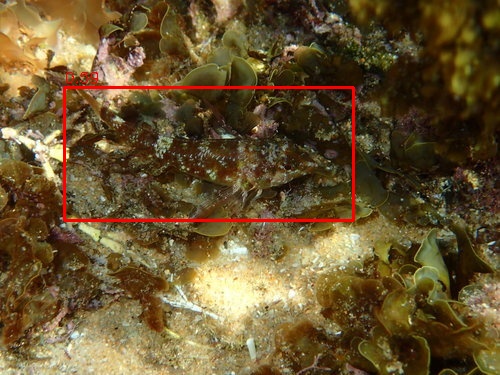} &
        \includegraphics[width=1.1cm, height=1.1cm]{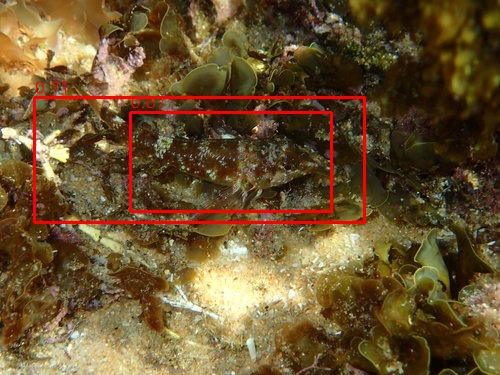} &
        \includegraphics[width=1.1cm, height=1.1cm]{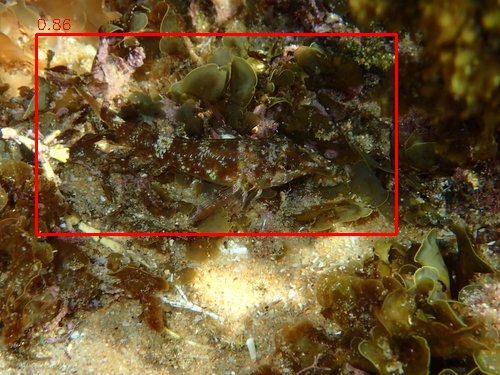} &
        \includegraphics[width=1.1cm, height=1.1cm]{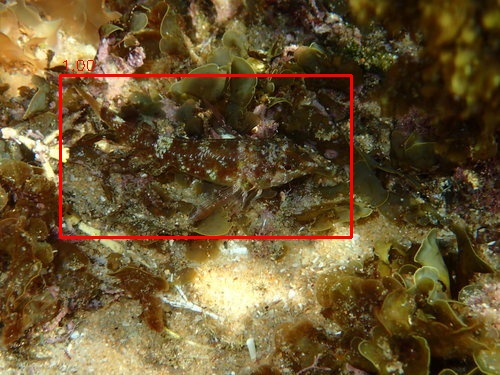} &
        \includegraphics[width=1.1cm, height=1.1cm]{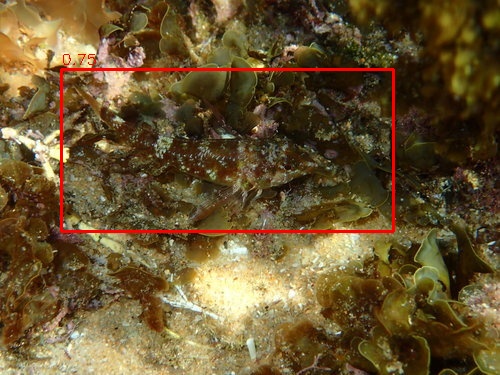} &
        \includegraphics[width=1.1cm, height=1.1cm]{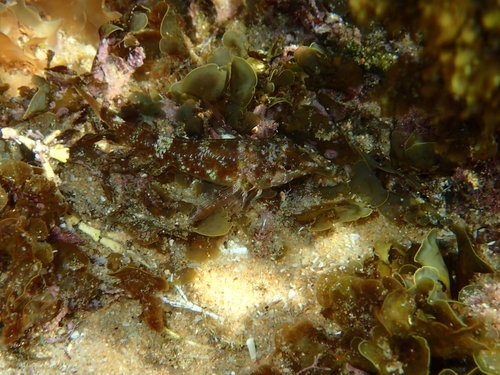} &
        \includegraphics[width=1.1cm, height=1.1cm]{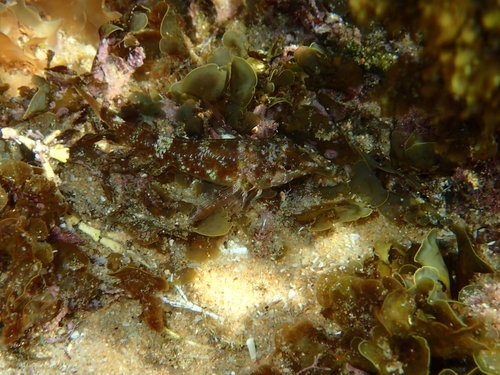} \\
        \includegraphics[width=1.1cm, height=1.1cm]{figures/pred_figure/2/GT.jpg} &
        \includegraphics[width=1.1cm, height=1.1cm]{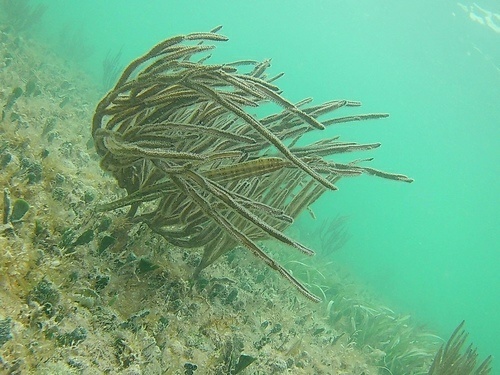} &
        \includegraphics[width=1.1cm, height=1.1cm]{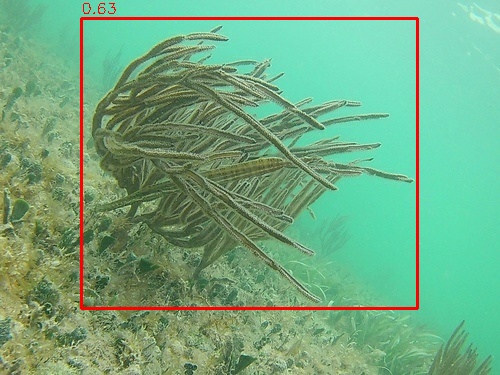} &
        \includegraphics[width=1.1cm, height=1.1cm]{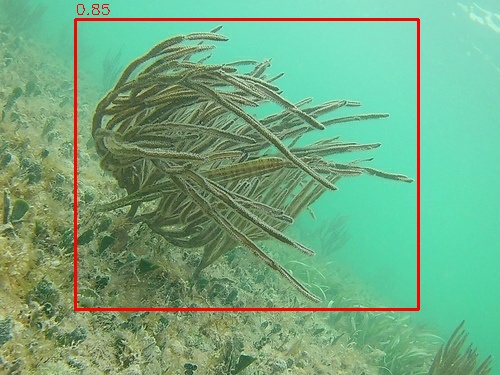} &
        \includegraphics[width=1.1cm, height=1.1cm]{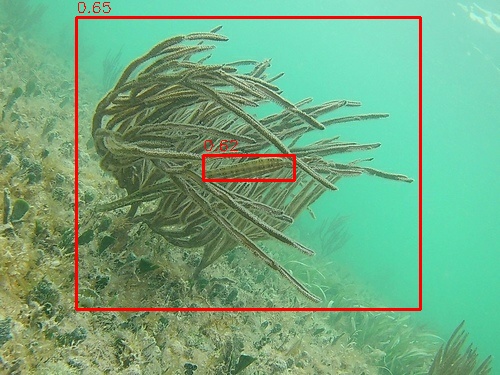} &
        \includegraphics[width=1.1cm, height=1.1cm]{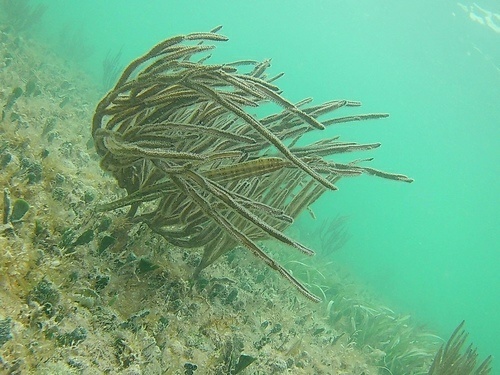} &
        \includegraphics[width=1.1cm, height=1.1cm]{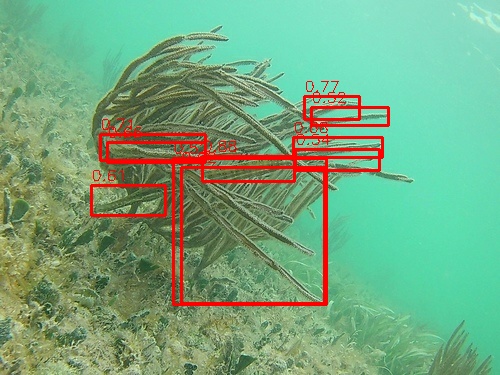} &
        \includegraphics[width=1.1cm, height=1.1cm]{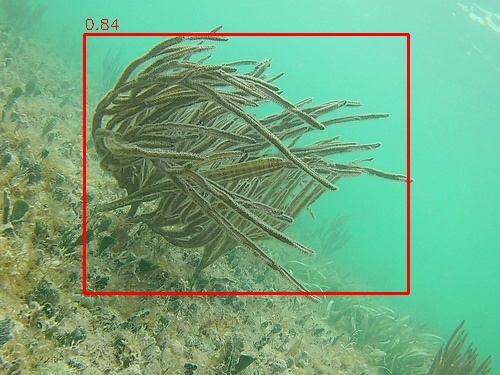} &
        \includegraphics[width=1.1cm, height=1.1cm]{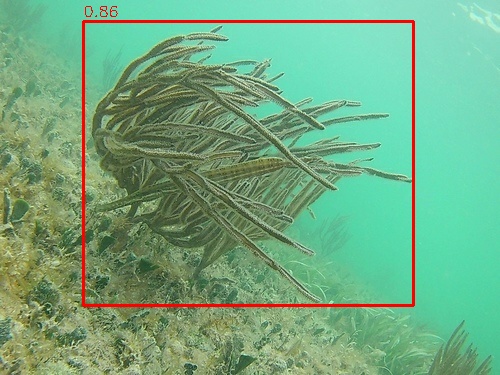} &
        \includegraphics[width=1.1cm, height=1.1cm]{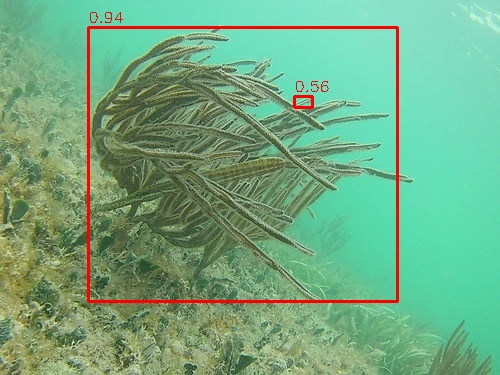} &
        \includegraphics[width=1.1cm, height=1.1cm]{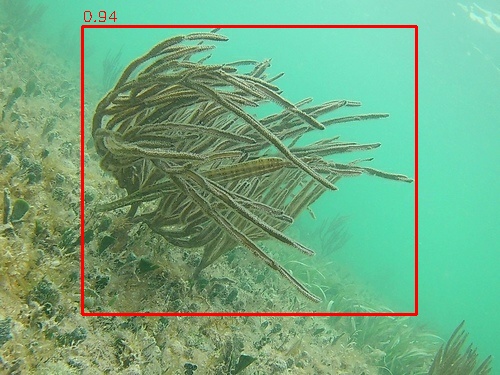} &
        \includegraphics[width=1.1cm, height=1.1cm]{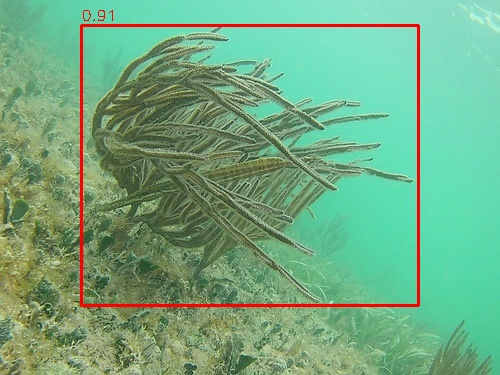} &
        \includegraphics[width=1.1cm, height=1.1cm]{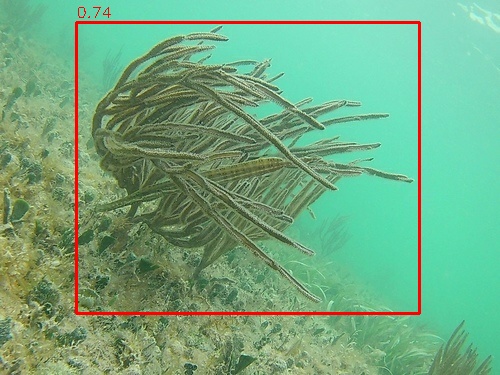} &
        \includegraphics[width=1.1cm, height=1.1cm]{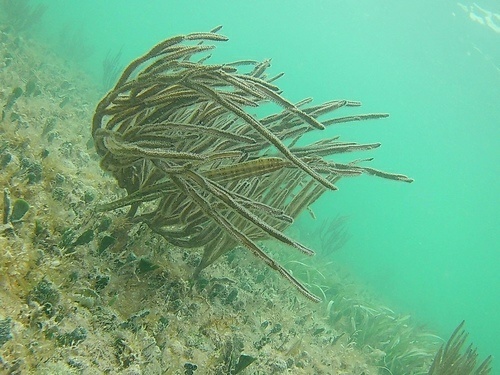} &
        \includegraphics[width=1.1cm, height=1.1cm]{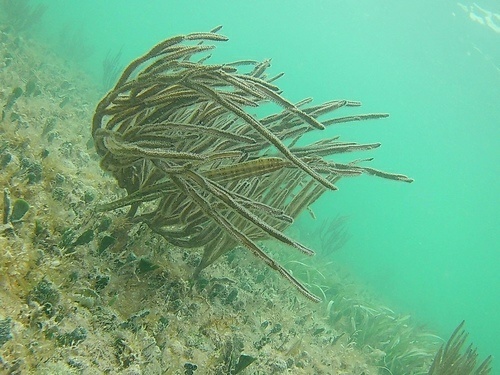} \\
        \includegraphics[width=1.1cm, height=1.1cm]{figures/pred_figure/3/GT.jpg} &
        \includegraphics[width=1.1cm, height=1.1cm]{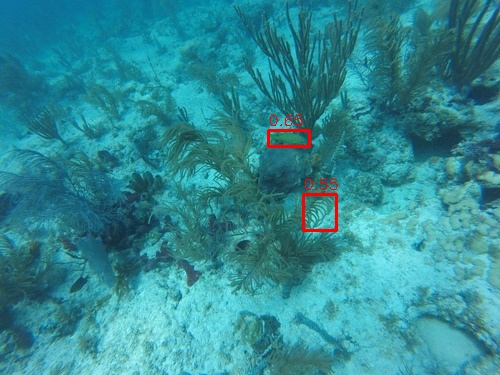} &
        \includegraphics[width=1.1cm, height=1.1cm]{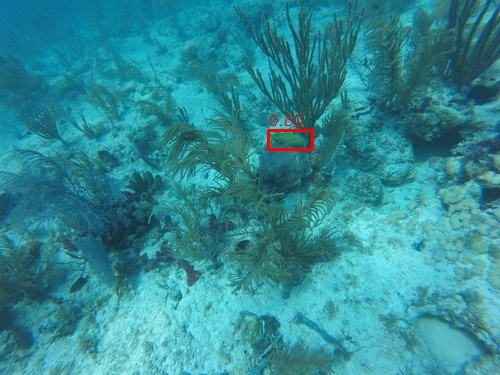} &
        \includegraphics[width=1.1cm, height=1.1cm]{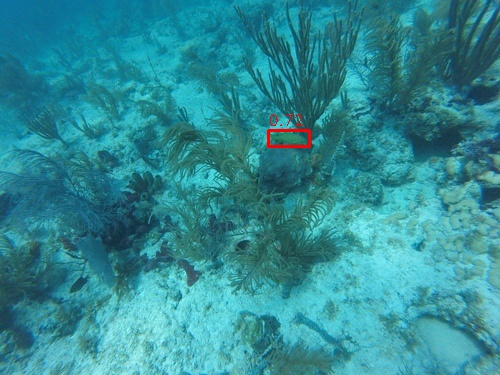} &
        \includegraphics[width=1.1cm, height=1.1cm]{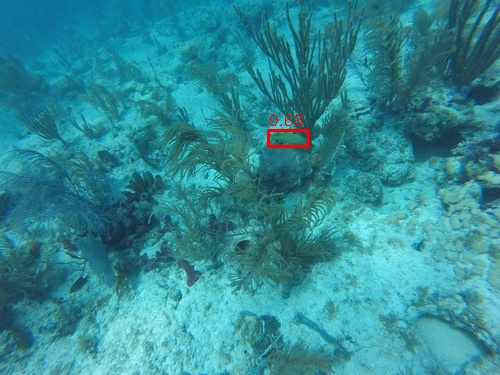} &
        \includegraphics[width=1.1cm, height=1.1cm]{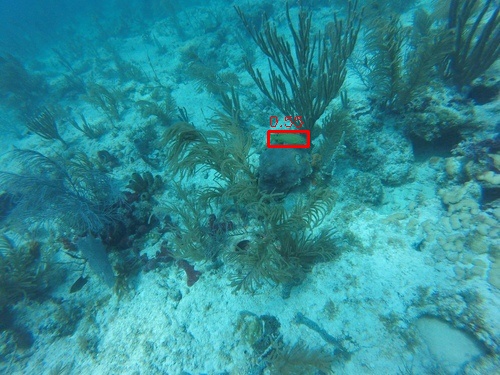} &
        \includegraphics[width=1.1cm, height=1.1cm]{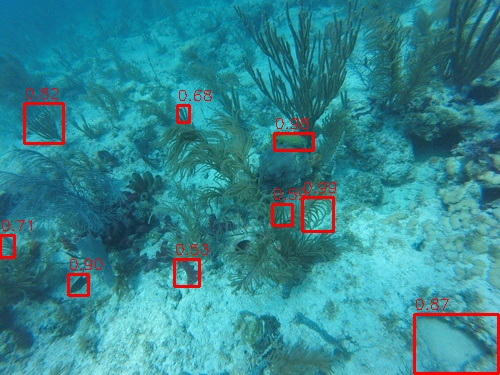} &
        \includegraphics[width=1.1cm, height=1.1cm]{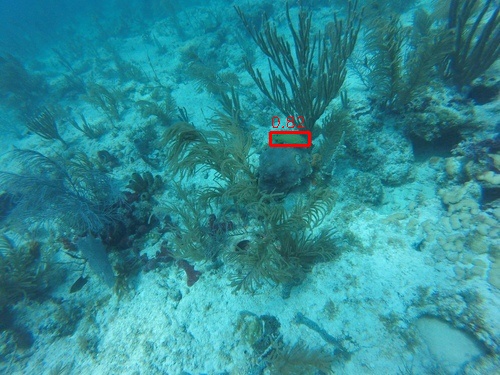} &
        \includegraphics[width=1.1cm, height=1.1cm]{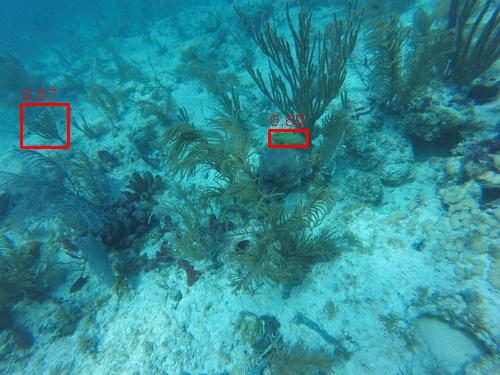} &
        \includegraphics[width=1.1cm, height=1.1cm]{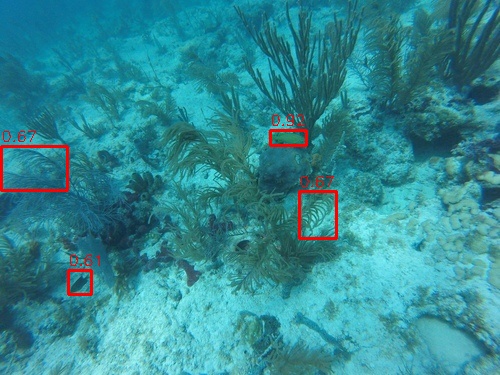} &
        \includegraphics[width=1.1cm, height=1.1cm]{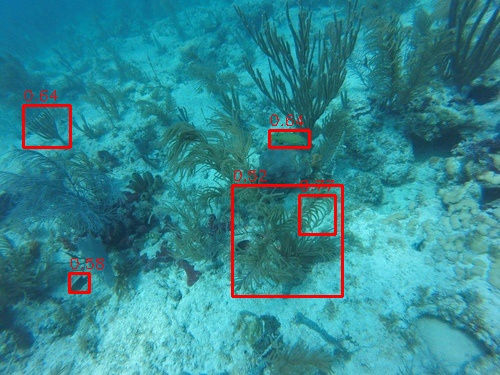} &
        \includegraphics[width=1.1cm, height=1.1cm]{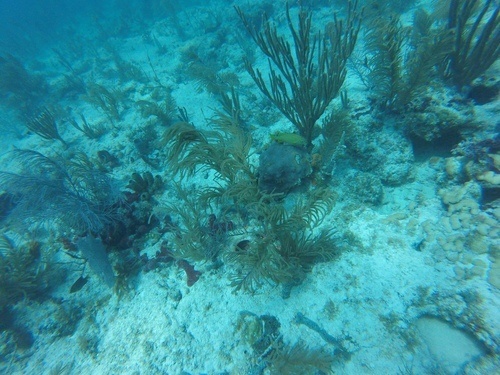} &
        \includegraphics[width=1.1cm, height=1.1cm]{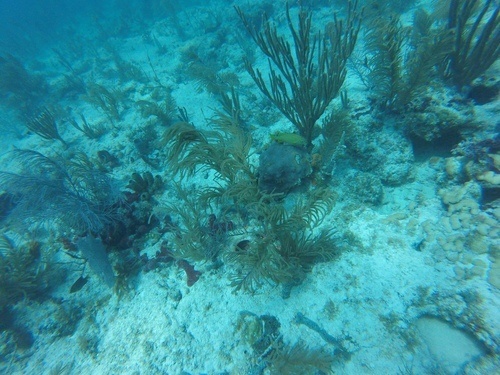} &
        \includegraphics[width=1.1cm, height=1.1cm]{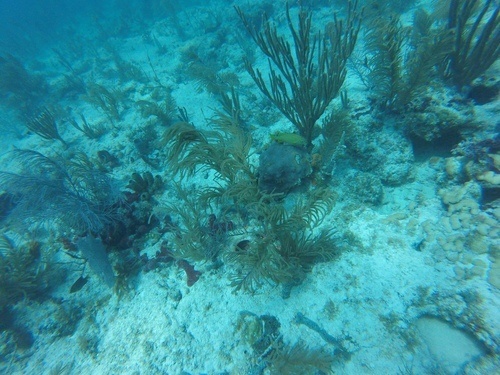} &
        \includegraphics[width=1.1cm, height=1.1cm]{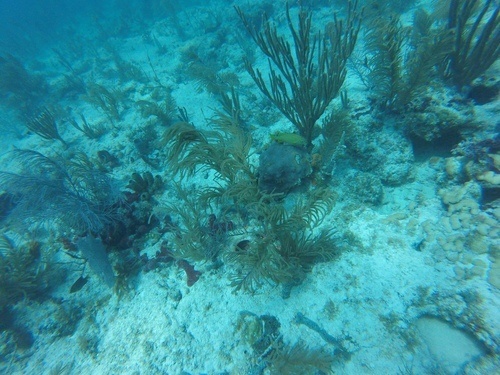} \\
        \includegraphics[width=1.1cm, height=1.1cm]{figures/pred_figure/4/GT.jpg} &
        \includegraphics[width=1.1cm, height=1.1cm]{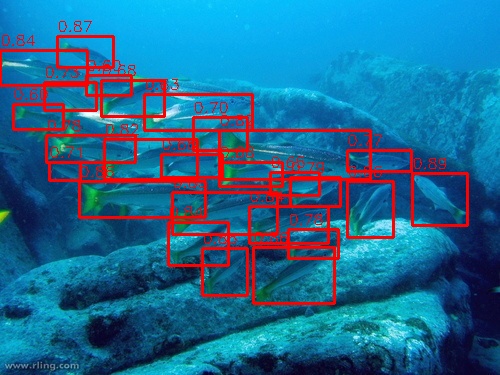} &
        \includegraphics[width=1.1cm, height=1.1cm]{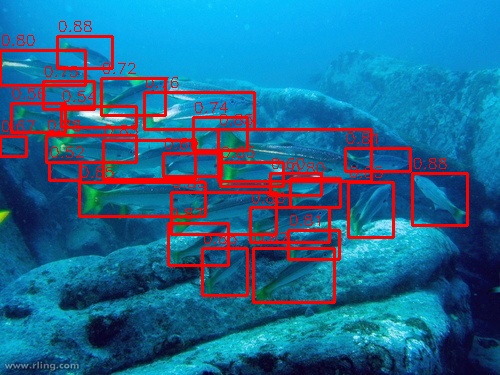} &
        \includegraphics[width=1.1cm, height=1.1cm]{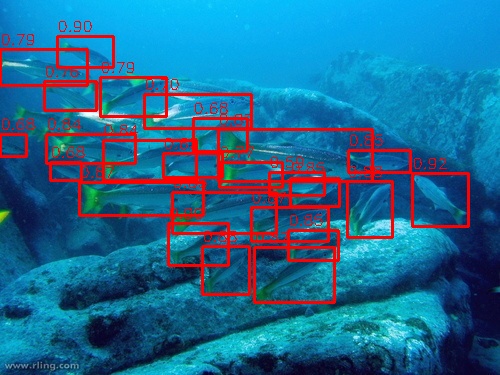} &
        \includegraphics[width=1.1cm, height=1.1cm]{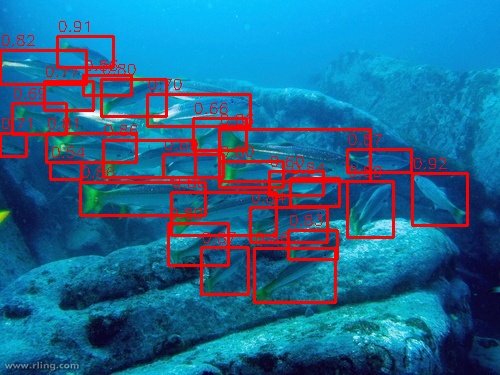} &
        \includegraphics[width=1.1cm, height=1.1cm]{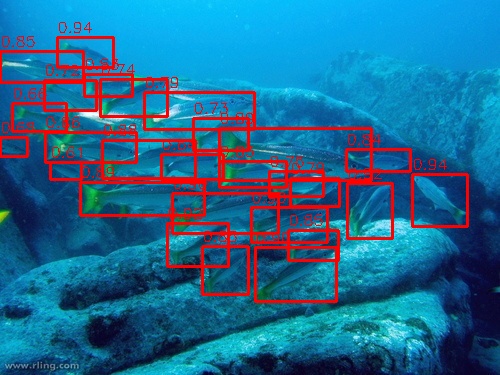} &
        \includegraphics[width=1.1cm, height=1.1cm]{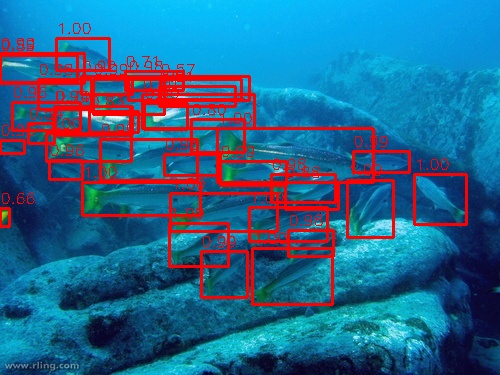} &
        \includegraphics[width=1.1cm, height=1.1cm]{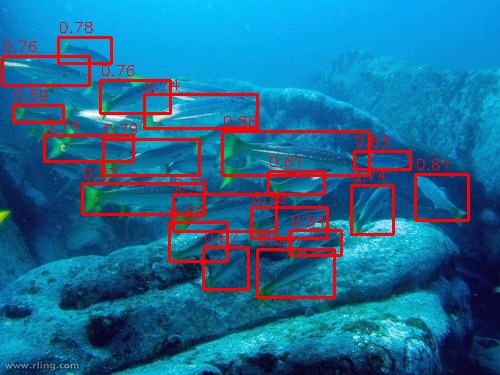} &
        \includegraphics[width=1.1cm, height=1.1cm]{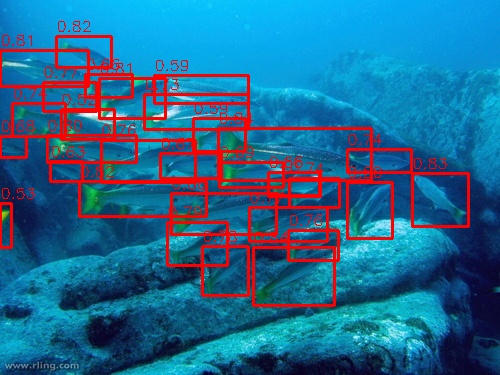} &
        \includegraphics[width=1.1cm, height=1.1cm]{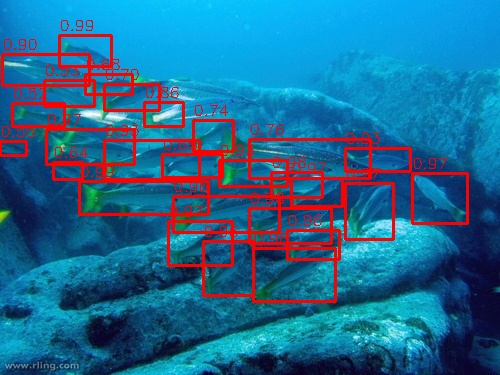} &
        \includegraphics[width=1.1cm, height=1.1cm]{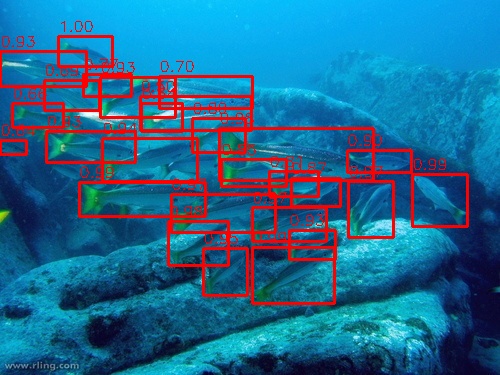} &
        \includegraphics[width=1.1cm, height=1.1cm]{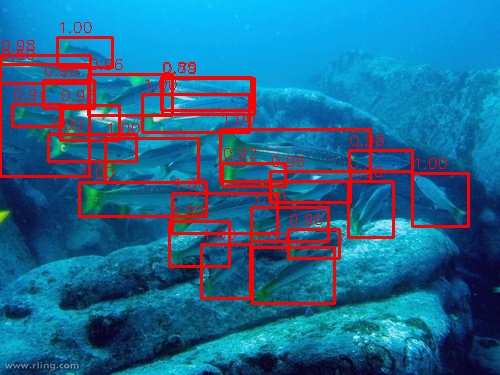} &
        \includegraphics[width=1.1cm, height=1.1cm]{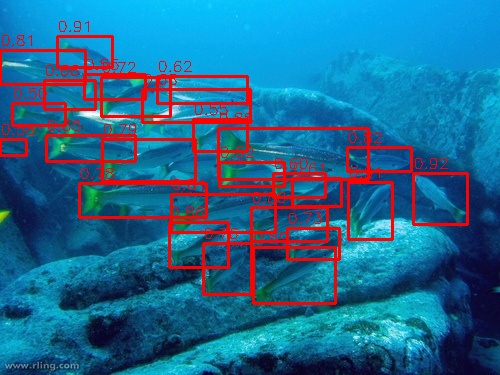} &
        \includegraphics[width=1.1cm, height=1.1cm]{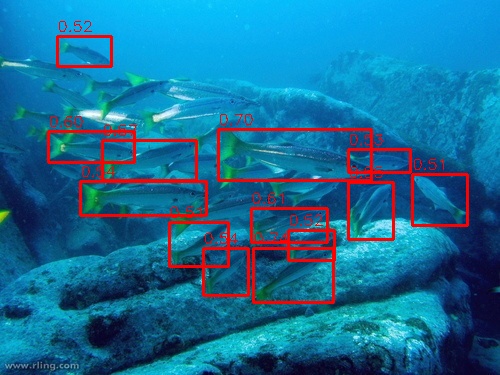} &
        \includegraphics[width=1.1cm, height=1.1cm]{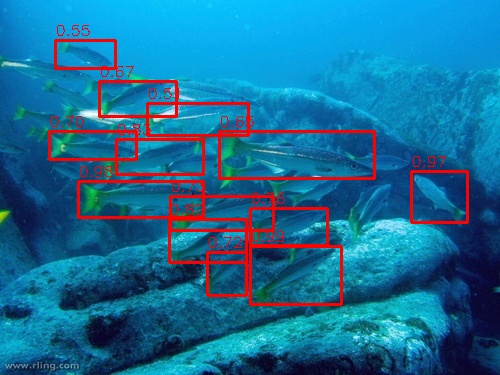} \\
        \end{tabular}
    \end{subfigure}
    \caption{Object detection results from 28 models across four challenging underwater images. Each row shows the original image followed by predictions from the models in the following order: GT (Ground Truth), Y8, Y10, Y11, Y12 (YOLOv8–v12 variants), YNAS (YOLO-NAS-L), DETR, DefD (Deformable DETR), RTD (RT-DETR), R-CNN (Faster, Cascade, and Sparse R-CNN), RNet (RetinaNet), FCOS, and MobSSD (MobileNetV2-SSD). The images represent varying detection difficulty: Image 1 shows a camouflaged fish with background-matching color and texture; Image 2 contains a partially occluded fish among visually similar corals; Image 3 includes two small fish with limited pixel footprint; and Image 4 depicts a dense school of fish with significant inter-object occlusion and low contrast against the background}
    \label{fig:model_pred}
\end{figure*}

In addition to quantitative metrics, visual inspection and cross domain testing provide further insight into model robustness under diverse underwater conditions. Fig.~\ref{fig:model_pred} presents outputs from 28 models across four representative images, selected to highlight common failure scenarios.

In camouflage conditions, where fish resemble the background, most models produce inaccurate or oversized bounding boxes due to low contrast. Occlusion by coral or overlapping structures leads to missed or fragmented detections, especially when background textures are similar to fish bodies. Small object detection remains particularly challenging, with limited pixel footprint resulting in frequent false negatives. In dense scenes with multiple fish and reduced visibility, models often merge instances or fail to localize targets precisely, revealing limitations in crowded environments.

To evaluate generalization, we tested all models on a separate dataset of 1500 unseen images from a different source \cite{Florencedataset}. As shown in Table~\ref{tab:generalization-performance}, top models from the YOLO family, including YOLO12l and YOLO11x, maintained high accuracy (mAP approximately 0.63) across object scales. Lightweight variants such as YOLOv8n and YOLOv10n also performed reliably. In contrast, transformer based detectors such as Deformable DETR showed a noticeable drop in accuracy, particularly for small objects, while region proposal models like Faster R-CNN showed moderate transferability. Single stage detectors such as RetinaNet and FCOS proved more stable under domain shift.

Together, these visual and cross dataset evaluations confirm the adaptability of the models trained on FishDet-M and highlight persistent limitations including camouflage, occlusion, small object detection, and dense scene separation. These remain open challenges for reliable fish detection in natural underwater environments.

\begin{table}[t]
\caption{Models Performance Comparison on unseen test data from a different source with sample size of 1500 \cite{Florencedataset}}
\tiny
\centering
\begin{tabular}{lcccccc}
\toprule
Model & mAP & mAP$_{50}$ & mAP$_{75}$ & AP$_S$ & AP$_M$ & AP$_L$ \\
\midrule
YOLOv8n \cite{yolov8_ultralytics} & 0.542 & 0.826 & 0.607 & 0.252 & 0.460 & 0.630 \\
YOLOv8s \cite{yolov8_ultralytics} & 0.575 & 0.853 & 0.646 & 0.295 & 0.492 & 0.661 \\
YOLOv8m \cite{yolov8_ultralytics} & 0.596 & 0.867 & 0.673 & 0.286 & 0.511 & 0.684 \\
YOLOv8l \cite{yolov8_ultralytics} & 0.615 & 0.879 & 0.699 & 0.322 & 0.523 & 0.705 \\
YOLOv8x \cite{yolov8_ultralytics} & 0.611 & 0.879 & 0.689 & 0.303 & 0.516 & 0.704 \\
YOLOv10n \cite{THU-MIGyolov10} & 0.579 & 0.842 & 0.660 & 0.242 & 0.486 & 0.675 \\
YOLOv10s \cite{THU-MIGyolov10} & 0.592 & 0.859 & 0.671 & 0.252 & 0.496 & 0.688 \\
YOLOv10m \cite{THU-MIGyolov10} & 0.622 & 0.877 & 0.704 & 0.286 & 0.519 & 0.718 \\
YOLO11n \cite{yolo11_ultralytics} & 0.584 & 0.842 & 0.662 & 0.228 & 0.489 & 0.686 \\
YOLO11s \cite{yolo11_ultralytics} & 0.607 & 0.863 & 0.685 & 0.266 & 0.515 & 0.704 \\
YOLO11m \cite{yolo11_ultralytics} & 0.609 & 0.866 & 0.688 & 0.272 & 0.511 & 0.705 \\
YOLO11l \cite{yolo11_ultralytics} & 0.622 & 0.875 & 0.706 & 0.287 & 0.521 & 0.717 \\
YOLO11x \cite{yolo11_ultralytics} & \textbf{0.629} & 0.882 & 0.715 & 0.302 & 0.534 & 0.720 \\
YOLO12n \cite{yolo12} & 0.586 & 0.865 & 0.660 & 0.255 & 0.497 & 0.680 \\
YOLO12s \cite{yolo12} & 0.626 & \textbf{0.887} & 0.713 & 0.290 & \textbf{0.535} & 0.719 \\
YOLO12m \cite{yolo12} & 0.625 & 0.886 & 0.713 & 0.285 & 0.532 & 0.721 \\
YOLO12l \cite{yolo12} & 0.629 & 0.876 & \textbf{0.715} & 0.298 & 0.534 & \textbf{0.723} \\
YOLO12x \cite{yolo12} & 0.625 & 0.874 & 0.706 & \textbf{0.303} & 0.525 & 0.721 \\
YOLO-NAS-l \cite{supergradients} & 0.602 & 0.840 & 0.683 & 0.283 & 0.511 & 0.695 \\
Faster R-CNN \cite{Fasterrcnn} & 0.452 & 0.729 & 0.498 & 0.156 & 0.337 & 0.552 \\
Cascade R-CNN \cite{Cascade} & 0.566 & 0.817 & 0.644 & 0.219 & 0.457 & 0.669 \\
Sparse R-CNN  R50 FPN \cite{Sparsercnn} & 0.486 & 0.736 & 0.554 & 0.116 & 0.336 & 0.620 \\
DETR \cite{detr} & 0.540 & 0.827 & 0.607 & 0.143 & 0.388 & 0.677 \\
Deformable-DETR \cite{Deformable} & 0.297 & 0.709 & 0.074 & 0.048 & 0.221 & 0.372 \\
RT-DETR-l \cite{RTDETR} & 0.501 &  0.759  & 0.556 & 0.136 & 0.366 & 0.644 \\
RetinaNet \cite{Retinanet} & 0.578 & 0.842 & 0.654 & 0.236 & 0.475 & 0.676 \\
MobileNetV2-SSD \cite{SSD} & 0.414 & 0.656 & 0.453 & 0.006 & 0.188 & 0.601 \\
FCOS \cite{Fcos} & 0.583 & 0.852 & 0.651 & 0.271 & 0.484 & 0.678 \\
\bottomrule

\end{tabular}
\label{tab:generalization-performance}
\end{table}
\subsection{CLIP-Based Model Selection Analysis}

\begin{figure}[t]
    \centering
    \includegraphics[width=\linewidth]{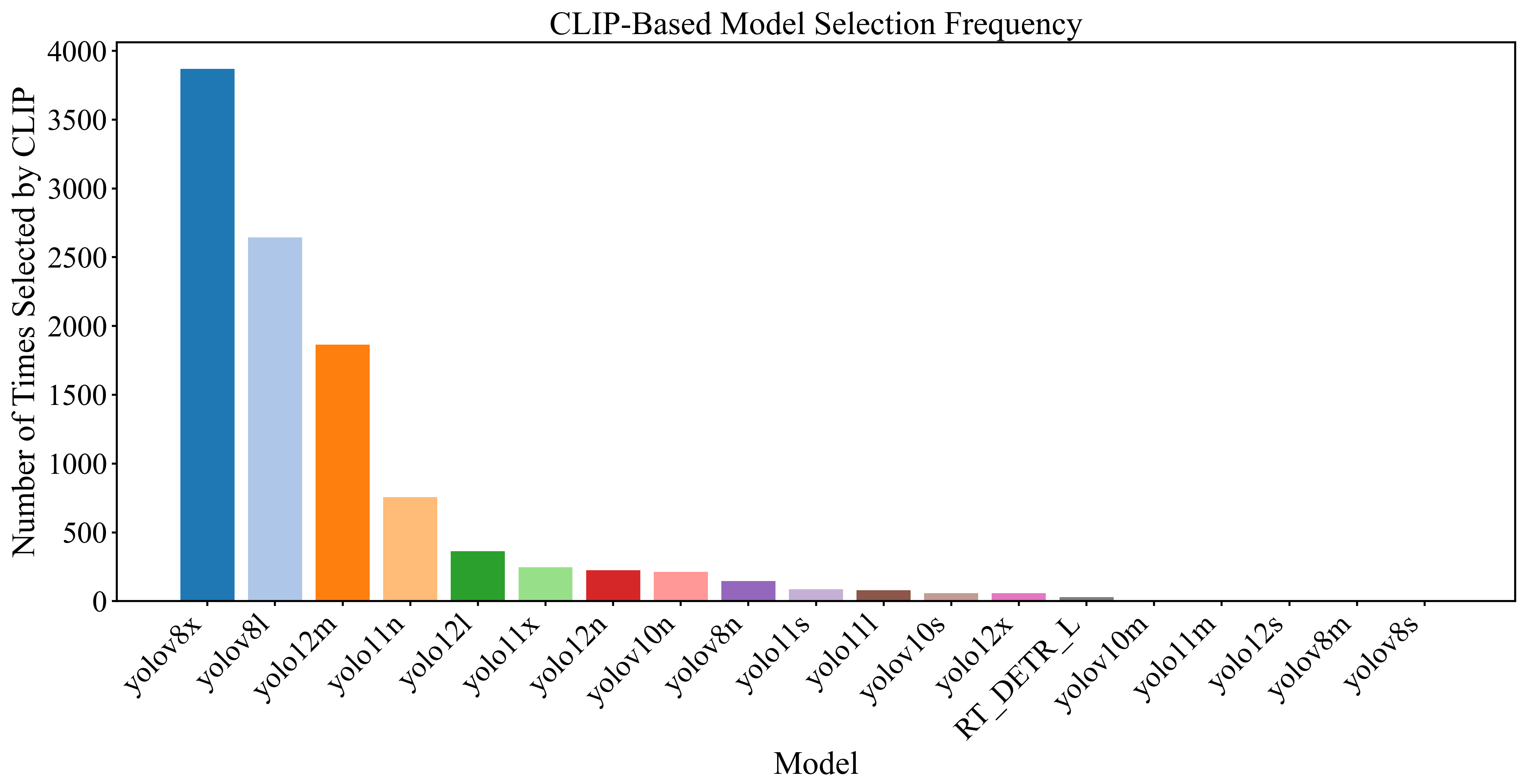}
    \caption{Frequency of model selections based on CLIP similarity across the full FishDet-M dataset.}
    \label{fig:clip_model_selection_frequency}
\end{figure}

We analyzed the behavior of the CLIP-guided model selector to understand its preferences across the FishDet-M dataset. As shown in Fig.~\ref{fig:clip_model_selection_frequency}, YOLOv8x was the most frequently selected model, followed by YOLOv8l, YOLOv12m, and YOLOv11n. These models likely dominated due to stronger feature representation and better alignment with CLIP's semantic cues. In contrast, older or lighter models (e.g., YOLOv10m, YOLOv8m) were selected less often.

\begin{figure}[t]
    \centering
    \includegraphics[width=0.99\linewidth]{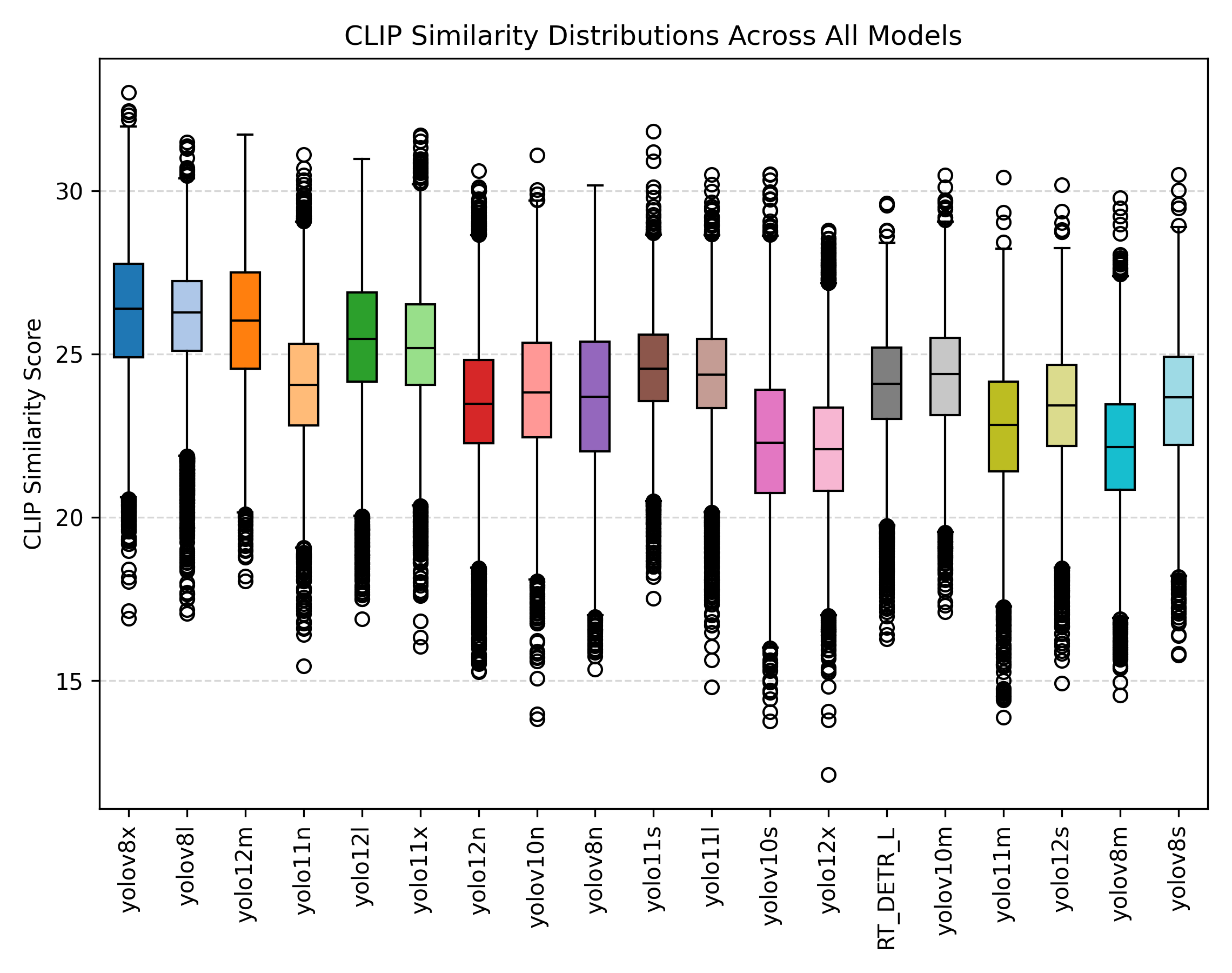}
    \caption{Distribution of CLIP similarity scores across all candidate models. Each box summarizes the semantic alignment over the entire dataset.}
    \label{fig:clip_similarity_boxplot}
\end{figure}

To further examine this trend, we visualized the distribution of CLIP similarity scores for each model in Fig.~\ref{fig:clip_similarity_boxplot}. YOLOv8x, YOLOv8l, and YOLOv12m consistently achieved higher median scores with compact interquartile ranges, indicating reliable semantic alignment. Models with broader or lower scoring distributions were less favored, reflecting CLIP’s tendency to prioritize stable semantic proximity.

These findings confirm that the CLIP selector does not act randomly but favors models with consistent visual-language correspondence. This supports the feasibility of language-driven model routing for adapting detection pipelines to diverse underwater conditions, as demonstrated in recent work on bias-aware underwater AI using CLIP similarity~\cite{saoud2025ebaaiethicsguidedbiasawareai}. The specific prompt used for CLIP-guided selection is provided in the supplementary material for full reproducibility.

\subsection{Discussion and Insights}

The evaluation on FishDet-M highlights consistent performance advantages of the YOLO family, particularly YOLO12x, which offered strong accuracy across object scales and efficient inference suitable for real-time marine applications \cite{yolo12, yolo11_ultralytics, liu2023sonar}. Lightweight YOLO variants also performed well, balancing speed and accuracy effectively.

Transformer based detectors such as DETR, Deformable DETR, and RT DETR-l showed limited performance, especially on small objects, and incurred higher latency \cite{detr, Deformable, RTDETR}. Region based models like Cascade R-CNN and single stage detectors such as RetinaNet and FCOS \cite{Cascade, Retinanet, Fcos} achieved moderate accuracy with better efficiency but did not match YOLO’s top results .

F1 score and precision recall analyses revealed that models behaved similarly at lower IoU thresholds but diverged under stricter matching (IoU 0.75), where lightweight and transformer models were more sensitive to localization demands. This underscores the need to evaluate models across varying thresholds to assess reliability.

Evaluation across individual datasets showed wide performance variation from 0.359 mAP on FishDataset \cite{Liu2023MultitaskModel} to 0.846 on DeepFish \cite{Saleh2020}. This reflects the diversity of underwater conditions. The merged FishDet-M benchmark provided a balanced training distribution, enabling generalizable models that avoid overfitting to specific domains.

Generalization tests on an unseen dataset confirmed this robustness. Many YOLO variants performed comparably or better on new data, reinforcing the value of diverse, consolidated training sets for deployment in real-world underwater applications \cite{Florencedataset}.

We also evaluated an adaptive model selection system, FishDet-M-CLIP, which used CLIP similarity to select contextually appropriate models. Although its accuracy (mAP 0.444) and speed (80 FPS) did not surpass the top YOLO models, it outperformed many traditional and transformer based models, validating the feasibility of language guided model routing for variable conditions.

Qualitative results revealed some failure cases including inaccurate localization in camouflage scenes, missed detections under occlusion, and poor handling of small or blurred objects. False positives around coral and complex structures also persisted, emphasizing ongoing challenges in fine grained discrimination and context-aware detection under underwater conditions.
\subsection{Dataset and Ethics}

FishDet-M is a large-scale benchmark for fish detection across diverse underwater settings \cite{yolov8_ultralytics}. It merges 13 publicly available and licensed datasets covering marine, brackish, and aquarium environments into a fish detection dataset. This integration addresses prior inconsistencies in dataset structure and evaluation \cite{mohankumar2025benchmark}.

The dataset includes 105{,}556 images and 296{,}885 fish annotations, with stratified splits for training, validation, and testing that maintain variation in habitat, clarity, and density. Image sizes range from $78\times53$ to $4608\times3456$ pixels, averaging $711\times465$, with fish counts from 1 to 256 per image. Annotations are labeled under a species-agnostic \textit{Fish} category to prioritize general detection while allowing downstream classification. A thorough quality assurance process ensured format consistency, label standardization, and removal of corrupted or ambiguous data.

All data originate from ethically sourced repositories \cite{Saleh2020, Le2023, 10377207, fish-clean_dataset, fish-video-ls42k_dataset}. The release aims to advance reproducible research in marine science, aquaculture, and conservation \cite{jiang2024dufish, salman2020hybrid}.

\section{Limitations and Future Directions}

Despite its comprehensiveness, FishDet-M retains several limitations. Real-world underwater environments remain highly variable, with extreme turbidity, dynamic lighting, and cluttered backgrounds posing significant challenges for current detection models~\cite{liu2025fishfinder,brackishMOT}. Future datasets may incorporate synthetic augmentation or multisensor data to improve robustness under such conditions.

At present, FishDet-M focuses exclusively on object detection and localization. However, downstream applications such as behavioral analysis and species-specific tracking require richer annotations, including pose, individual identity, and activity labels~\cite{jalal2025deepfins}. Expanding the annotation scope would enable deeper biological insights.

While YOLO-based models outperform others in handling small or occluded fish, limitations remain in complex scenes~\cite{li2023occlusion}. Multi-scale feature fusion and context-aware detection architectures could help mitigate these issues and enhance robustness under occlusion and scale variation.

Generalization analysis further underscores the need to diversify training data. Although models trained on FishDet-M perform well on external benchmarks, broader inclusion of geographic regions, water types, and depths would enhance global applicability.

In terms of adaptive inference, our CLIP-guided model selector shows promise by dynamically routing inputs to suitable detectors. However, this strategy introduces computational overhead and marginally reduces accuracy compared to the top fixed models. Exploring optimized prompts, confidence-weighted voting, or ensemble strategies may offer better speed–accuracy tradeoffs for real-time systems.

Explainability remains an open frontier. Integrating XAI techniques could elucidate model decisions and failure modes in complex underwater environments. This is critical for high-stakes applications in marine conservation and aquaculture, where trust and interpretability are essential.

As reported in Table~\ref{tab:generalization-performance}, top-performing detectors such asvYOLOv12x achieve mAP values exceeding 0.62 on unseen data, outperforming their scores on the FishDet-M test set. This suggests a strong generalization capacity resulting from training on an aggregated dataset such as FishDet-M. This trend extends beyond YOLO, as other detectors such as Faster R-CNN (0.452 mAP), RetinaNet (0.578 mAP), and DETR (0.540 mAP) also exhibit higher scores.

FishDet-M provides a strong foundation for advancing fish detection research. Continued progress will depend on extending annotation depth, improving cross-domain generalization, and refining adaptive and interpretable inference strategies for dynamic underwater settings.

\section{Conclusion}

FishDet-M is a unified benchmark dataset developed to address persistent challenges in underwater fish detection. By merging 13 heterogeneous sources into a standardized repository, it resolves issues related to fragmentation and inconsistent evaluation protocols.
We benchmarked 28 object detection models, including YOLOv8 through YOLOv12, Transformer based architectures, and R-CNN variants. The results indicate the consistent superiority of YOLO models, particularly YOLO12x and YOLO11m, in balancing detection accuracy, robustness across object scales, and inference efficiency. Transformer models exhibited reduced performance when detecting small or occluded fish and imposed greater computational demands.
Evaluation across diverse aquatic conditions revealed substantial variation in detection difficulty. Generalization tests demonstrated that models trained on FishDet-M maintain reliable performance on previously unseen data, underscoring the dataset’s utility in supporting adaptable and field deployable detection systems.
By releasing FishDet-M alongside its annotation scripts and trained models, we aim to foster reproducible research and accelerate innovation in marine science, ecological monitoring, and intelligent aquatic systems.

\section*{Code Availability}

All FishDet-M resources, including the dataset, benchmarking scripts, pretrained models, and documentation, are publicly available through the official project page: \url{https://lyessaadsaoud.github.io/FishDet-M/}.

\section*{Acknowledgments}
This work is supported in part by Khalifa University Center for Autonomous Robotic Systems (KUCARS) under Award RC1-2018-KUCARS and in part by CIRA under Award 8474000419 and 8434000534.

\section*{Declaration}
During the preparation of this work, we used Grammarly and AI tools to improve the English grammar and flow of the paper.


\bibliographystyle{IEEEtran}
\bibliography{fixed_citations_cleaned}

\end{document}


\setcounter{table}{0}
\renewcommand{\thetable}{S\Roman{table}}




\title{Supplementary Material: 
\textit{FishDet-M}}

\author{
    Muayad~Abujabal,~Lyes~Saad~Saoud,~and~Irfan~Hussain
}
\maketitle

\noindent
\begin{strip}
This supplementary document presents comprehensive performance metrics for object detection models evaluated on the FishDet-M benchmark. All models were trained and tested across the 13 constituent underwater datasets to ensure a fair and consistent comparison. Detailed evaluation results are provided in Tables~\ref{tab:fish4knowledge-detection} through~\ref{tab:fishextend-detection}, including mean average precision (mAP), scale-based AP scores, and backbone comparisons. Additionally, the text prompt configuration used for the CLIP-guided model selection process is included in Section V.
\end{strip}

\maketitle



%
\IEEEpeerreviewmaketitle

\begin{table}[t]
\tiny
\centering
\caption{Detection model metrics on Fish4Knowledge \cite{fish4knowledge-dataset_dataset}}

\begin{tabular}{lcccccc}
\toprule
Model & mAP & mAP$_{50}$ & mAP$_{75}$ & AP$_S$ & AP$_M$ & AP$_L$ \\
\midrule
YOLOv8n \cite{yolov8_ultralytics} & 0.427 & 0.791 & 0.397 & 0.395 & 0.767 & 0.891 \\
YOLOv8s \cite{yolov8_ultralytics} & 0.451 & 0.818 & 0.428 & 0.417 & 0.822 & 0.963 \\
YOLOv8m \cite{yolov8_ultralytics} & 0.460 & 0.826 & 0.441 & 0.428 & 0.818 & 0.946 \\
YOLOv8l \cite{yolov8_ultralytics} & 0.462 & 0.829 & 0.439 & 0.428 & 0.843 & 0.926 \\
YOLOv8x \cite{yolov8_ultralytics} & 0.468 & 0.830 & 0.464 & 0.433 & 0.852 & 0.936 \\
YOLOv10n \cite{THU-MIGyolov10} & 0.429 & 0.795 & 0.408 & 0.398 & 0.774 & 0.946 \\
YOLOv10s \cite{THU-MIGyolov10} & 0.451 & 0.817 & 0.424 & 0.417 & 0.826 & 0.951 \\
YOLOv10m \cite{THU-MIGyolov10} & 0.469 & 0.842 & 0.454 & 0.435 & 0.850 & 0.945 \\
YOLO11n \cite{yolo11_ultralytics} & 0.447 & 0.821 & 0.417 & 0.411 & 0.829 & 0.952 \\
YOLO11s \cite{yolo11_ultralytics} & 0.469 & 0.845 & 0.444 & 0.433 & 0.850 & 0.936 \\
YOLO11m \cite{yolo11_ultralytics} & 0.481 & 0.853 & 0.470 & 0.447 & 0.865 & 0.927 \\
YOLO11l \cite{yolo11_ultralytics} & 0.479 & 0.855 & 0.468 & 0.445 & 0.864 & 0.932 \\
YOLO11x \cite{yolo11_ultralytics} & 0.472 & 0.846 & 0.456 & 0.437 & 0.854 & 0.948 \\
YOLO12n \cite{yolo12} & 0.431 & 0.795 & 0.409 & 0.397 & 0.773 & 0.943 \\
YOLO12s \cite{yolo12} & 0.456 & 0.829 & 0.439 & 0.423 & 0.819 & \textbf{0.981} \\
YOLO12m \cite{yolo12} & 0.468 & 0.840 & 0.446 & 0.433 & 0.846 & 0.938 \\
YOLO12l \cite{yolo12} & 0.479 & 0.857 & 0.465 & 0.445 & 0.855 & 0.956 \\
YOLO12x \cite{yolo12} & \textbf{0.485} & \textbf{0.860} & \textbf{0.477} & \textbf{0.449} & \textbf{0.857} & 0.940 \\
YOLO-NAS-l \cite{supergradients} & 0.448 & 0.812 & 0.435 & 0.415 & 0.813 & 0.943 \\
Faster R-CNN \cite{Fasterrcnn} & 0.352 & 0.725 & 0.287 & 0.333 & 0.627 & 0.780 \\
Cascade R-CNN \cite{Cascade} & 0.436 & 0.803 & 0.419 & 0.409 & 0.745 & 0.861 \\
Sparse R-CNN \cite{Sparsercnn} & 0.298 & 0.595 & 0.260 & 0.276 & 0.667 & 0.757 \\
DETR \cite{detr} & 0.298 & 0.678 & 0.210 & 0.267 & 0.690 & 0.752 \\
Deformable-DETR \cite{Deformable} & 0.208 & 0.604 & 0.072 & 0.187 & 0.493 & 0.522 \\
RT-DETR-l \cite{RTDETR} & 0.410 & 0.771 & 0.379 & 0.374 & 0.763 & 0.899 \\
RetinaNet \cite{Retinanet} & 0.432 & 0.804 & 0.402 & 0.399 & 0.783 & 0.934 \\
MobileNetv2-SSD \cite{SSD} & 0.149 & 0.375 & 0.095 & 0.107 & 0.577 & 0.867 \\
FCOS-r50-FPN \cite{Fcos} & 0.430 & 0.817 & 0.405 & 0.405 & 0.761 & 0.924 \\
\bottomrule
\end{tabular}
\label{tab:fish4knowledge-detection}
\end{table}

\begin{table}[t]
\centering
\caption{Detection model metrics on Fish Dataset \cite{Liu2023MultitaskModel}}
\tiny
\begin{tabular}{lcccccc}
\toprule
Model & mAP & mAP$_{50}$ & mAP$_{75}$ & AP$_S$ & AP$_M$ & AP$_L$ \\
\midrule
YOLOv8n \cite{yolov8_ultralytics} & 0.490 & 0.776 & 0.551 & 0.286 & 0.522 & 0.699 \\
YOLOv8s \cite{yolov8_ultralytics} & 0.529 & 0.803 & 0.603 & 0.292 & 0.558 & 0.747 \\
YOLOv8m \cite{yolov8_ultralytics} & 0.547 & 0.812 & 0.619 & 0.302 & 0.576 & 0.771 \\
YOLOv8l \cite{yolov8_ultralytics} & 0.558 & 0.824 & 0.643 & 0.320 & 0.593 & 0.770 \\
YOLOv8x \cite{yolov8_ultralytics} & 0.573 & 0.835 & 0.649 & 0.330 & 0.611 & 0.776 \\
YOLOv10n \cite{THU-MIGyolov10} & 0.516 & 0.783 & 0.588 & 0.287 & 0.536 & 0.767 \\
YOLOv10s \cite{THU-MIGyolov10} & 0.548 & 0.815 & 0.617 & 0.312 & 0.576 & 0.785 \\
YOLOv10m \cite{THU-MIGyolov10} & 0.572 & 0.832 & 0.621 & 0.324 & 0.599 & 0.822 \\
YOLO11n \cite{yolo11_ultralytics} & 0.528 & 0.792 & 0.592 & 0.266 & 0.559 & 0.797 \\
YOLO11s \cite{yolo11_ultralytics} & 0.549 & 0.806 & 0.607 & 0.275 & 0.580 & 0.813 \\
YOLO11m \cite{yolo11_ultralytics} & 0.566 & 0.816 & 0.630 & 0.298 & 0.605 & 0.821 \\
YOLO11l \cite{yolo11_ultralytics} & 0.568 & 0.826 & 0.640 & 0.310 & 0.600 & 0.810 \\
YOLO11x \cite{yolo11_ultralytics} & \textbf{0.584} & \textbf{0.842} & 0.651 & \textbf{0.360} & 0.616 & 0.801 \\
YOLO12n \cite{yolo12} & 0.519 & 0.797 & 0.579 & 0.295 & 0.541 & 0.748 \\
YOLO12s \cite{yolo12} & 0.550 & 0.821 & 0.617 & 0.293 & 0.578 & 0.794 \\
YOLO12m \cite{yolo12} & 0.585 & 0.846 & \textbf{0.672} & 0.348 & \textbf{0.616} & 0.810 \\
YOLO12l \cite{yolo12} & 0.573 & 0.825 & 0.641 & 0.314 & 0.608 & \textbf{0.826} \\
YOLO12x \cite{yolo12} & 0.577 & 0.827 & 0.642 & 0.314 & 0.613 & 0.821 \\
YOLO-NAS-l \cite{supergradients} & 0.538 & 0.789 & 0.612 & 0.263 & 0.583 & 0.776 \\
Faster R-CNN \cite{Fasterrcnn} & 0.395 & 0.690 & 0.428 & 0.249 & 0.412 & 0.559 \\
Cascade R-CNN \cite{Cascade} & 0.495 & 0.769 & 0.569 & 0.276 & 0.507 & 0.730 \\
Sparse R-CNN \cite{Sparsercnn} & 0.401 & 0.640 & 0.450 & 0.176 & 0.421 & 0.673 \\
DETR \cite{detr} & 0.443 & 0.764 & 0.453 & 0.180 & 0.482 & 0.750 \\
Deformable-DETR \cite{Deformable} & 0.378 & 0.729 & 0.332 & 0.150 & 0.398 & 0.646 \\
RT-DETR-l \cite{RTDETR} & 0.509 & 0.801 & 0.563 & 0.245 & 0.540 & 0.740 \\
RetinaNet \cite{Retinanet} & 0.501 & 0.784 & 0.571 & 0.266 & 0.525 & 0.738 \\
MobileNetv2-SSD \cite{SSD} & 0.289 & 0.539 & 0.287 & 0.054 & 0.293 & 0.630 \\
FCOS-r50-FPN \cite{Fcos} & 0.498 & 0.792 & 0.545 & 0.293 & 0.523 & 0.717 \\
\bottomrule
\end{tabular}
\label{tab:fishdataset-detection}
\end{table}

\begin{table}[t]
\centering
\caption{Detection model metrics on Fish Video \cite{fish-video_dataset}}
\tiny
\begin{tabular}{lcccccc}
\toprule
Model & mAP & mAP$_{50}$ & mAP$_{75}$ & AP$_S$ & AP$_M$ & AP$_L$ \\
\midrule
YOLOv8n \cite{yolov8_ultralytics} & 0.635 & 0.910 & 0.730 & 0.385 & 0.624 & 0.763 \\
YOLOv8s \cite{yolov8_ultralytics} & 0.663 & 0.927 & 0.747 & 0.397 & 0.653 & 0.787 \\
YOLOv8m \cite{yolov8_ultralytics} & 0.668 & 0.932 & 0.762 & 0.457 & 0.651 & 0.789 \\
YOLOv8l \cite{yolov8_ultralytics} & 0.674 & 0.932 & 0.774 & 0.439 & 0.658 & 0.797 \\
YOLOv8x \cite{yolov8_ultralytics} & 0.680 & 0.934 & 0.770 & 0.469 & 0.665 & 0.802 \\
YOLOv10n \cite{THU-MIGyolov10} & 0.638 & 0.913 & 0.733 & 0.402 & 0.627 & 0.777 \\
YOLOv10s \cite{THU-MIGyolov10} & 0.656 & 0.920 & 0.758 & 0.430 & 0.642 & 0.793 \\
YOLOv10m \cite{THU-MIGyolov10} & 0.673 & 0.931 & \textbf{0.788} & 0.478 & 0.650 & 0.819 \\
YOLO11n \cite{yolo11_ultralytics} & 0.652 & 0.920 & 0.748 & 0.420 & 0.636 & 0.798 \\
YOLO11s \cite{yolo11_ultralytics} & 0.676 & 0.940 & 0.774 & 0.452 & 0.656 & 0.822 \\
YOLO11m \cite{yolo11_ultralytics} & 0.692 & 0.940 & 0.799 & 0.459 & 0.670 & 0.834 \\
YOLO11l \cite{yolo11_ultralytics} & 0.685 & 0.936 & 0.784 & 0.479 & 0.664 & 0.830 \\
YOLO11x \cite{yolo11_ultralytics} & 0.683 & 0.936 & 0.789 & 0.462 & 0.669 & 0.815 \\
YOLO12n \cite{yolo12} & 0.643 & 0.922 & 0.733 & 0.387 & 0.624 & 0.787 \\
YOLO12s \cite{yolo12} & 0.670 & 0.929 & 0.765 & 0.425 & 0.650 & 0.810 \\
YOLO12m \cite{yolo12} & 0.682 & 0.939 & 0.782 & 0.468 & 0.665 & 0.818 \\
YOLO12l \cite{yolo12} & 0.686 & 0.939 & 0.786 & 0.449 & 0.668 & 0.824 \\
YOLO12x \cite{yolo12} & \textbf{0.687} & \textbf{0.940} & 0.794 & 0.464 & 0.670 & 0.816 \\
YOLO-NAS-l \cite{supergradients} & 0.665 & 0.93 & 0.777 & \textbf{0.472} & 0.648 & 0.791 \\
Faster R-CNN \cite{Fasterrcnn} & 0.504 & 0.823 & 0.565 & 0.220 & 0.496 & 0.651 \\
Cascade R-CNN \cite{Cascade}& 0.607 & 0.891 & 0.701 & 0.290 & 0.601 & 0.752 \\
Sparse R-CNN \cite{Sparsercnn} & 0.475 & 0.708 & 0.562 & 0.150 & 0.460 & 0.701 \\
DETR \cite{detr} & 0.499 & 0.850 & 0.523 & 0.165 & 0.463 & 0.721 \\
Deformable-DETR \cite{Deformable} & 0.294 & 0.738 & 0.089 & 0.088 & 0.295 & 0.394 \\
RT-DETR-l \cite{RTDETR} & 0.629 & 0.923 & 0.722 & 0.299 & 0.615 & 0.778 \\
RetinaNet \cite{Retinanet} & 0.608 & 0.893 & 0.709 & 0.302 & 0.599 & 0.757 \\
MobileNetv2-SSD \cite{SSD} & 0.349 & 0.675 & 0.320 & 0.054 & 0.294 & 0.645 \\
FCOS-r50-FPN \cite{Fcos} & 0.594 & 0.912 & 0.675 & 0.296 & 0.589 & 0.732 \\
\bottomrule
\end{tabular}
\label{tab:fishvideo-detection}
\end{table}

\begin{table}[t]
\centering
\caption{Detection model metrics on Fishnet \cite{10377207}}
\tiny
\begin{tabular}{lcccccc}
\toprule
Model & mAP & mAP$_{50}$ & mAP$_{75}$ & AP$_S$ & AP$_M$ & AP$_L$ \\
\midrule
YOLOv8n \cite{yolov8_ultralytics} & 0.462 & 0.822 & 0.483 & 0.242 & 0.429 & 0.655 \\
YOLOv8s \cite{yolov8_ultralytics} & 0.481 & 0.847 & 0.509 & 0.267 & 0.452 & 0.668 \\
YOLOv8m \cite{yolov8_ultralytics} & 0.493 & 0.862 & 0.522 & 0.280 & 0.463 & 0.680 \\
YOLOv8l \cite{yolov8_ultralytics} & 0.501 & 0.869 & 0.539 & 0.289 & 0.473 & 0.685 \\
YOLOv8x \cite{yolov8_ultralytics} & 0.503 & 0.871 & 0.536 & 0.291 & 0.475 & 0.686 \\
YOLOv10n \cite{THU-MIGyolov10} & 0.470 & 0.831 & 0.502 & 0.254 & 0.433 & 0.663 \\
YOLOv10s \cite{THU-MIGyolov10} & 0.485 & 0.850 & 0.517 & 0.269 & 0.448 & 0.673 \\
YOLOv10m \cite{THU-MIGyolov10} & 0.498 & 0.867 & 0.535 & 0.289 & 0.465 & 0.680 \\
YOLO11n \cite{yolo11_ultralytics} & 0.480 & 0.838 & 0.510 & 0.252 & 0.451 & 0.675 \\
YOLO11s \cite{yolo11_ultralytics} & 0.493 & 0.858 & 0.526 & 0.273 & 0.466 & 0.680 \\
YOLO11m \cite{yolo11_ultralytics} & 0.500 & 0.873 & 0.534 & 0.289 & 0.474 & 0.681 \\
YOLO11l \cite{yolo11_ultralytics} & 0.505 & 0.877 & 0.538 & 0.296 & 0.477 & 0.687 \\
YOLO11x \cite{yolo11_ultralytics} & 0.504 & 0.876 & 0.539 & 0.294 & 0.478 & 0.684 \\
YOLO12n \cite{yolo12} & 0.474 & 0.834 & 0.499 & 0.246 & 0.445 & 0.666 \\
YOLO12s \cite{yolo12} & 0.494 & 0.861 & 0.526 & 0.275 & 0.470 & 0.681 \\
YOLO12m \cite{yolo12} & 0.504 & 0.876 & 0.540 & 0.292 & 0.478 & 0.687 \\
YOLO12l \cite{yolo12} & 0.508 & 0.878 & 0.547 & 0.298 & 0.485 & 0.687 \\
YOLO12x \cite{yolo12} & \textbf{0.512} & \textbf{0.884} & \textbf{0.553} & \textbf{0.310} & \textbf{0.489} & \textbf{0.687} \\
YOLO-NAS-l \cite{supergradients} & 0.493 & 0.839 & 0.529 & 0.277 & 0.467 & 0.673 \\
Faster R-CNN \cite{Fasterrcnn} & 0.404 & 0.716 & 0.412 & 0.170 & 0.359 & 0.594 \\
Cascade R-CNN \cite{Cascade} & 0.472 & 0.773 & 0.505 & 0.239 & 0.426 & 0.661 \\
Sparse R-CNN \cite{Sparsercnn} & 0.387 & 0.643 & 0.422 & 0.103 & 0.326 & 0.627 \\
DETR \cite{detr} & 0.439 & 0.739 & 0.472 & 0.138 & 0.401 & 0.668 \\
Deformable-DETR \cite{Deformable} & 0.391 & 0.717 & 0.381 & 0.128 & 0.362 & 0.588 \\
RT-DETR-l \cite{RTDETR} & 0.476 & 0.846 & 0.499 & 0.234 & 0.467 & 0.672 \\
RetinaNet \cite{Retinanet} & 0.476 & 0.780 & 0.512 & 0.217 & 0.435 & 0.681 \\
MobileNetv2-SSD \cite{SSD} & 0.345 & 0.590 & 0.373 & 0.020 & 0.260 & 0.633 \\
FCOS-r50-FPN \cite{Fcos} & 0.480 & 0.807 & 0.505 & 0.257 & 0.437 & 0.657 \\
\bottomrule
\end{tabular}
\label{tab:fishnet-detection}
\end{table}

\begin{table}[t]
\centering
\caption{Detection model metrics on TrashCan 1.0 \cite{hong2020trashcan}}
\tiny
\begin{tabular}{lcccccc}
\toprule
Model & mAP & mAP$_{50}$ & mAP$_{75}$ & AP$_S$ & AP$_M$ & AP$_L$ \\
\midrule
YOLOv8n \cite{yolov8_ultralytics} & 0.259 & 0.490 & 0.222 & 0.161 & 0.289 & 0.788 \\
YOLOv8s \cite{yolov8_ultralytics} & 0.301 & 0.550 & 0.260 & 0.181 & 0.338 & 0.826 \\
YOLOv8m \cite{yolov8_ultralytics} & 0.370 & 0.677 & 0.352 & 0.248 & 0.408 & 0.820 \\
YOLOv8l \cite{yolov8_ultralytics} & 0.377 & 0.675 & 0.391 & 0.257 & 0.417 & \textbf{0.875} \\
YOLOv8x \cite{yolov8_ultralytics} & 0.364 & 0.658 & 0.326 & 0.241 & 0.423 & 0.776 \\
YOLOv10n \cite{THU-MIGyolov10} & 0.354 & 0.642 & 0.335 & 0.252 & 0.380 & 0.850 \\
YOLOv10s \cite{THU-MIGyolov10} & 0.364 & 0.655 & 0.330 & 0.243 & 0.414 & 0.801 \\
YOLOv10m \cite{THU-MIGyolov10} & 0.373 & 0.688 & 0.347 & 0.234 & 0.433 & 0.789 \\
YOLO11n \cite{yolo11_ultralytics} & 0.372 & 0.663 & 0.348 & 0.253 & 0.407 & 0.801 \\
YOLO11s \cite{yolo11_ultralytics} & 0.373 & 0.709 & 0.363 & 0.220 & 0.436 & 0.751 \\
YOLO11m \cite{yolo11_ultralytics} & 0.395 & 0.706 & 0.396 & 0.265 & 0.466 & 0.727 \\
YOLO11l \cite{yolo11_ultralytics} & 0.340 & 0.653 & 0.322 & 0.216 & 0.401 & 0.711 \\
YOLO11x \cite{yolo11_ultralytics} & 0.383 & 0.686 & 0.351 & 0.268 & 0.435 & 0.850 \\
YOLO12n \cite{yolo12} & 0.322 & 0.604 & 0.331 & 0.227 & 0.342 & 0.820 \\
YOLO12s \cite{yolo12} & 0.396 & 0.702 & 0.359 & 0.294 & 0.427 & 0.801 \\
YOLO12m \cite{yolo12} & 0.381 & 0.714 & 0.360 & 0.281 & 0.418 & 0.727 \\
YOLO12l \cite{yolo12} & 0.368 & 0.695 & 0.308 & 0.266 & 0.410 & 0.751 \\
YOLO12x \cite{yolo12} & 0.391 & \textbf{0.736} & 0.335 & 0.231 & \textbf{0.490} & 0.671 \\
YOLO-NAS-l \cite{supergradients} & 0.296 & 0.533 & 0.224 & 0.125 & 0.387 & 0.751 \\
Faster R-CNN \cite{Fasterrcnn} & 0.156 & 0.326 & 0.107 & 0.134 & 0.175 & 0.338 \\
Cascade R-CNN \cite{Cascade} & 0.253 & 0.481 & 0.277 & 0.197 & 0.262 & 0.766 \\
Sparse R-CNN \cite{Sparsercnn} & 0.269 & 0.544 & 0.261 & 0.236 & 0.282 & 0.701 \\
DETR \cite{detr} & 0.250 & 0.499 & 0.236 & 0.115 & 0.297 & \textbf{0.888} \\
Deformable-DETR \cite{Deformable} & 0.138 & 0.402 & 0.072 & 0.083 & 0.205 & 0.605 \\
RT-DETR-l \cite{RTDETR} & 0.256 & 0.501 & 0.240 & 0.158 & 0.273 & 0.763 \\
RetinaNet \cite{Retinanet} & 0.264 & 0.520 & 0.236 & 0.175 & 0.265 & 0.832 \\
MobileNetv2-SSD \cite{SSD} & 0.151 & 0.323 & 0.132 & 0.064 & 0.158 & 0.776 \\
FCOS-r50-FPN \cite{Fcos} & 0.230 & 0.426 & 0.226 & 0.116 & 0.274 & 0.729 \\
\bottomrule
\end{tabular}
\label{tab:trashcan-detection}
\end{table}

\begin{table}[t]
\centering
\caption{Detection model metrics on UIIS \cite{Lian_2023_ICCV}}
\tiny
\begin{tabular}{lcccccc}
\toprule
Model & mAP & mAP$_{50}$ & mAP$_{75}$ & AP$_S$ & AP$_M$ & AP$_L$ \\
\midrule
YOLOv8n \cite{yolov8_ultralytics} & 0.490 & 0.776 & 0.551 & 0.286 & 0.522 & 0.699 \\
YOLOv8s \cite{yolov8_ultralytics} & 0.529 & 0.803 & 0.603 & 0.292 & 0.558 & 0.747 \\
YOLOv8m \cite{yolov8_ultralytics} & 0.547 & 0.812 & 0.619 & 0.302 & 0.576 & 0.771 \\
YOLOv8l \cite{yolov8_ultralytics} & 0.558 & 0.824 & 0.643 & 0.320 & 0.593 & 0.770 \\
YOLOv8x \cite{yolov8_ultralytics} & 0.573 & 0.835 & 0.649 & 0.330 & 0.611 & 0.776 \\
YOLOv10n \cite{THU-MIGyolov10} & 0.516 & 0.783 & 0.588 & 0.287 & 0.536 & 0.767 \\
YOLOv10s \cite{THU-MIGyolov10} & 0.548 & 0.815 & 0.617 & 0.312 & 0.576 & 0.785 \\
YOLOv10m \cite{THU-MIGyolov10} & 0.572 & 0.832 & 0.621 & 0.324 & 0.599 & 0.822 \\
YOLO11n \cite{yolo11_ultralytics} & 0.528 & 0.792 & 0.592 & 0.266 & 0.559 & 0.797 \\
YOLO11s \cite{yolo11_ultralytics} & 0.549 & 0.806 & 0.607 & 0.275 & 0.580 & 0.813 \\
YOLO11m \cite{yolo11_ultralytics} & 0.566 & 0.816 & 0.630 & 0.298 & 0.605 & 0.821 \\
YOLO11l \cite{yolo11_ultralytics} & 0.568 & 0.826 & 0.640 & 0.310 & 0.600 & 0.810 \\
YOLO11x \cite{yolo11_ultralytics} & \textbf{0.584} & \textbf{0.842} & 0.651 & \textbf{0.360} & 0.616 & 0.801 \\
YOLO12n \cite{yolo12} & 0.519 & 0.797 & 0.579 & 0.295 & 0.541 & 0.748 \\
YOLO12s \cite{yolo12} & 0.550 & 0.821 & 0.617 & 0.293 & 0.578 & 0.794 \\
YOLO12m \cite{yolo12} & 0.585 & 0.846 & \textbf{0.672} & 0.348 & \textbf{0.616} & 0.810 \\
YOLO12l \cite{yolo12} & 0.573 & 0.825 & 0.641 & 0.314 & 0.608 & \textbf{0.826} \\
YOLO12x \cite{yolo12} & 0.577 & 0.827 & 0.642 & 0.314 & 0.613 & 0.821 \\
YOLO-NAS-l \cite{supergradients} & 0.538 & 0.789 & 0.612 & 0.263 & 0.583 & 0.776 \\
Faster R-CNN \cite{Fasterrcnn} & 0.395 & 0.690 & 0.428 & 0.249 & 0.412 & 0.559 \\
Cascade R-CNN \cite{Cascade} & 0.495 & 0.769 & 0.569 & 0.276 & 0.507 & 0.730 \\
Sparse R-CNN \cite{Sparsercnn} & 0.401 & 0.640 & 0.450 & 0.176 & 0.421 & 0.673 \\
DETR \cite{detr} & 0.443 & 0.764 & 0.453 & 0.180 & 0.482 & 0.750 \\
Deformable-DETR \cite{Deformable} & 0.378 & 0.729 & 0.332 & 0.150 & 0.398 & 0.646 \\
RT-DETR-l \cite{RTDETR} & 0.509 & 0.801 & 0.563 & 0.245 & 0.54 & 0.74 \\
RetinaNet \cite{Retinanet} & 0.501 & 0.784 & 0.571 & 0.266 & 0.525 & 0.738 \\
MobileNetv2-SSD \cite{SSD} & 0.289 & 0.539 & 0.287 & 0.054 & 0.293 & 0.630 \\
FCOS-r50-FPN \cite{Fcos} & 0.498 & 0.792 & 0.545 & 0.293 & 0.523 & 0.717 \\
\bottomrule
\end{tabular}
\label{tab:uiis-detection}
\end{table}

\begin{table}[t]
\centering
\caption{Detection model metrics on Aquarium \cite{detect-aqurium_dataset}}
\tiny
\begin{tabular}{lcccccc}
\toprule
Model & mAP & mAP$_{50}$ & mAP$_{75}$ & AP$_S$ & AP$_M$ & AP$_L$ \\
\midrule
YOLOv8n \cite{yolov8_ultralytics} & 0.431 & 0.752 & 0.451 & 0.130 & 0.391 & 0.524 \\
YOLOv8s \cite{yolov8_ultralytics} & 0.480 & 0.812 & 0.494 & 0.188 & 0.442 & 0.564 \\
YOLOv8m \cite{yolov8_ultralytics} & 0.504 & 0.842 & 0.535 & 0.221 & 0.469 & 0.588 \\
YOLOv8l \cite{yolov8_ultralytics} & 0.519 & 0.848 & 0.531 & 0.222 & 0.491 & 0.597 \\
YOLOv8x \cite{yolov8_ultralytics} & 0.534 & 0.864 & 0.555 & 0.234 & 0.503 & 0.615 \\
YOLOv10n \cite{THU-MIGyolov10} & 0.451 & 0.788 & 0.458 & 0.178 & 0.414 & 0.536 \\
YOLOv10s \cite{THU-MIGyolov10} & 0.500 & 0.837 & 0.504 & 0.208 & 0.456 & 0.587 \\
YOLOv10m \cite{THU-MIGyolov10} & \textbf{0.576} & \textbf{0.886} & \textbf{0.624} & 0.252 & 0.543 & 0.661 \\
YOLO11n \cite{yolo11_ultralytics} & 0.490 & 0.822 & 0.514 & 0.194 & 0.444 & 0.585 \\
YOLO11s \cite{yolo11_ultralytics} & 0.533 & 0.851 & 0.562 & 0.168 & 0.493 & 0.629 \\
YOLO11m \cite{yolo11_ultralytics} & 0.574 & 0.877 & 0.626 & 0.195 & 0.543 & 0.665 \\
YOLO11l \cite{yolo11_ultralytics} & 0.580 & 0.890 & 0.632 & \textbf{0.266} & 0.540 & 0.672 \\
YOLO11x \cite{yolo11_ultralytics} & 0.551 & 0.868 & 0.593 & 0.256 & 0.520 & 0.636 \\
YOLO12n \cite{yolo12} & 0.449 & 0.790 & 0.452 & 0.148 & 0.395 & 0.550 \\
YOLO12s \cite{yolo12} & 0.522 & 0.859 & 0.545 & 0.216 & 0.474 & 0.620 \\
YOLO12m \cite{yolo12} & 0.540 & 0.873 & 0.560 & 0.232 & 0.500 & 0.628 \\
YOLO12l \cite{yolo12} & 0.586 & \textbf{0.905} & 0.637 & 0.267 & \textbf{0.561} & 0.664 \\
YOLO12x \cite{yolo12} & 0.577 & 0.888 & 0.623 & 0.212 & 0.547 & 0.664 \\
YOLO-NAS-l \cite{supergradients} & 0.511 & 0.827 & 0.527 & 0.234 & 0.479 & 0.596 \\
Faster R-CNN \cite{Fasterrcnn} & 0.371 & 0.701 & 0.356 & 0.114 & 0.331 & 0.453 \\
Cascade R-CNN \cite{Cascade} & 0.457 & 0.777 & 0.466 & 0.210 & 0.414 & 0.544 \\
Sparse R-CNN \cite{Sparsercnn} & 0.346 & 0.562 & 0.376 & 0.094 & 0.279 & 0.467 \\
DETR \cite{detr} & 0.368 & 0.724 & 0.340 & 0.087 & 0.279 & 0.516 \\
Deformable-DETR \cite{Deformable} & 0.316 & 0.699 & 0.248 & 0.066 & 0.252 & 0.445 \\
RT-DETR-l \cite{RTDETR} & 0.469 & 0.814 & 0.473 & 0.183 & 0.429 & 0.569 \\
RetinaNet \cite{Retinanet} & 0.485 & 0.831 & 0.486 & 0.204 & 0.443 & 0.576 \\
MobileNetv2-SSD \cite{SSD} & 0.194 & 0.454 & 0.146 & 0.001 & 0.084 & 0.363 \\
FCOS-r50-FPN \cite{Fcos} & 0.442 & 0.772 & 0.463 & 0.164 & 0.409 & 0.520 \\
\bottomrule
\end{tabular}
\label{tab:aquarium-detection}
\end{table}

\begin{table}[t]
\centering
\caption{Detection model metrics on Brackish-MOT \cite{brackishMOT}}
\tiny
\begin{tabular}{lcccccc}
\toprule
Model & mAP & mAP$_{50}$ & mAP$_{75}$ & AP$_S$ & AP$_M$ & AP$_L$ \\
\midrule
YOLOv8n \cite{yolov8_ultralytics} & 0.367 & 0.662 & 0.371 & 0.052 & 0.377 & 0.367 \\
YOLOv8s \cite{yolov8_ultralytics} & 0.383 & 0.707 & 0.369 & 0.081 & 0.381 & 0.420 \\
YOLOv8m \cite{yolov8_ultralytics} & 0.435 & 0.757 & 0.452 & 0.202 & 0.443 & 0.439 \\
YOLOv8l \cite{yolov8_ultralytics} & 0.445 & 0.762 & 0.466 & 0.209 & 0.450 & 0.453 \\
YOLOv8x \cite{yolov8_ultralytics} & 0.449 & 0.769 & 0.464 & 0.193 & 0.463 & 0.429 \\
YOLOv10n \cite{THU-MIGyolov10} & 0.356 & 0.636 & 0.364 & 0.116 & 0.362 & 0.356 \\
YOLOv10s \cite{THU-MIGyolov10} & 0.403 & 0.692 & 0.424 & 0.144 & 0.406 & 0.419 \\
YOLOv10m \cite{THU-MIGyolov10} & 0.423 & 0.723 & 0.446 & 0.160 & 0.430 & 0.429 \\
YOLO11n \cite{yolo11_ultralytics} & 0.352 & 0.667 & 0.329 & 0.090 & 0.356 & 0.351 \\
YOLO11s \cite{yolo11_ultralytics} & 0.425 & 0.747 & 0.439 & 0.173 & 0.438 & 0.393 \\
YOLO11m \cite{yolo11_ultralytics} & 0.424 & 0.738 & 0.447 & 0.208 & 0.429 & 0.434 \\
YOLO11l \cite{yolo11_ultralytics} & 0.423 & 0.749 & 0.428 & 0.187 & 0.430 & 0.435 \\
YOLO11x \cite{yolo11_ultralytics} & 0.432 & 0.744 & 0.460 & 0.211 & 0.444 & 0.424 \\
YOLO12n \cite{yolo12} & 0.347 & 0.651 & 0.334 & 0.035 & 0.352 & 0.374 \\
YOLO12s \cite{yolo12} & 0.403 & 0.713 & 0.403 & 0.149 & 0.406 & 0.417 \\
YOLO12m \cite{yolo12} & 0.442 & 0.762 & 0.463 & 0.186 & 0.442 & 0.465 \\
YOLO12l \cite{yolo12} & \textbf{0.461} & \textbf{0.798} & \textbf{0.485} & 0.198 & \textbf{0.477} & 0.439 \\
YOLO12x \cite{yolo12} & 0.454 & 0.769 & 0.486 & \textbf{0.226} & 0.465 & 0.436 \\
YOLO-NAS-l \cite{supergradients} & 0.399 & 0.719 & 0.395 & 0.119 & 0.413 & 0.370 \\
Faster R-CNN \cite{Fasterrcnn} & 0.303 & 0.577 & 0.293 & 0.115 & 0.313 & 0.295 \\
Cascade R-CNN \cite{Cascade} & 0.363 & 0.665 & 0.362 & 0.163 & 0.376 & 0.351 \\
Sparse R-CNN \cite{Sparsercnn} & 0.249 & 0.513 & 0.225 & 0.129 & 0.257 & 0.296 \\
DETR \cite{detr} & 0.199 & 0.553 & 0.096 & 0.019 & 0.168 & 0.353 \\
Deformable-DETR \cite{Deformable} & 0.089 & 0.354 & 0.006 & 0.001 & 0.077 & 0.159 \\
RT-DETR-l \cite{RTDETR} & 0.400 & 0.729 & 0.403 & 0.097 & 0.406 & 0.402 \\
RetinaNet \cite{Retinanet} & 0.335 & 0.704 & 0.266 & 0.047 & 0.333 & 0.382 \\
MobileNetv2-SSD \cite{SSD} & 0.083 & 0.272 & 0.038 & 0.000 & 0.058 & 0.199 \\
FCOS-r50-FPN \cite{Fcos} & 0.386 & 0.736 & 0.363 & 0.182 & 0.392 & 0.382 \\
\bottomrule
\end{tabular}
\label{tab:brackish-mot-detection}
\end{table}

\begin{table}[t]
\centering
\caption{Detection metrics on labeled Med. Fish \cite{10.3389/fmars.2023.1151758}}
\tiny
\begin{tabular}{lcccccc}
\toprule
Model & mAP & mAP$_{50}$ & mAP$_{75}$ & AP$_S$ & AP$_M$ & AP$_L$ \\
\midrule
YOLOv8n \cite{yolov8_ultralytics} & 0.254 & 0.526 & 0.208 & 0.077 & 0.237 & 0.496 \\
YOLOv8s \cite{yolov8_ultralytics} & 0.312 & 0.623 & 0.273 & 0.099 & 0.313 & 0.553 \\
YOLOv8m \cite{yolov8_ultralytics} & 0.348 & 0.663 & 0.322 & 0.117 & 0.354 & 0.598 \\
YOLOv8l \cite{yolov8_ultralytics} & 0.365 & 0.692 & 0.336 & 0.129 & 0.371 & 0.617 \\
YOLOv8x \cite{yolov8_ultralytics} & 0.376 & 0.699 & 0.350 & 0.137 & 0.389 & 0.625 \\
YOLOv10n \cite{THU-MIGyolov10} & 0.268 & 0.551 & 0.223 & 0.084 & 0.252 & 0.518 \\
YOLOv10s \cite{THU-MIGyolov10} & 0.319 & 0.631 & 0.290 & 0.111 & 0.315 & 0.576 \\
YOLOv10m \cite{THU-MIGyolov10} & 0.368 & 0.695 & 0.339 & 0.121 & 0.377 & 0.624 \\
YOLO11n \cite{yolo11_ultralytics} & 0.280 & 0.575 & 0.230 & 0.084 & 0.264 & 0.533 \\
YOLO11s \cite{yolo11_ultralytics} & 0.338 & 0.658 & 0.300 & 0.107 & 0.340 & 0.595 \\
YOLO11m \cite{yolo11_ultralytics} & 0.379 & 0.713 & 0.354 & 0.130 & 0.395 & 0.623 \\
YOLO11l \cite{yolo11_ultralytics} & 0.389 & 0.719 & 0.367 & 0.137 & 0.403 & 0.637 \\
YOLO11x \cite{yolo11_ultralytics} & 0.389 & 0.715 & 0.363 & 0.146 & 0.403 & 0.635 \\
YOLO12n \cite{yolo12} & 0.264 & 0.553 & 0.209 & 0.076 & 0.247 & 0.516 \\
YOLO12s \cite{yolo12} & 0.333 & 0.651 & 0.294 & 0.107 & 0.334 & 0.590 \\
YOLO12m \cite{yolo12} & 0.373 & 0.698 & 0.349 & 0.134 & 0.382 & 0.618 \\
YOLO12l \cite{yolo12} & 0.389 & 0.723 & 0.369 & 0.151 & 0.399 & 0.635 \\
YOLO12x \cite{yolo12} & \textbf{0.406} & \textbf{0.740} & \textbf{0.394} & \textbf{0.161} & \textbf{0.415} & \textbf{0.655} \\
YOLO-NAS-l \cite{supergradients} & 0.343 & 0.658 & 0.313 & 0.118 & 0.356 & 0.582 \\
Faster R-CNN \cite{Fasterrcnn} & 0.309 & 0.638 & 0.257 & 0.150 & 0.334 & 0.466 \\
Cascade R-CNN \cite{Cascade} & 0.407 & 0.736 & 0.403 & 0.203 & 0.420 & 0.614 \\
Sparse R-CNN \cite{Sparsercnn} & 0.176 & 0.344 & 0.154 & 0.058 & 0.172 & 0.345 \\
DETR \cite{detr} & 0.219 & 0.531 & 0.148 & 0.029 & 0.199 & 0.467 \\
Deformable-DETR \cite{Deformable} & 0.104 & 0.343 & 0.020 & 0.012 & 0.093 & 0.294 \\
RT-DETR-l \cite{RTDETR} & 0.305 & 0.623 & 0.259 & 0.073 & 0.308 & 0.553 \\
RetinaNet \cite{Retinanet} & 0.340 & 0.654 & 0.315 & 0.090 & 0.336 & 0.594 \\
MobileNetv2-SSD \cite{SSD} & 0.057 & 0.132 & 0.046 & 0.000 & 0.012 & 0.261 \\
FCOS-r50-FPN \cite{Fcos} & 0.385 & 0.725 & 0.352 & 0.164 & 0.411 & 0.585 \\
\bottomrule
\end{tabular}
\label{tab:labelled-mediterranean-fish-detection}
\end{table}

\begin{table}[t]
\centering
\captionof{table}{Detection model metrics on DeepFish \cite{Saleh2020} }
\tiny
\begin{tabular}{lcccccc}
\toprule
Model & mAP & mAP$_{50}$ & mAP$_{75}$ & AP$_S$ & AP$_M$ & AP$_L$ \\
\midrule
YOLOv8n \cite{yolov8_ultralytics} & 0.631 & 0.868 & 0.761 & -1.000 & -1.000 & 0.641 \\
YOLOv8s \cite{yolov8_ultralytics} & 0.702 & 0.931 & 0.834 & -1.000 & -1.000 & 0.715 \\
YOLOv8m \cite{yolov8_ultralytics} & 0.730 & 0.946 & 0.862 & -1.000 & -1.000 & 0.744 \\
YOLOv8l \cite{yolov8_ultralytics} & 0.776 & 0.956 & 0.921 & -1.000 & -1.000 & 0.793 \\
YOLOv8x \cite{yolov8_ultralytics} & 0.772 & 0.961 & 0.890 & -1.000 & -1.000 & 0.789 \\
YOLOv10n \cite{THU-MIGyolov10} & 0.662 & 0.893 & 0.773 & -1.000 & -1.000 & 0.674 \\
YOLOv10s \cite{THU-MIGyolov10} & 0.752 & 0.931 & 0.851 & -1.000 & -1.000 & 0.768 \\
YOLOv10m \cite{THU-MIGyolov10} & \textbf{0.825} & \textbf{0.968} & 0.896 & -1.000 & -1.000 & 0.844 \\
YOLO11n \cite{yolo11_ultralytics} & 0.726 & 0.926 & 0.866 & -1.000 & -1.000 & 0.740 \\
YOLO11s \cite{yolo11_ultralytics} & 0.827 & 0.968 & \textbf{0.949} & -1.000 & -1.000 & 0.842 \\
YOLO11m \cite{yolo11_ultralytics} & 0.848 & \textbf{0.978} & \textbf{0.950} & -1.000 & -1.000 & 0.862 \\
YOLO11l \cite{yolo11_ultralytics} & 0.826 & 0.974 & 0.926 & -1.000 & -1.000 & 0.839 \\
YOLO11x \cite{yolo11_ultralytics} & 0.798 & 0.954 & 0.888 & -1.000 & -1.000 & 0.816 \\
YOLO12n \cite{yolo12} & 0.685 & 0.892 & 0.802 & -1.000 & -1.000 & 0.697 \\
YOLO12s \cite{yolo12} & 0.764 & 0.939 & 0.881 & -1.000 & -1.000 & 0.779 \\
YOLO12m \cite{yolo12} & 0.819 & 0.966 & 0.936 & -1.000 & -1.000 & 0.836 \\
YOLO12l \cite{yolo12} & \textbf{0.850} & 0.970 & 0.914 & -1.000 & -1.000 & \textbf{0.867} \\
YOLO12x \cite{yolo12} & 0.846 & 0.968 & 0.931 & -1.000 & -1.000 & 0.863 \\
YOLO-NAS-l \cite{supergradients} & 0.752 & 0.911 & 0.845 & -1.000 & -1.000 & 0.773 \\
Faster R-CNN \cite{Fasterrcnn} & 0.534 & 0.827 & 0.669 & -1.000 & -1.000 & 0.542 \\
Cascade R-CNN \cite{Cascade} & 0.724 & 0.938 & 0.845 & -1.000 & -1.000 & 0.736 \\
Sparse R-CNN \cite{Sparsercnn} & 0.693 & 0.952 & 0.829 & -1.000 & -1.000 & 0.702 \\
DETR \cite{detr} & 0.722 & 0.959 & 0.821 & -1.000 & -1.000 & 0.732 \\
Deformable-DETR \cite{Deformable} & 0.356 & 0.868 & 0.107 & -1.000 & -1.000 & 0.364 \\
RT-DETR-l \cite{RTDETR} & 0.736 & 0.939 & 0.879 & -1.000 & -1.000 & 0.748 \\
RetinaNet \cite{Retinanet} & 0.743 & 0.961 & 0.876 & -1.000 & -1.000 & 0.755 \\
MobileNetv2-SSD \cite{SSD} & 0.523 & 0.778 & 0.613 & -1.000 & -1.000 & 0.531 \\
FCOS-r50-FPN \cite{Fcos} & 0.686 & 0.908 & 0.792 & -1.000 & -1.000 & 0.697 \\
\bottomrule
\end{tabular}
\label{tab:deepfish-detection}
\end{table}

\begin{table}[t]
\centering
\captionof{table}{Detection model metrics on FISH-Video \cite{fish-video-ls42k_dataset} }
\tiny
\begin{tabular}{lcccccc}
\toprule
Model & mAP & mAP$_{50}$ & mAP$_{75}$ & AP$_S$ & AP$_M$ & AP$_L$ \\
\midrule
YOLOv8n \cite{yolov8_ultralytics} & 0.579 & 0.980 & 0.667 & 0.365 & 0.586 & -1.000 \\
YOLOv8s \cite{yolov8_ultralytics} & 0.580 & 0.999 & 0.601 & 0.299 & 0.586 & -1.000 \\
YOLOv8m \cite{yolov8_ultralytics} & 0.591 & \textbf{1.000} & 0.674 & 0.299 & 0.597 & -1.000 \\
YOLOv8l \cite{yolov8_ultralytics} & 0.585 & 0.999 & 0.655 & 0.333 & 0.591 & -1.000 \\
YOLOv8x \cite{yolov8_ultralytics} & 0.590 & 0.990 & 0.681 & 0.299 & 0.596 & -1.000 \\
YOLOv10n \cite{THU-MIGyolov10} & 0.580 & 0.985 & 0.654 & 0.321 & 0.582 & -1.000 \\
YOLOv10s \cite{THU-MIGyolov10} & 0.580 & 0.999 & 0.629 & 0.327 & 0.585 & -1.000 \\
YOLOv10m \cite{THU-MIGyolov10} & 0.600 & 0.999 & \textbf{0.725} & 0.224 & \textbf{0.608} & -1.000 \\
YOLO11n \cite{yolo11_ultralytics} & 0.567 & 0.969 & 0.589 & 0.234 & 0.575 & -1.000 \\
YOLO11s \cite{yolo11_ultralytics} & 0.585 & 0.999 & 0.654 & 0.233 & 0.593 & -1.000 \\
YOLO11m \cite{yolo11_ultralytics} & 0.591 & 0.990 & 0.683 & 0.199 & 0.599 & -1.000 \\
YOLO11l \cite{yolo11_ultralytics} & 0.583 & \textbf{1.000} & 0.667 & 0.233 & 0.591 & -1.000 \\
YOLO11x \cite{yolo11_ultralytics} & 0.585 & 0.990 & 0.623 & 0.233 & 0.594 & -1.000 \\
YOLO12n \cite{yolo12} & 0.578 & \textbf{1.000} & 0.660 & 0.333 & 0.582 & -1.000 \\
YOLO12s \cite{yolo12} & 0.567 & \textbf{1.000} & 0.608 & 0.200 & 0.576 & -1.000 \\
YOLO12m \cite{yolo12} & 0.594 & \textbf{1.000} & 0.673 & 0.267 & 0.600 & -1.000 \\
YOLO12l \cite{yolo12} & 0.594 & \textbf{1.000} & 0.692 & 0.266 & 0.599 & -1.000 \\
YOLO12x \cite{yolo12} & 0.593 & \textbf{1.000} & 0.651 & 0.266 & 0.600 & -1.000 \\
YOLO-NAS-l \cite{supergradients} & 0.595 & 0.999 & 0.668 & 0.299 & 0.602 & -1.000 \\
Faster R-CNN \cite{Fasterrcnn} & 0.533 & 0.952 & 0.530 & 0.278 & 0.536 & -1.000 \\
Cascade R-CNN \cite{Cascade} & 0.582 & 0.989 & 0.624 & \textbf{0.494} & 0.583 & -1.000 \\
Sparse R-CNN \cite{Sparsercnn} & 0.540 & 0.979 & 0.563 & \textbf{0.611} & 0.540 & -1.000 \\
DETR \cite{detr} & 0.523 & 0.971 & 0.492 & 0.544 & 0.525 & -1.000 \\
Deformable-DETR \cite{Deformable} & 0.345 & 0.915 & 0.094 & 0.451 & 0.350 & -1.000 \\
RT-DETR-l \cite{RTDETR} & 0.588 & 0.999 & 0.651 & 0.435 & 0.594 & -1.000 \\
RetinaNet \cite{Retinanet} & 0.583 & 0.989 & 0.638 & 0.553 & 0.583 & -1.000 \\
MobileNetv2-SSD \cite{SSD} & 0.471 & 0.874 & 0.421 & 0.089 & 0.483 & -1.000 \\
FCOS-r50-FPN \cite{Fcos} & 0.567 & 0.977 & 0.655 & 0.213 & 0.573 & -1.000 \\
\bottomrule
\end{tabular}
\label{tab:fishvideo2-detection}
\end{table}

\begin{table}[t]{}
\centering
\captionof{table}{Detection model metrics on AAS \cite{Le2023} }
\tiny
\begin{tabular}{lcccccc}
\toprule
Model & mAP & mAP$_{50}$ & mAP$_{75}$ & AP$_S$ & AP$_M$ & AP$_L$ \\
\midrule
YOLOv8n \cite{yolov8_ultralytics} & 0.424 & 0.651 & 0.474 & 0.071 & 0.292 & 0.452 \\
YOLOv8s \cite{yolov8_ultralytics} & 0.480 & 0.699 & 0.532 & 0.096 & 0.341 & 0.509 \\
YOLOv8m \cite{yolov8_ultralytics} & 0.518 & 0.728 & 0.590 & 0.064 & 0.344 & 0.553 \\
YOLOv8l \cite{yolov8_ultralytics} & 0.542 & 0.743 & 0.611 & 0.081 & 0.360 & 0.578 \\
YOLOv8x \cite{yolov8_ultralytics} & 0.562 & 0.754 & 0.638 & \textbf{0.100} & 0.347 & 0.599 \\
YOLOv10n \cite{THU-MIGyolov10} & 0.478 & 0.687 & 0.532 & 0.085 & 0.318 & 0.507 \\
YOLOv10s \cite{THU-MIGyolov10} & 0.530 & 0.727 & 0.598 & 0.092 & 0.357 & 0.564 \\
YOLOv10m \cite{THU-MIGyolov10} & 0.608 & 0.776 & 0.663 & 0.038 & 0.380 & 0.645 \\
YOLO11n \cite{yolo11_ultralytics} & 0.520 & 0.707 & 0.585 & 0.096 & 0.330 & 0.553 \\
YOLO11s \cite{yolo11_ultralytics} & 0.593 & 0.765 & 0.661 & 0.082 & 0.384 & 0.625 \\
YOLO11m \cite{yolo11_ultralytics} & 0.607 & 0.776 & 0.661 & 0.070 & 0.380 & 0.643 \\
YOLO11l \cite{yolo11_ultralytics} & \textbf{0.615} & 0.789 & 0.669 & 0.093 & 0.386 & 0.650 \\
YOLO11x \cite{yolo11_ultralytics} & 0.586 & 0.763 & 0.652 & \textbf{0.100} & 0.375 & 0.622 \\
YOLO12n \cite{yolo12} & 0.462 & 0.675 & 0.525 & 0.072 & 0.310 & 0.496 \\
YOLO12s \cite{yolo12} & 0.560 & 0.744 & 0.620 & 0.086 & 0.357 & 0.595 \\
YOLO12m \cite{yolo12} & 0.589 & 0.770 & 0.659 & 0.090 & 0.377 & 0.625 \\
YOLO12l \cite{yolo12} & \textbf{0.647} & 0.797 & \textbf{0.701} & 0.093 & \textbf{0.419} & \textbf{0.679} \\
YOLO12x \cite{yolo12} & 0.645 & \textbf{0.802} & 0.699 & 0.062 & 0.407 & \textbf{0.680} \\
YOLO-NAS-l \cite{supergradients} & 0.546 & 0.723 & 0.611 & 0.054 & 0.361 & 0.584 \\
Faster R-CNN \cite{Fasterrcnn} & 0.330 & 0.566 & 0.362 & 0.026 & 0.206 & 0.361 \\
Cascade R-CNN \cite{Cascade} & 0.465 & 0.689 & 0.528 & 0.082 & 0.334 & 0.495 \\
Sparse R-CNN \cite{Sparsercnn} & 0.444 & 0.717 & 0.493 & 0.047 & 0.326 & 0.464 \\
DETR \cite{detr} & 0.505 & 0.699 & 0.584 & 0.066 & 0.312 & 0.555 \\
Deformable-DETR \cite{Deformable} & 0.453 & 0.754 & 0.534 & 0.080 & 0.268 & 0.480 \\
RT-DETR-l \cite{RTDETR} & 0.526 & 0.737 & 0.612 & 0.062 & 0.291 & 0.560 \\
RetinaNet \cite{Retinanet} & 0.536 & 0.775 & 0.608 & 0.051 & 0.326 & 0.566 \\
MobileNetv2-SSD \cite{SSD} & 0.341 & 0.540 & 0.380 & 0.019 & 0.136 & 0.383 \\
FCOS-r50-FPN \cite{Fcos} & 0.461 & 0.707 & 0.504 & 0.087 & 0.279 & 0.497 \\
\bottomrule
\end{tabular}
\label{tab:aas-detection}
\end{table}

\begin{table}[t]
\centering
\captionof{table}{Detection model metrics on FishExtend \cite{fish-clean_dataset} }
\tiny
\begin{tabular}{lcccccc}
\toprule
Model & mAP & mAP$_{50}$ & mAP$_{75}$ & AP$_S$ & AP$_M$ & AP$_L$ \\
\midrule
YOLOv8n \cite{yolov8_ultralytics} & 0.506 & 0.771 & 0.537 & -1.000 & 0.427 & 0.619 \\
YOLOv8s \cite{yolov8_ultralytics} & 0.533 & 0.785 & 0.596 & -1.000 & 0.447 & 0.652 \\
YOLOv8m \cite{yolov8_ultralytics} & 0.536 & 0.783 & 0.586 & -1.000 & 0.419 & 0.678 \\
YOLOv8l \cite{yolov8_ultralytics} & 0.560 & 0.807 & 0.620 & -1.000 & 0.456 & 0.703 \\
YOLOv8x \cite{yolov8_ultralytics} & 0.552 & 0.795 & 0.610 & -1.000 & 0.438 & 0.704 \\
YOLOv10n \cite{THU-MIGyolov10} & 0.517 & 0.767 & 0.559 & -1.000 & 0.405 & 0.670 \\
YOLOv10s \cite{THU-MIGyolov10} & 0.525 & 0.782 & 0.550 & -1.000 & 0.395 & 0.684 \\
YOLOv10m \cite{THU-MIGyolov10} & 0.578 & 0.829 & 0.618 & -1.000 & 0.473 & 0.724 \\
YOLO11n \cite{yolo11_ultralytics} & 0.539 & 0.788 & 0.609 & -1.000 & 0.408 & 0.698 \\
YOLO11s \cite{yolo11_ultralytics} & 0.579 & 0.826 & 0.651 & -1.000 & \textbf{0.483} & 0.708 \\
YOLO11m \cite{yolo11_ultralytics} & 0.574 & 0.828 & 0.627 & -1.000 & 0.467 & 0.711 \\
YOLO11l \cite{yolo11_ultralytics} & 0.582 & \textbf{0.837} & 0.626 & -1.000 & 0.470 & \textbf{0.730} \\
YOLO11x \cite{yolo11_ultralytics} & 0.566 & 0.826 & 0.608 & -1.000 & 0.459 & 0.721 \\
YOLO12n \cite{yolo12} & 0.522 & 0.781 & 0.567 & -1.000 & 0.415 & 0.677 \\
YOLO12s \cite{yolo12} & 0.560 & 0.810 & 0.612 & -1.000 & 0.459 & 0.705 \\
YOLO12m \cite{yolo12} & 0.571 & 0.817 & 0.627 & -1.000 & 0.449 & 0.721 \\
YOLO12l \cite{yolo12} & \textbf{0.587} & \textbf{0.847} & \textbf{0.658} & -1.000 & 0.477 & 0.729 \\
YOLO12x \cite{yolo12} & 0.576 & 0.822 & 0.637 & -1.000 & 0.460 & 0.722 \\
YOLO-NAS-l \cite{supergradients} & 0.569 & 0.816 & 0.628 & -1.000 & 0.444 & 0.717 \\
Faster R-CNN \cite{Fasterrcnn} & 0.429 & 0.707 & 0.452 & -1.000 & 0.292 & 0.581 \\
Cascade R-CNN \cite{Cascade}& 0.536 & 0.791 & 0.596 & -1.000 & 0.415 & 0.672 \\
Sparse R-CNN \cite{Sparsercnn} & 0.452 & 0.716 & 0.483 & -1.000 & 0.303 & 0.612 \\
DETR \cite{detr} & 0.465 & 0.769 & 0.490 & -1.000 & 0.310 & 0.677 \\
Deformable-DETR \cite{Deformable} & 0.241 & 0.709 & 0.020 & -1.000 & 0.168 & 0.325 \\
RT-DETR-l \cite{RTDETR} & 0.549 & 0.814 & 0.600 & -1.000 & 0.408 & 0.705 \\
RetinaNet \cite{Retinanet} & 0.534 & 0.797 & 0.581 & -1.000 & 0.413 & 0.680 \\
MobileNetv2-SSD \cite{SSD} & 0.458 & 0.783 & 0.435 & -1.000 & 0.351 & 0.593 \\
FCOS-r50-FPN \cite{Fcos} & 0.503 & 0.785 & 0.515 & -1.000 & 0.369 & 0.645 \\
\bottomrule
\end{tabular}
\label{tab:fishextend-detection}
\end{table}


\begin{figure}[t]
\tiny
\centering
\begin{lstlisting}
YOLO_MODELS_WITH_DESCRIPTIONS = {
    "yolov10m.pt": "A robust YOLOv10 model, good for detecting a moderate number of fish in typical aquatic conditions.",
    "yolov10n.pt": "An ultra-fast YOLOv10 model, ideal for detecting fish in live video streams or on low-power underwater cameras where speed is critical.",
    "yolov10s.pt": "A small and agile YOLOv10 model, suitable for fish detection in clear water with decent speed.",
    "yolov8x.pt": "The most precise YOLOv8 model, excellent for identifying individual fish or rare species in high-resolution images with many overlapping fish or challenging backgrounds.",
    "yolov8n.pt": "A featherlight YOLOv8 model for rapid fish presence detection, perfect for high-throughput video analysis.",
    "yolov8s.pt": "A balanced YOLOv8 model for general fish detection, offering good speed and accuracy for common scenarios.",
    "yolov8m.pt": "A versatile YOLOv8 model for fish detection, handling a range of fish sizes and water clarities effectively.",
    "yolov8l.pt": "A powerful YOLOv8 model for high-resolution fish detection, capturing subtle details in diverse aquatic environments.",
    "yolo12x.pt": "An extremely precise YOLOv12 model, designed for critical fish monitoring where missed detections are costly, even in complex, dark, or murky waters.",
    "yolo12n.pt": "An ultra-efficient YOLOv12 model for immediate fish alerts, perfect for simple presence detection in real-time streams.",
    "yolo12s.pt": "A compact YOLOv12 model, balancing detection accuracy and speed for typical fish surveys.",
    "yolo12m.pt": "A highly adaptive YOLOv12 model, effective for varied fish populations and environmental conditions.",
    "yolo12l.pt": "A high-performance YOLOv12 model, strong for detecting diverse fish species with robust detail capture.",
    "yolo11x.pt": "A top-tier YOLOv11 model, excelling at identifying a wide variety of fish in high-definition imagery, even with occlusions.",
    "yolo11s.pt": "An efficient YOLOv11 model for quick fish counts and general monitoring in relatively clear water.",
    "yolo11m.pt": "A reliable YOLOv11 model, a workhorse for consistent fish detection performance across different lighting.",
    "yolo11n.pt": "The most lightweight YOLOv11 model, specifically tuned for identifying fish on tiny, battery-operated underwater sensors.",
    "yolo11l.pt": "A specialized YOLOv11 model, expertly trained on extensive marine datasets for comprehensive and accurate fish identification.",
    "RT_DETR_L.pt": "An advanced transformer-based model, potentially superior for detecting fish with complex shapes or spatial arrangements, offering robust performance in structured environments."
}
\end{lstlisting}
\label{CLIP_prompt}
\end{figure}

\bibliographystyle{IEEEtran}
\bibliography{fixed_citations_cleaned}